%% file: main.tex
\documentclass[10pt,twocolumn,letterpaper]{article}

\usepackage{cvpr}      %

\usepackage{comment}
\usepackage[percent]{overpic}
\usepackage{multirow}

\include{packages}
\include{macros}
\usepackage{array}
\newcolumntype{P}[1]{>{\centering\arraybackslash}p{#1}}

\begin{document}

\title{Latent-NeRF for Shape-Guided Generation of 3D Shapes and Textures}

\author{Gal Metzer\textsuperscript{*}
\qquad
Elad Richardson\textsuperscript{*}
\qquad
Or Patashnik
\qquad
Raja Giryes
\qquad
Daniel Cohen-Or\\ \\
Tel Aviv University 
}

\twocolumn[{%
\renewcommand\twocolumn[1][]{#1}%
\vspace{-1em}
\maketitle
\vspace{-1em}
\input{figures/teaser/fig}
}]

\input{0_abstract}
\vspace{-0.75cm}
\input{1_introduction}

\input{2_related_work}

\input{3_method}

\input{4_experiments}

\input{6_conclusion.tex}

{\small
\bibliographystyle{ieee_fullname}
\bibliography{egbib}
}

\end{document}

%% file: packages.tex
\usepackage{graphicx}
\usepackage{amsmath}
\usepackage{amssymb}
\usepackage{booktabs}

\usepackage[pagebackref,breaklinks,colorlinks]{hyperref}

\usepackage[capitalize]{cleveref}

\usepackage{xcolor}

\usepackage{algorithm, algorithmic}

\usepackage{lipsum}  
\usepackage{pifont}
\usepackage{enumitem,amssymb}

\usepackage{physics}

%% file: macros.tex
\crefname{section}{Sec.}{Secs.}
\Crefname{section}{Section}{Sections}
\Crefname{table}{Table}{Tables}
\crefname{table}{Tab.}{Tabs.}

\definecolor{mydarkgreen}{rgb}{0.02,0.6,0.02}

\newif\ifdraft
\draftfalse

\newcommand{\meshsketch}[0]{Sketch-Shape}

\newcommand{\latentpaint}[0]{Latent-Paint}

\newlist{todolist}{itemize}{2}
\setlist[todolist]{label=$\square$}

%% file: figures/teaser/fig.tex
\begin{center}
    \centering
    \vspace{-0.2cm}
    {\scriptsize 
    \newcommand{\pl}{-8.0}
     \setlength{\fboxrule}{0pt}
    \begin{overpic}[width=0.96\linewidth]{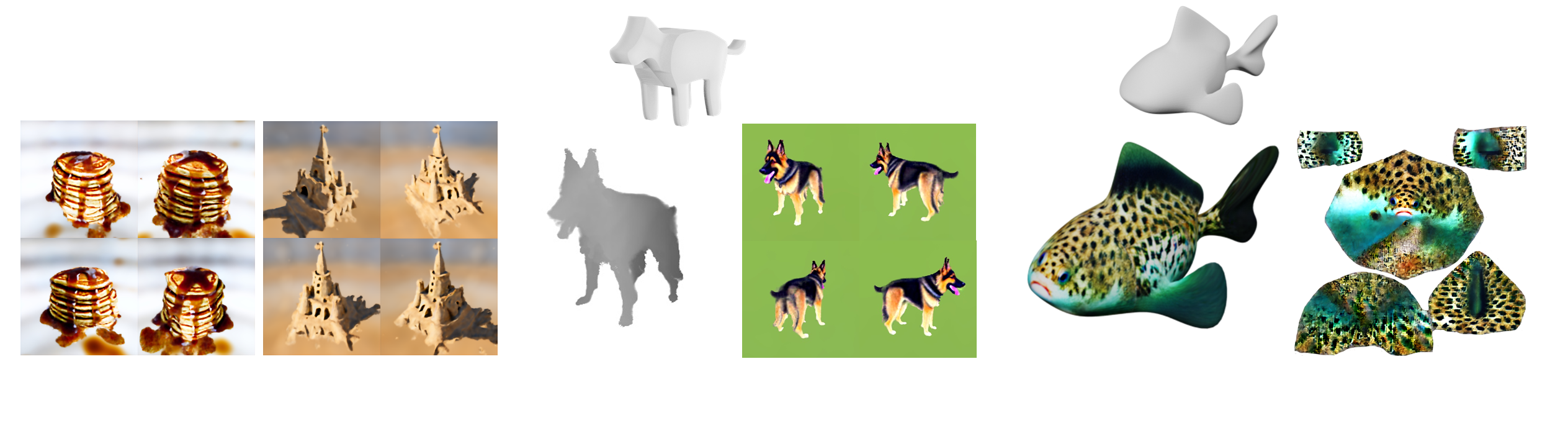}
    \put(10,  1.0){\textcolor{black}{\large Latent-NeRF}}
    \put(42,  1.0){\textcolor{black}{\large \meshsketch}}
    \put(75,  1.0){\textcolor{black}{\large Latent-Paint}}
    \put(3.2,  22.5){\fbox{%
    \parbox{0.1\textwidth}{
    \centering
    \textcolor{black}{ ``A stack of pancakes covered in maple syrup''}
    }%
    }}
    \put(16.2,  22.5){\fbox{%
    \parbox{0.15\textwidth}{
    \centering
    \textcolor{black}{ ``A highly detailed sandcastle''}
    }%
    }}
    \put(47,  22.5){\fbox{%
    \parbox{0.1\textwidth}{
    \centering
    \textcolor{black}{ ``A German Shepherd''}
    }%
    }}
    \put(82, 22.5){\fbox{%
    \parbox{0.1\textwidth}{
    \centering
    \textcolor{black}{ ``A fish with leopard spots''}
    }%
    }}
    \end{overpic}
    }
    \captionof{figure}{Our three text-guided models: a purely text-guided Latent-NeRF, Latent-NeRF with \meshsketch{} guidance for more exact control over the generated shape, and Latent-Paint for texture generation for explicit shapes. The top row represents the models' inputs.}
    \label{teaser}
\end{center}%

%% file: 0_abstract.tex
\begin{abstract}
\vspace{-0.4cm}
Text-guided image generation has progressed rapidly in recent years, inspiring major breakthroughs in text-guided shape generation. Recently, it has been shown that using score distillation, one can successfully text-guide a NeRF model to generate a 3D object. We adapt the score distillation to the publicly available, and computationally efficient, Latent Diffusion Models, which apply the entire diffusion process in a compact latent space of a pretrained autoencoder. As NeRFs operate in image space, a na\"{i}ve solution for guiding them with latent score distillation would require encoding to the latent space at each guidance step. Instead, we propose to bring the NeRF to the latent space, resulting in a Latent-NeRF.
Analyzing our Latent-NeRF, we show that while Text-to-3D models can generate impressive results, they are inherently unconstrained and may lack the ability to guide or enforce a specific 3D structure. To assist and direct the 3D generation, we propose to guide our Latent-NeRF using a \meshsketch: an abstract geometry that defines the coarse structure of the desired object. Then, we present means to integrate such a constraint directly into a Latent-NeRF. This unique combination of text and shape guidance allows for increased control over the generation process.
We also show that latent score distillation can be successfully applied directly on 3D meshes. This allows for generating high-quality textures on a given geometry. Our experiments validate the power of our different forms of guidance and the efficiency of using latent rendering.

\end{abstract}

%% file: 1_introduction.tex
\section{Introduction}
\label{seq:intro}
Text-guided image generation has seen tremendous success in recent years, primarily due to the breathtaking development in Language-Image models~\cite{radford2021learning, jia2021scaling,li2022blip} and diffusion models~\cite{rombach2021highresolution, imagen, ramesh2021zeroshot, dhariwal2021diffusion,ho2020denoising,ramesh2022hierarchical}.
These breakthroughs have also resulted in fast progression for text-guided shape generation~\cite{zeng2022lion,michel2022text2mesh,chen2022tango}. Most recently, it has been shown~\cite{poole2022dreamfusion} that one can directly use score distillation from a 2D diffusion model to guide the generation of a 3D object represented as a Neural Radiance Field (NeRF)~\cite{mildenhall2021nerf}.

While Text-to-3D can generate impressive results, it is inherently unconstrained and may lack the ability to guide or enforce a 3D structure.
In this paper, we show how to introduce shape-guidance to the generation process to guide it toward a specific shape, thus allowing increased control over the generation process.
Our method builds upon two models, a NeRF model \cite{mildenhall2021nerf}, and a Latent Diffusion Model (LDM)~\cite{rombach2021highresolution}.
Latent Models, which apply the entire diffusion process in a compact latent space, have recently gained popularity due to their efficiency and publicly available pretrained checkpoints. 
As score distillation was previously applied only on RGB diffusion models, we first present two key modifications to the NeRF model that are better paired with guidance from a latent model. First, instead of representing our NeRF in the standard RGB space, we propose a \textit{Latent-NeRF} which operates directly in the latent space of the LDM, thus avoiding the burden of encoding a rendered RGB image to a latent space for each and every guiding step.
Secondly, we show that after training, one can easily transform a Latent-NeRF back into a regular NeRF. This allows further refinement in RGB space, where we can also introduce shading constraints or apply further guidance from RGB diffusion models~\cite{imagen}. This is achieved by introducing a learnable linear layer that can be optionally added to a trained Latent-NeRF, where the linear layer is initialized using an approximate mapping between the latent and RGB values~\cite{linear_latent_approximation}.

Our first form of shape-guidance is applied using a coarse 3D model, which we call a \textit{\meshsketch{}}.
Given a \meshsketch{}, we apply soft constraint during the NeRF optimization process to guide its occupancy based on the given shape. Easily combined with Latent-NeRF optimization, the additional constraint can be tuned to meet a desired level of strictness.
Using a \meshsketch{} allows users to define their base geometry, where Latent-NeRF then refines the shape and introduces texture based on a guiding prompt. 

We further present \textit{\latentpaint}, another form of shape-guidance where the generation process is applied directly on a given 3D mesh, and we have not only the structure but also the exact parameterization of the input mesh. This is achieved by representing a texture map in the latent space and propagating the guidance gradients directly to the texture map through the rendered mesh. By doing so, we allow for the first time to colorize a mesh using guidance from a pretrained diffusion model and enjoy its expressiveness. 

We evaluate our different forms of guidance under a variety of scenarios and show that together with our latent-based guidance, they offer a compelling solution for constrained shape and texture generation.

%% file: 2_related_work.tex
\section{Related Work}
\label{seq:related}

\paragraph{3D Shape Generation}
3D shape synthesis is a longstanding problem in computer graphics and computer vision.
In recent years, with the emergence of neural networks, the research in 3D modeling has immensely advanced.
The most conventional supervision type is applied directly with 3D shapes, through different representations such as implicit functions~\cite{IMNET,park2019deepsdf,hertz2022spaghetti}, meshes~\cite{gao2019sdm, yang2022dsg} or point clouds~\cite{yang2019pointflow, li2018point}.
As 3D supervision is often difficult to obtain, other works use images to guide the generative task~\cite{ Chan2021piGANPI, chan2022efficient, Niemeyer2021CAMPARICD}. In fact, even when 3D data is available, 2D renderings are sometimes chosen as the supervising primitive~\cite{chen2021ngp, gao2022get3d, bautista2022gaudi}. 
For example, in GET3D~\cite{gao2022get3d}, two generators are trained, one generates a 3D SDF, and the other a texture field. The output textured mesh is then obtained in a differentiable manner by utilizing DMTet~\cite{shen2021dmtet}. These generators are adversarially trained with a dataset of 2D images.
In \cite{Watson2022Novel} a diffusion model has been used to generate multiple views of a given input image. Yet, it has been trained in a supervised manner on a multi-view dataset, unlike our work which does not require a dataset.

\paragraph{Text-to-3D with 2D Supervision}

Recently, the success of text-guided synthesis in numerous domains~\cite{patashnik2021styleclip, gal2022stylegan, tevet2022motionclip, avrahami2022blended_latent, bar2022text2live}, has motivated a surge of works that use Language-Image models to guide 3D scenes representations.
CLIP-Forge~\cite{sanghi2021clipforge} consists of two separate components, an implicit autoencoder conditioned on shape codes, and a normalizing flow model that is trained to generate shape codes according to CLIP embeddings. CLIP-Forge exploits the fact that CLIP has a joint text-image embedding space to train on image embeddings and infer on text embeddings, achieving text-to-shape capabilities.
Text2Mesh~\cite{michel2022text2mesh} introduced mesh colorization and geometric fine-tuning by optimizing an initial mesh through differential rendering and CLIP~\cite{radford2021learning} guidance.
TANGO~\cite{chen2022tango} follows a similar optimization scheme, while improving results by considering an explicit shading model. 
CLIP-Mesh~\cite{khalid2022clipmesh} optimizes an initial spherical mesh according to a target text prompt, using a modified CLIP loss that accounts for the gap and ambiguity between image/text CLIP embeddings. Similarly to our method, they also use UV texture mapping to bake colors into the mesh.
DreamFields~\cite{jain2021dreamfields} employs CLIP guidance as well, but uses NeRFs to represent the 3D object instead of an explicit triangular mesh, together with a dedicated sparsity loss.
CLIPNeRF~\cite{wang2022clip} pretrains a disentangled NeRF representation network on rendered object datasets, which is then used to constraint a NeRF scene optimization under CLIP loss, between random renderings of the NeRF and target image or text CLIP embedding.
DreamFusion~\cite{poole2022dreamfusion} introduced, for the first time, the use of largely successful pretrained 2D diffusion models for text-guided 3D object generation. DreamFusion uses a proprietary 2D diffusion model~\cite{imagen} to supervise the generation of 3D objects represented by NeRFs. 
To guide a NeRF scene using the pretrained diffusion model, the authors derive a \textit{Score-Distillation} loss, see Section~\ref{sec:prelim} for more details.

\paragraph{Neural Rendering}
The recent rapid progression of neural networks has immensely advanced the performance of differential renderers. Particularly NeRF~\cite{mildenhall2021nerf, barron2021mip, mueller2022instant} have shown astounding performance on novel view generation and relighting, also extending to other applications like 3D reconstruction~\cite{yariv2021volume}.
Thanks to their differentiable nature, it has been recently shown that one can introduce different neural objectives during training to guide the 3D modeling.

%% file: 3_method.tex
\section{Method}
\label{sec:method}
Here we present our shape-guidance solution. We describe the Latent-NeRF framework, presented in Figure~\ref{fig:overview_latent_nerf}, and then introduce different guidance controls that can be combined with Latent-NeRF for controlling its generation. Yet, before showing our solution, we provide a quick overview of two recently proposed techniques that are highly relevant to our method.

\subsection{Preliminaries}
\label{sec:prelim}
A {\bf latent diffusion model (LDM)}~\cite{rombach2021highresolution} is a specific form of a diffusion model that is trained to denoise \textit{latent codes} of a pretrained autoencoder, instead of the high-resolution images directly. 
First, an autoencoder composed of an encoder $\mathcal{E} $, and a decoder $\mathcal{D} $ is trained to reconstruct natural images $x \sim X$, where $X$ is the image training dataset, in the following form: $\Tilde{x}=\mathcal{D}(\mathcal{E}(x))$.
The autoencoder is trained with a reconstruction loss, perceptual loss~\cite{zhang2018unreasonable} and a patch-based adversarial loss~\cite{isola2017image}.
Then, given the trained autoencoder, a denoising diffusion probabilistic model (DDPM)~\cite{ho2020denoising} is trained to generate a spatial latent $z$ from noise, according to the distribution $z=\mathcal{E}(x)~s.t.~x\sim X$. 
In order to generate a novel image, a latent $\Tilde{z}$ is sampled from this learned distribution, using the trained DDPM, and passed to the decoder to obtain the final image $\mathcal{D}(\Tilde{z})$. 
Operating in the latent space requires less compute, and leads to faster training and sampling, which makes LDM widely popular. In fact, the recent Stable Diffusion model is also an LDM.

\textbf{Score Distillation} is a method that enables using a diffusion model as a critic, \ie, using it as a loss without explicitly back-propagating through the diffusion process. It has been introduced in DreamFusion~\cite{poole2022dreamfusion} for guiding 3D generation using the Imagen model \cite{imagen}. 
To perform score distillation, noise is first added to a given image (e.g., one view of the NeRF's output). Then, the diffusion model is used to predict the added noise from the noised image. Finally, the difference between the predicted and added noises is used for calculating per-pixel gradients.
For NeRF, the gradients are back-propagated for updating the 3D NeRF model. 

Going into more detail, at each iteration of the score distillation optimization, a rendered image $x$ is noised to a randomly drawn time step $t$,
\begin{equation}
x_t=\sqrt{\Bar{\alpha_t}} x + \sqrt{1 - \Bar{\alpha_t}} \epsilon,
\label{eq:noise_step}
\end{equation}
where $\epsilon \sim \mathcal{N}(0, I)$, and $\Bar{\alpha_t}$ is a time-dependent constant specified by the diffusion model.
Then, the per-pixel score distillation gradients are taken to be
\begin{equation}
\nabla_xL_{SDS}=w(t) (\epsilon_\phi(x_t,t,T)-\epsilon),
\label{eq:sds_loss}
\end{equation}
where $\epsilon_\phi$ is the diffusion model's denoiser (which approximates the noise to be removed), $\phi$ are the denoiser's parameters, $T$ is an optional guiding text prompt, and $w(t)$ is a constant multiplier that depends on $\alpha_t$. 
During training, gradients are propagated from the pixel gradients to the NeRF parameters and gradually change the 3D object.
Please refer to~\cite{poole2022dreamfusion} for the complete details and derivation of Score Distillation.
Note that DreamFusion uses the proprietary Imagen~\cite{imagen} model that is very computationally demanding. We rely on the publicly available Stable Diffusion model and the re-implementation of DreamFusion~\cite{stabledreamfusion} (it operates in the RGB space and not the latent as we propose).%

\input{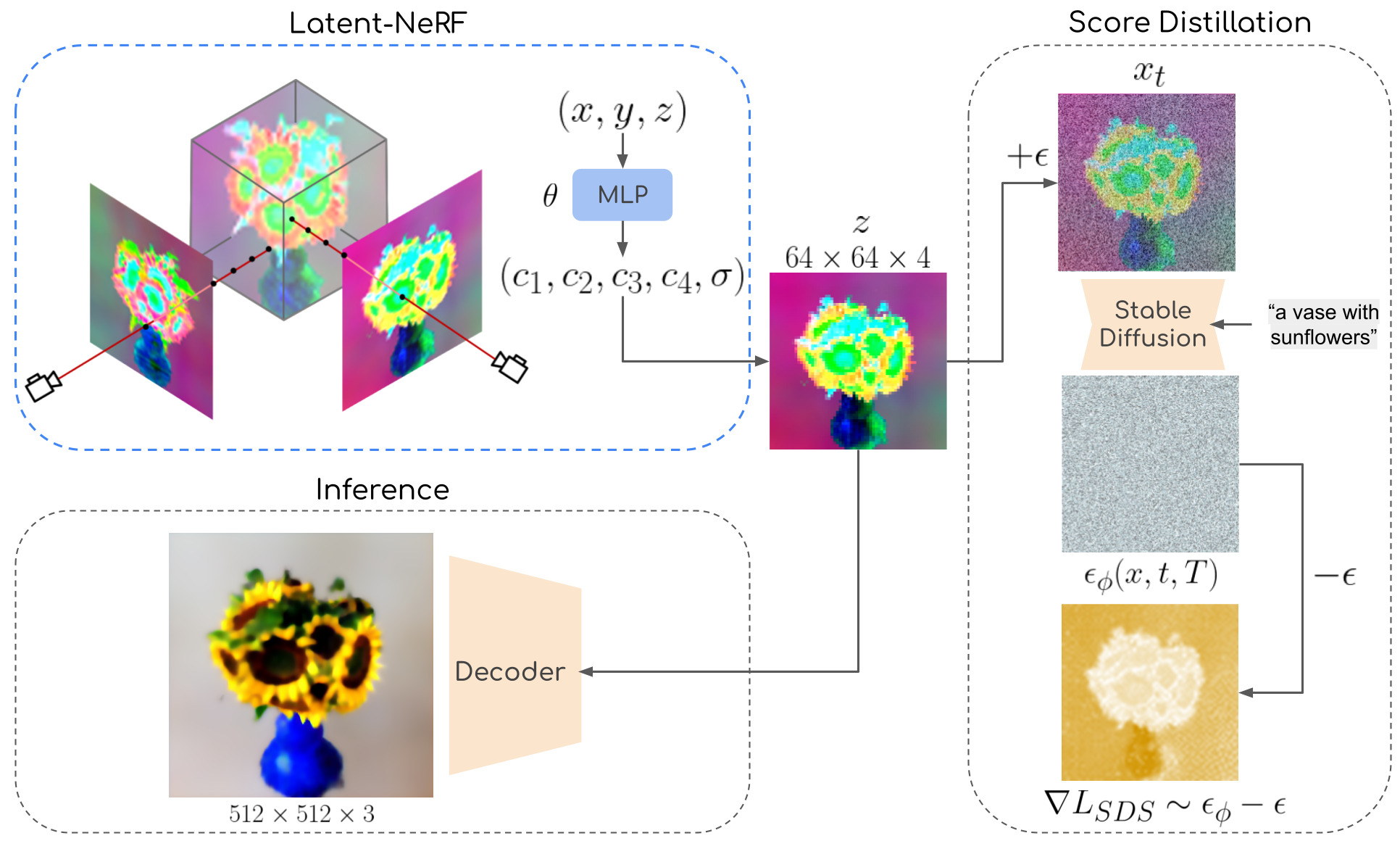}

\subsection{Latent-NeRF}
\label{sec:latent_nerf}
We now turn to describe our Latent-NeRF approach.
In this method, a NeRF model is optimized to render 2D feature maps in Stable Diffusion's latent space $\mathcal{Z}$.
Latent-NeRF outputs four pseudo-color channels, $(c_1, c_2, c_3, c_4)$, corresponding to the four latent features that stable diffusion operates over, and a volume density $\sigma$. Figure~\ref{fig:overview_latent_nerf} illustrates this process.
Representing the scene using NeRF implicitly imposes spatial consistency between different views, due to the spatial radiance field and rendering equation.
Still, the fact that $\mathcal{Z}$ \textbf{can} be represented by a NeRF with spatial consistencies is non-trivial. Previous works~\cite{avrahami2022blended_latent, linear_latent_approximation}
showed that super-pixels in $\mathcal{Z}$ depend mainly on individual patches in the output image.
This can be attributed to the high resolution ($64\times64$) and low channel-wise depth ($4$) of this latent space, which encourages local dependency over the autoencoder's image and latent spaces. 
Assuming $\mathcal{Z}$ is a near patch level representation of its corresponding RGB image makes the latents nearly equivariant to spatial transformations of the scene, which justifies the use of NeRFs for representing the 3D scenes. 

\paragraph{Text Guidance}
The vanilla form of Latent-NeRF is text-guided, with no other constrains for the scene generation. In this setting, we employ the following loss:
$$
L = \lambda_{SDS}L_{SDS}  + \lambda_{sparse}L_{sparse},
$$
where $L_{SDS}$ is the Score-Distillation loss depicted in Figure~\ref{fig:overview_latent_nerf}. 
Note, that the exact value of this loss is not accessible. Instead, the gradients implied by it are approximated by a single forward pass through the denoiser. These gradients are directly passed to the autograd solver.
The loss $L_{sparse}=BE(w_{blend})$ suggested in~\cite{stabledreamfusion} prevents floating ``radiance clouds'' by penalizing the binary entropy of ill-defined background masks $w_{blend}$. Namely, it encourages a strict blending of the object NeRF and background NeRF.

\input{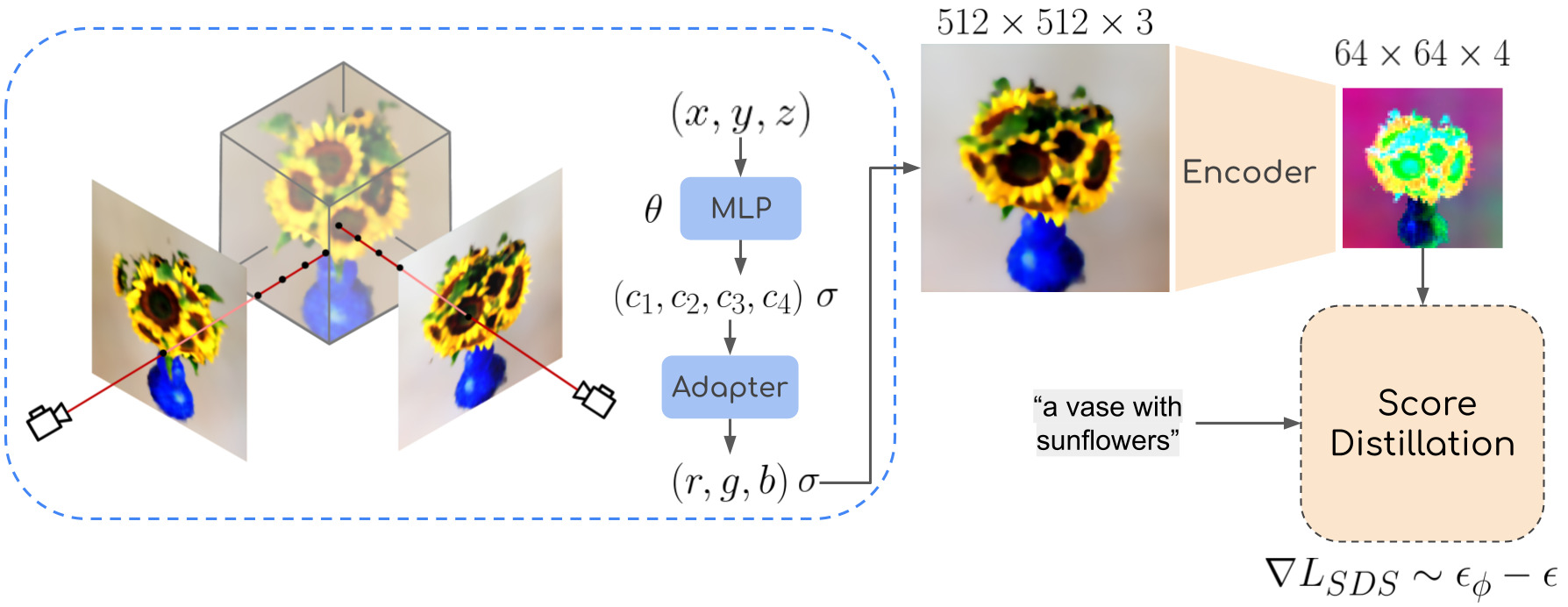}

\paragraph{RGB Refinement}
Using Latent-NeRF, one may successfully learn to represent 3D scenes even when optimizing solely in latent space. Still, in some cases, it could be beneficial to further refine the model by fine-tuning it in pixel space, and have the NeRF model operate directly in RGB.
To do so, we must convert the NeRF that was trained in latent space to a NeRF that operates in RGB.
This requires converting the MLP's output from the four latent channels to three RGB channels such that the initial rendered RGB image is close to the decoder output, when applied to the rendered latent of the original model.
Interestingly, it has been shown~\cite{linear_latent_approximation} that a linear approximation is sufficient to predict plausible RGB colors given a single four-channel latent super pixel, via the following transformation
\begin{equation}
\begin{pmatrix}
           \hat{r} \\
           \hat{g} \\
           \hat{b} 
\end{pmatrix}
=
\begin{pmatrix}
           0.298 & 0.187 & -0.158 & -0.184\\
           0.207 & 0.286 & 0.189 & -0.271\\
           0.208 & 0.173 & 0.264 & -0.473\\

\end{pmatrix}
\begin{pmatrix}
           c_1 \\
           c_2 \\
           c_3 \\
           c_4

\end{pmatrix},
\label{eq:latent_preview}
\end{equation}
which was calculated using pairs of RGB images and their corresponding latent codes over a collection of natural images. 
As our NeRF model is already composed of a set of fully connected layers, we simply add another linear layer that is initialized using the weights in Equation~\ref{eq:latent_preview}. 
This converts our Latent-NeRF to operate in pixel space and ensures that our refinement process starts from a valid result.
The additional layer is then fine-tuned together with the rest of the model, to create the refined and final output.
The overall fine-tuning procedure is illustrated in Figure~\ref{fig:rgb_finetune_overview}.

\input{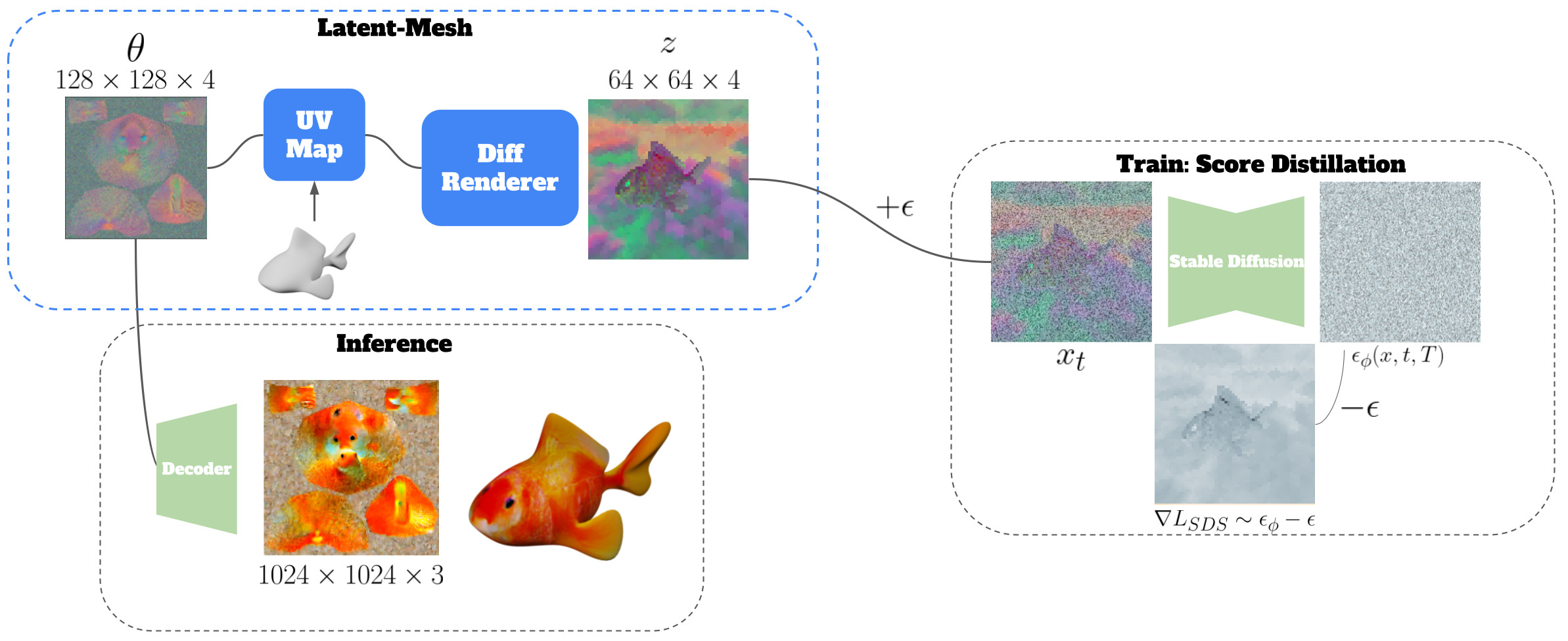}

\subsection{\meshsketch{} Guidance}

Next, we introduce a novel technique for guiding the Latent-NeRF generation based on a coarse geometry, which we call a \meshsketch.
A \meshsketch~is an abstract coarse alignment of simple 3D primitives like spheres, boxes, cylinders, etc., that together depict an outline of a more complex object. Figures~\ref{fig:skecth_mesh_animals}, \ref{fig:skecth_mesh_house}, \ref{fig:general_sketchshapes} illustrate such simple shapes.
Ideally, we would like the output density of our MLP to match that of the \meshsketch, such that the output Latent-NeRF result resembles the input shape. Nevertheless, we would also like the new NeRF to have the capacity to create new details and geometries that match the input text prompt and improve the fidelity of the shape. 
To achieve this lenient constraint, we encourage the NeRF's occupancy to match the winding-number~\cite{jacobson2013robust, barill2018fast} indicator of the \meshsketch,~but with decaying importance near the surface to allow new geometries. This loss reads as
\vspace{-0.2cm}
\begin{equation}
    L_{\meshsketch} = CE(\alpha_{NeRF}(p), \alpha_{GT}(p)) \cdot (1- e^{-\frac{d^2}{2\sigma_S}}).  
    \label{eq:lenient_constraint}
\end{equation}
This loss implies that the occupancy should be well constrained away from the surface, and free to be set by score distillation near the surface. This loss is applied in addition to the Latent-NeRF loss, over the entire point set $p$ that is used by the NeRF's volumetric rendering.  $d$ represents the distance of $p$ from the surface, and $\sigma_S$ is a hyperparameter that controls how lenient the loss is, \textit{i.e.}, lower $\sigma_S$ values imply a tighter constraint to the input \meshsketch. 
Applying the loss only on the sampled point-set $p$, makes it more efficient as these points are already evaluated as part of the Latent-NeRF rendering process.

\subsection{\latentpaint{} of Explicit Shapes}
\label{sec:latent_mesh}
We now move to a more strict constraint, where the guidance is based on an exact structure of a given shape, e.g., provided in the form of a mesh. We call this approach \latentpaint, which leads to the generation of novel textures for a given shape.
Our method generates texture over a UV texture map, which can either be supplied by the input mesh, or calculated on-the-fly using XAtlas~\cite{xatlas}. To color a mesh, we first initialize a random latent texture image of size $H \times  W \times 4$, where $H$ and $W$ can be chosen according to the desired texture granularity. We set them both to be $128$ in our experiments.

Figure~\ref{fig:overview_latent_mesh} presents the training process. At each score distillation iteration, we render the mesh with a differentiable renderer~\cite{KaolinLibrary} to obtain a $64 \times 64 \times 4$ feature map that is \textit{pseudo-colored} by the latent texture image.
Then, we apply the score distillation loss from Equation~\ref{eq:sds_loss} in the same way it is applied for Latent-NeRF. Yet, instead of back-propagating the loss to the NeRF's MLP parameters, we optimize the deep texture image by back-propagating through the differentiable renderer.
To get the final RGB texture image, we simply pass the \textbf{latent texture} image through Stable Diffusion's decoder $\mathcal{D} $ once, to get a larger high-quality RGB texture.

%% file: figures/3_method/overview.tex
\begin{figure}[t]
    \centering
    \includegraphics[width=\columnwidth]{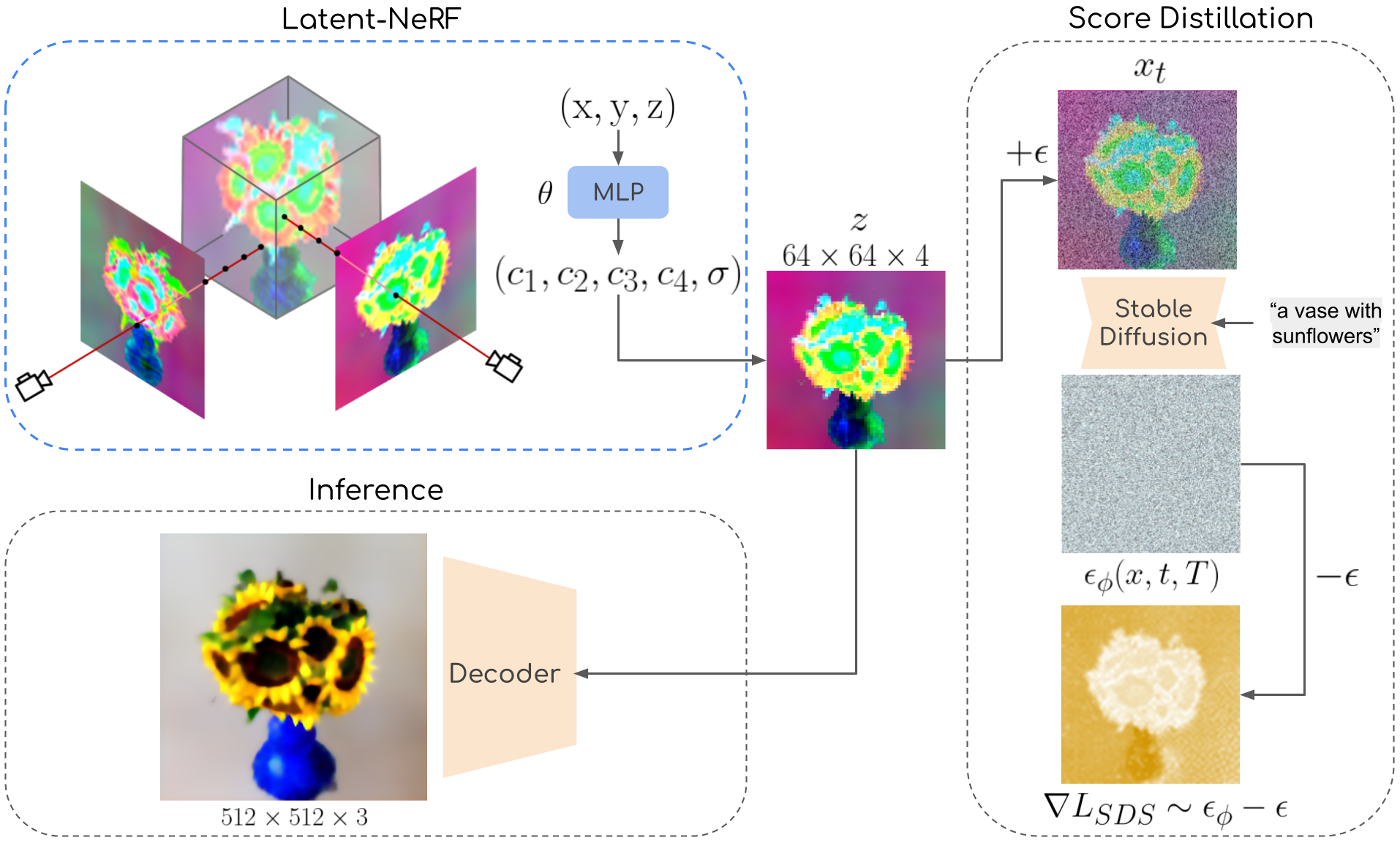}
    \caption{An overview of Latent-NeRF trained with a similar score distillation scheme as proposed by~\cite{poole2022dreamfusion}.
    At each training iteration we render the scene from a random view point to produce a feature map $z$. Then, $z$ is noised with $\epsilon$ according to a random diffusion step $t$. The noised version of $z$, \ie, $x_t$, is denoised using Stable Diffusion~\cite{rombach2021highresolution}, with the input text prompt. Finally, the input noise is subtracted from the predicted noise by Stable Diffusion, to approximate per-pixel gradients that are back propagated to the NeRF representation.}
    \label{fig:overview_latent_nerf}
\end{figure}

%% file: figures/3_method/finetune.tex
\begin{figure}[t]
    \centering
    \includegraphics[width=\columnwidth]{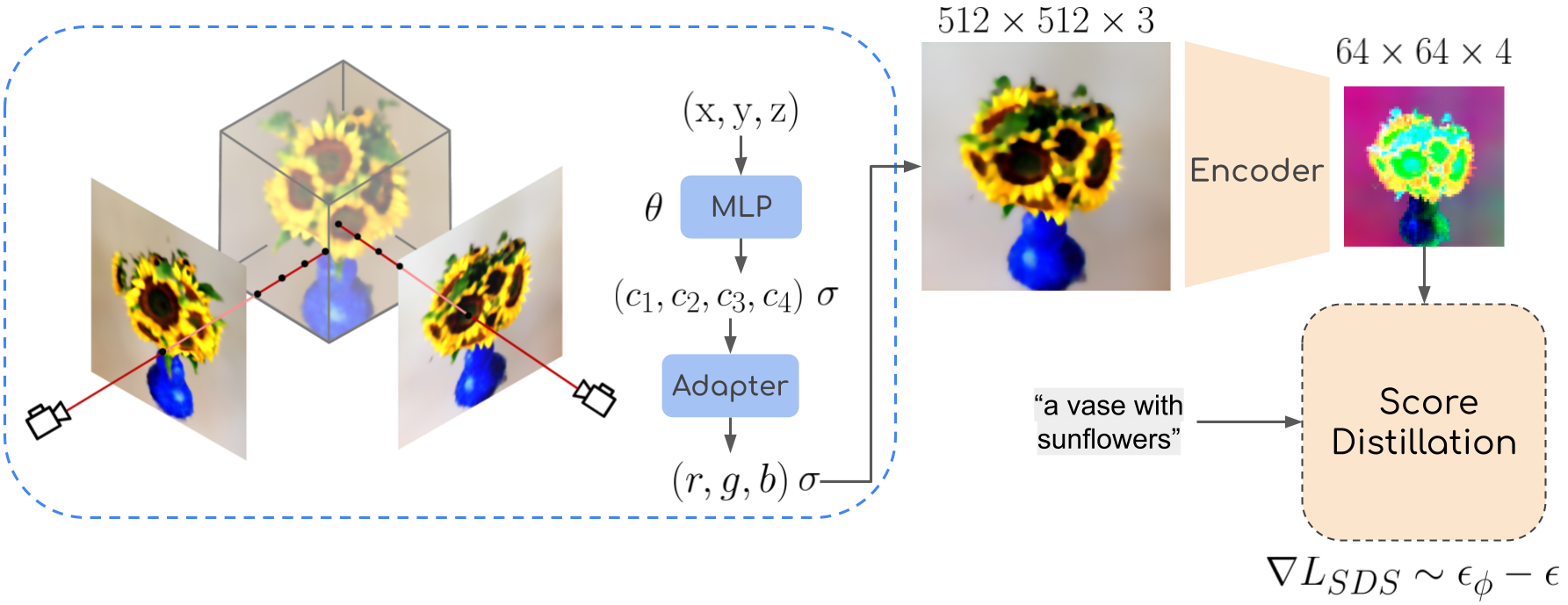}
    \caption{To perform fine-tuning in RGB, we take a Latent-NeRF model that was trained in \textbf{latent space} (conventional setting), apply the matrix adapter to the channel output of the model to get an RGB preview, and then continue optimizing the MLP's weights through supervision in RGB to improve the result. }
    \label{fig:rgb_finetune_overview}
\end{figure}

%% file: figures/3_method/overview_latent_mesh.tex
\begin{figure}[b]
    \centering
    \vspace{-0.4cm}
    \includegraphics[width=\columnwidth]{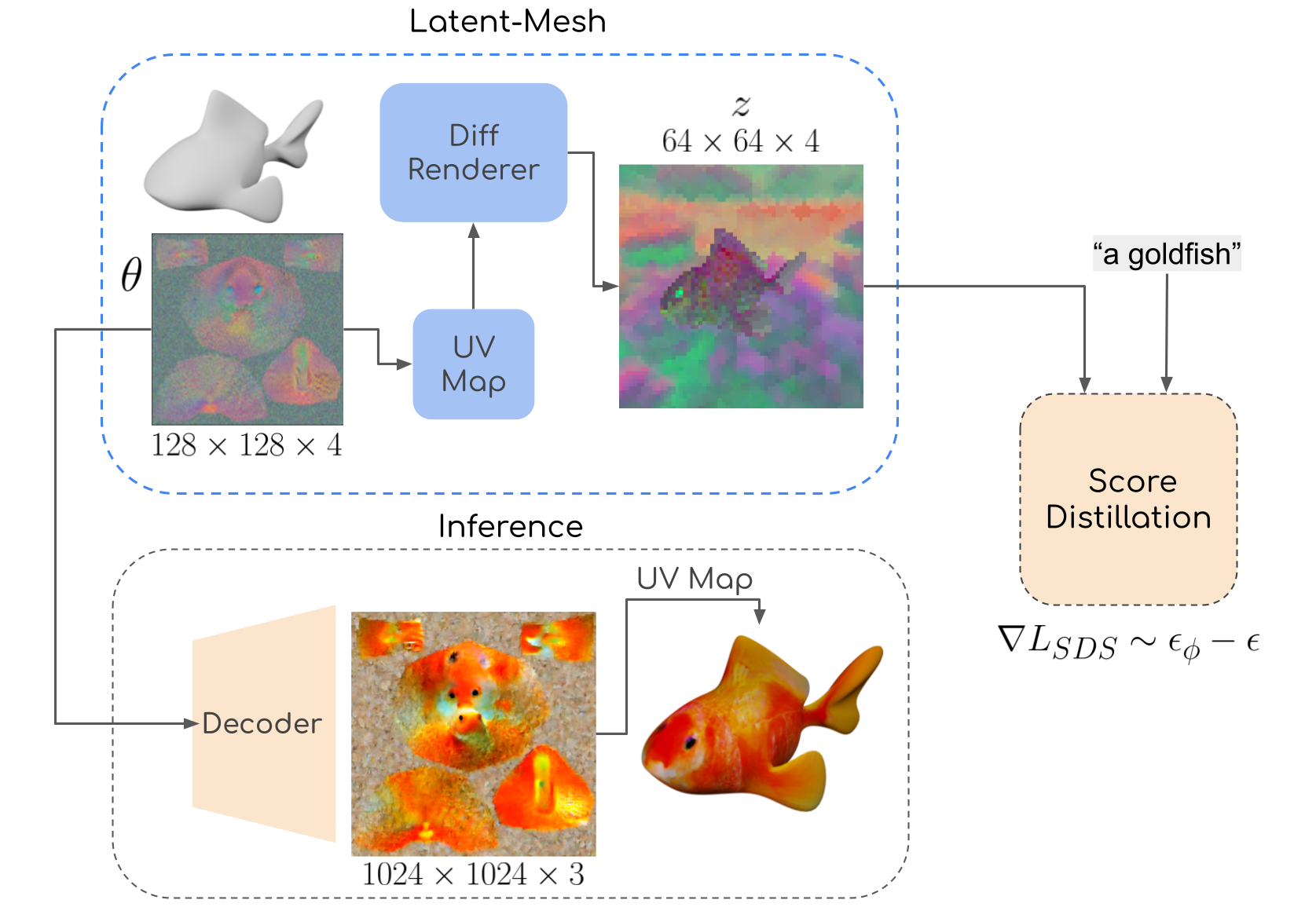}
    \caption{An overview of Latent-Paint texture generation, trained with a similar score distillation loss~\cite{poole2022dreamfusion}. The object is represented as a textured mesh, with a latent texture image that has four pseudo-color channels.
    At each iteration, the mesh is rendered with a differential renderer, to create a feature map from a random viewpoint.
    Score distillation is applied to the rendered feature map, and the gradients back-propagate to the latent texture image through the differential renderer. See Section~\ref{sec:latent_mesh}.
    }
    \label{fig:overview_latent_mesh}
\end{figure}

%% file: 4_experiments.tex
\section{Evaluation}
\input{figures/4_exp/latent_nerf/text2nerf_directions/fig.tex}

\label{sec:experiments}
We now validate the effectiveness of our different forms of guidance through a variety of experiments.
\paragraph{Implementation Details}
We use the Stable Diffusion implementation by HuggingFace Diffusers, with the \texttt{v1-4} checkpoint. For score distillation, we use the code-base provided by \cite{stabledreamfusion}, with Instant-NGP~\cite{mueller2022instant} as our NeRF model. Latent-NeRF usually takes less than 15 minutes to converge on a single V100, while using an RGB-NeRF with Stable Diffusion~\cite{stabledreamfusion} takes about 30 minutes, due to the increased overhead from encoding into the latent space. Note that  DreamFusion~\cite{poole2022dreamfusion} takes about 1.5 hours on 4 TPUs. This clearly shows the computational advantage of Latent-NeRF.

\subsection{Text-Guided Generation}

\input{figures/4_exp/latent_nerf/text2nerf_comparison/fig.tex}

\paragraph{Qualitative Results}
We begin by demonstrating the effectiveness of the latent rendering approach with Latent-NeRF.
In \cref{teaser,fig:text2nerf_directions,fig:rgb_finetune}, we show several results obtained by our method.
In the supplementary material we provide additional results of different objects, including video visualizations.
In Figure~\ref{fig:text2nerf_directions}, we show the consistency of our learned shapes from several viewpoints. 
Next, we use the baseline set by DreamFusion~\cite{poole2022dreamfusion} to qualitatively compare our approach (with the proposed RGB refinement) against other methods.
As can be seen in Figure~\ref{fig:text2nerf_comparison}, Latent-NeRF achieves significantly better results than DreamFields~\cite{jain2021dreamfields} and CLIPMesh~\cite{khalid2022clipmesh}.
We believe that the better quality of DreamFusion can be attributed to the high quality of Imagen~\cite{imagen}, but unfortunately, we cannot validate this as the model is not publicly available to the community.

\paragraph{RGB Refinement}
Figure~\ref{fig:rgb_finetune} shows the quality improvement achieved by our RGB refinement method. It reveals that RGB refinement is mostly useful for complex objects or for regions with detailed textures.
Refinement iterations in the RGB space are about $\times2$ slower than iterations in latent space, thus, increasing the runtime to more than 30 minutes. Thanks to our linear mapping from a Latent-NeRF to an RGB-NeRF, practitioners may apply the refinement method only after the 3D shape has already converged with the more efficient latent training. This allows a fast exploration of 3D shapes, and an optional polishing step with RGB refinement.

\input{figures/4_exp/rgb_finetune/fig.tex}

\input{figures/4_exp/textual_inversion/cat.tex}

\paragraph{Textual-Inversion}
As our Latent-NeRF is supervised by Stable-Diffusion, we can also use \textit{Textual Inversion}~\cite{gal2022textual_inversion} tokens as part of the input text prompt. This allows conditioning the object generation on specific objects and styles, defined only by input images. Results using \textit{Textual Inversion} are presented in Figure~\ref{fig:textual_inversion}.

\input{figures/4_exp/sketch_mesh/animals/fig.tex}
\input{figures/4_exp/sketch_mesh/house/fig.tex}

\input{figures/4_exp/sketch_mesh/general/fig}
\input{figures/4_exp/latent-mesh/misc/misc}

\input{figures/4_exp/latent-mesh/shoes2}
\input{figures/4_exp/sketch_mesh/ablation/fig}

\input{figures/4_exp/latent-mesh/fish}
\input{figures/4_exp/limitations/fig.tex}

\input{figures/4_exp/sketch_mesh/weight_ablation/fig}

\subsection{\meshsketch~Guidance}

Figure~\ref{fig:skecth_mesh_animals} shows different \meshsketch~results with the same conditioning mesh. The different text prompts are able to guide the shape toward refined geometries that better match the text prompt. The rough \meshsketch{} in this figure, was quickly designed in Blender~\cite{blender} and allows us to easily define a coarse shape that guides the Latent-NeRF.
Moreover, Figure~\ref{fig:skecth_weight_ablation} depicts an ablation over the lenient parameter $\sigma_S$ from \cref{eq:lenient_constraint}.
When $\sigma_S$ is set to $0.05$, the generated mesh takes the form of the conditioning shape (shown in Figure~\ref{fig:skecth_mesh_animals}).
As $\sigma_S$ grows, more details are added on top of the base shape, until little to no resemblance is observed at $\sigma_S=1.5$.
Figure~\ref{fig:skecth_mesh_house} contains additional results of different shapes generated with the same conditioning mesh, here a coarse house shape. The normals visualization (bottom row) shows that our method can add fine geometric details.

Figure~\ref{fig:general_sketchshapes} demonstrates that our proposed approach can successfully work with a variety of different \meshsketch s. Notice that our method handles a variety of different shapes and also works well with shapes extruded from 2D sketches.
We also exhibit in Figure~\ref{fig:ablation_prompt} the effectiveness of shape-guidance, by showing results of the same prompts with and without the shape loss.

\subsection{\latentpaint{} Generation}
We tested \latentpaint{} on a variety of input shapes shown in \cref{fig:latent_mesh_misc,fig:latent_mesh_shoe}.
As all of the shapes in these figures do not contain precomputed UV parameterization, we use XAtlas~\cite{xatlas} to compute such parameterization automatically.
In contrast, the fish mesh in Figure~\ref{fig:latent_mesh_fish} (obtained from~\cite{keenan3D}) already contains high quality UV parameterization, which we are also able work with.
Figure~\ref{fig:latent_mesh_shoe} compares our \latentpaint{} approach to two closely related methods, Tango~\cite{chen2022tango} and CLIPMesh~\cite{khalid2022clipmesh}. As can be seen, our approach achieves more precise textures thanks to the guidance from the diffusion model.

Note that \latentpaint{} can work without assuming or computing any UV-map by simply optimizing per-face latent attributes. Yet, we found it better to use a UV-map for two main reasons: (i) The UV map makes the texture granularity independent of the geometric resolution, \textit{i.e.}, coarse geometries do not imply course colorization; and (ii) texture maps are easier to use with downstream applications like MeshLab~\cite{meshlab} and Blender~\cite{blender}.

%% file: figures/4_exp/latent_nerf/text2nerf_directions/fig.tex
\begin{figure}[b]
    \centering
    \setlength{\tabcolsep}{0.5pt}
    {\scriptsize
    \begin{tabular}{c c c c c}
        \includegraphics[width=0.19\linewidth]{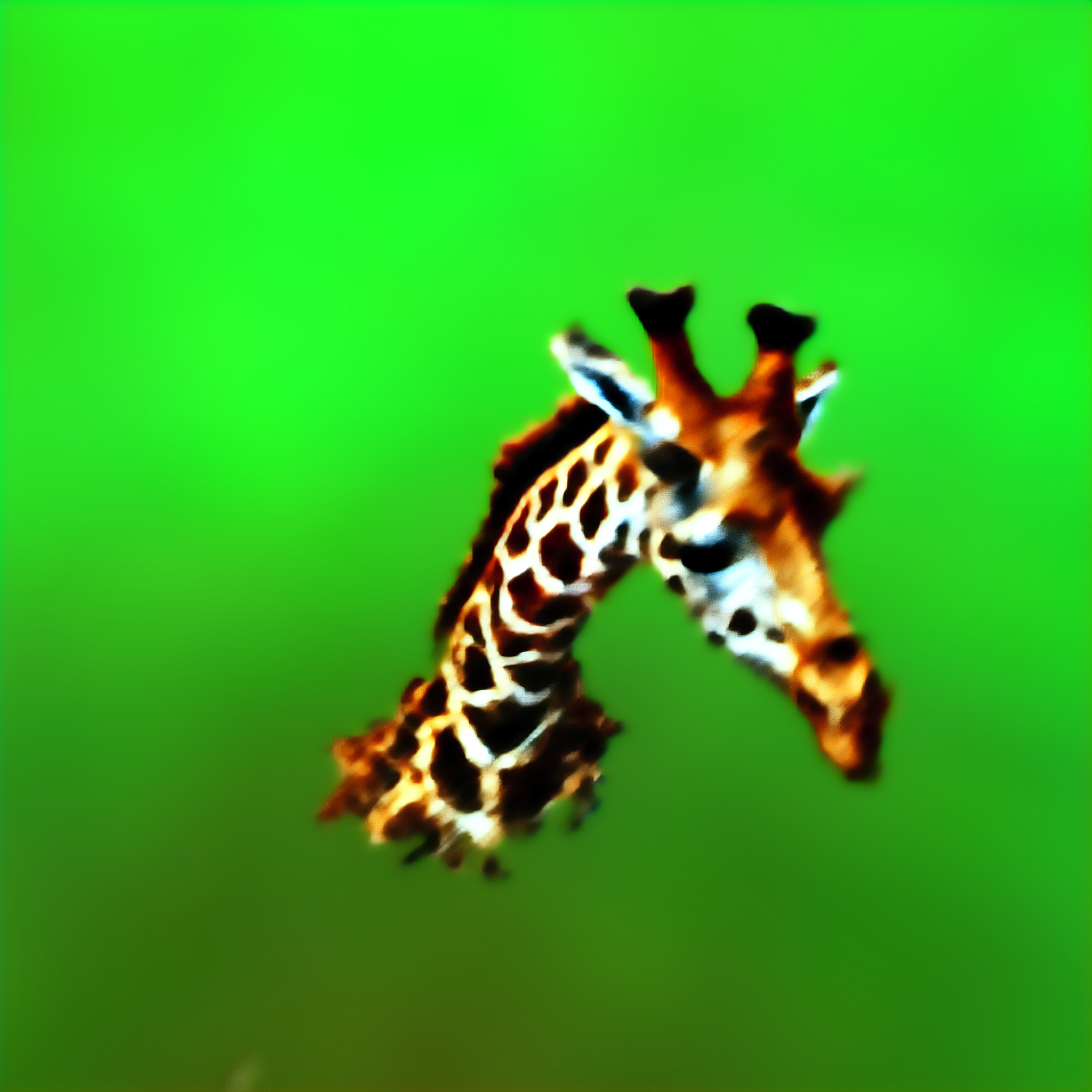} & 
        \includegraphics[width=0.19\linewidth]{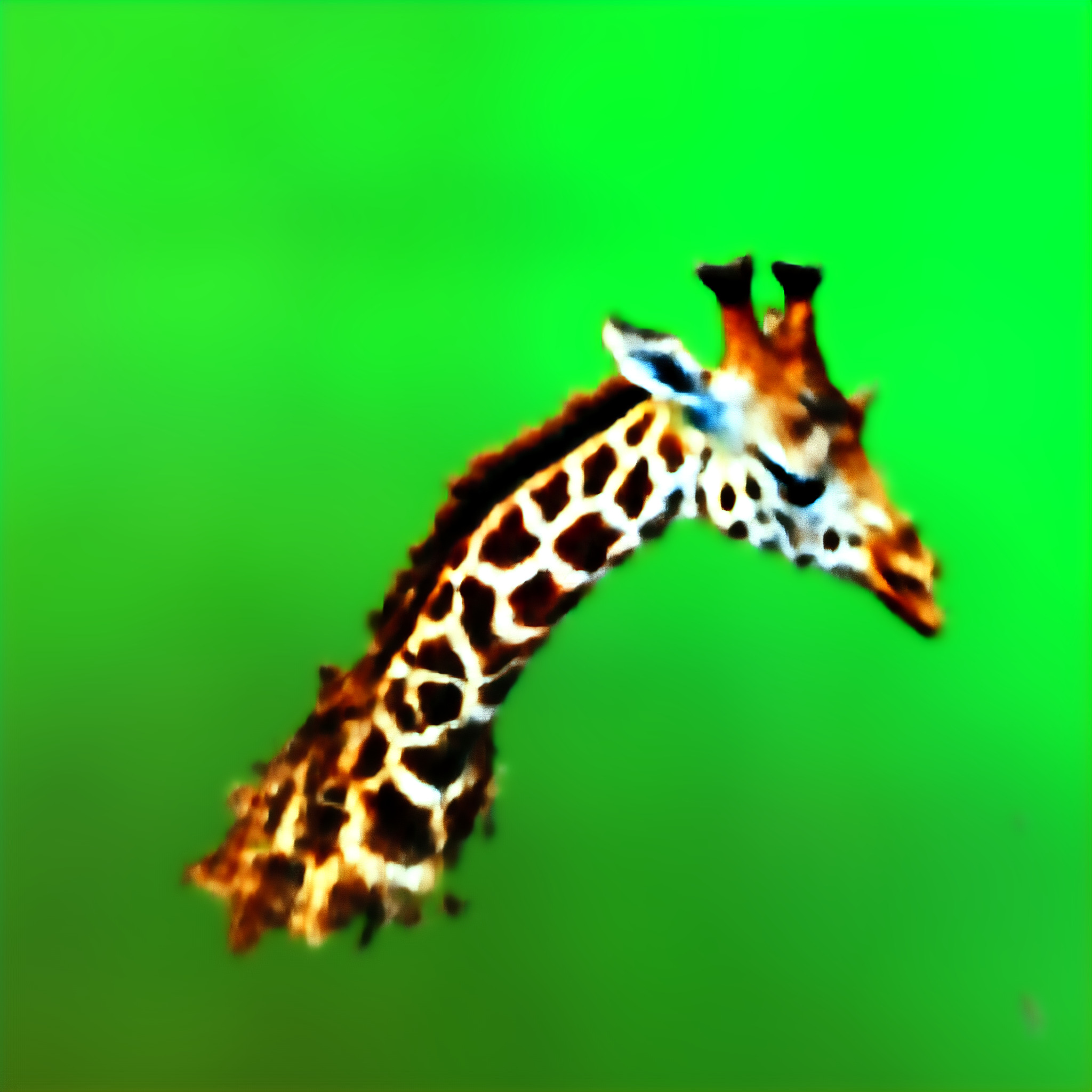} & 
        \includegraphics[width=0.19\linewidth]{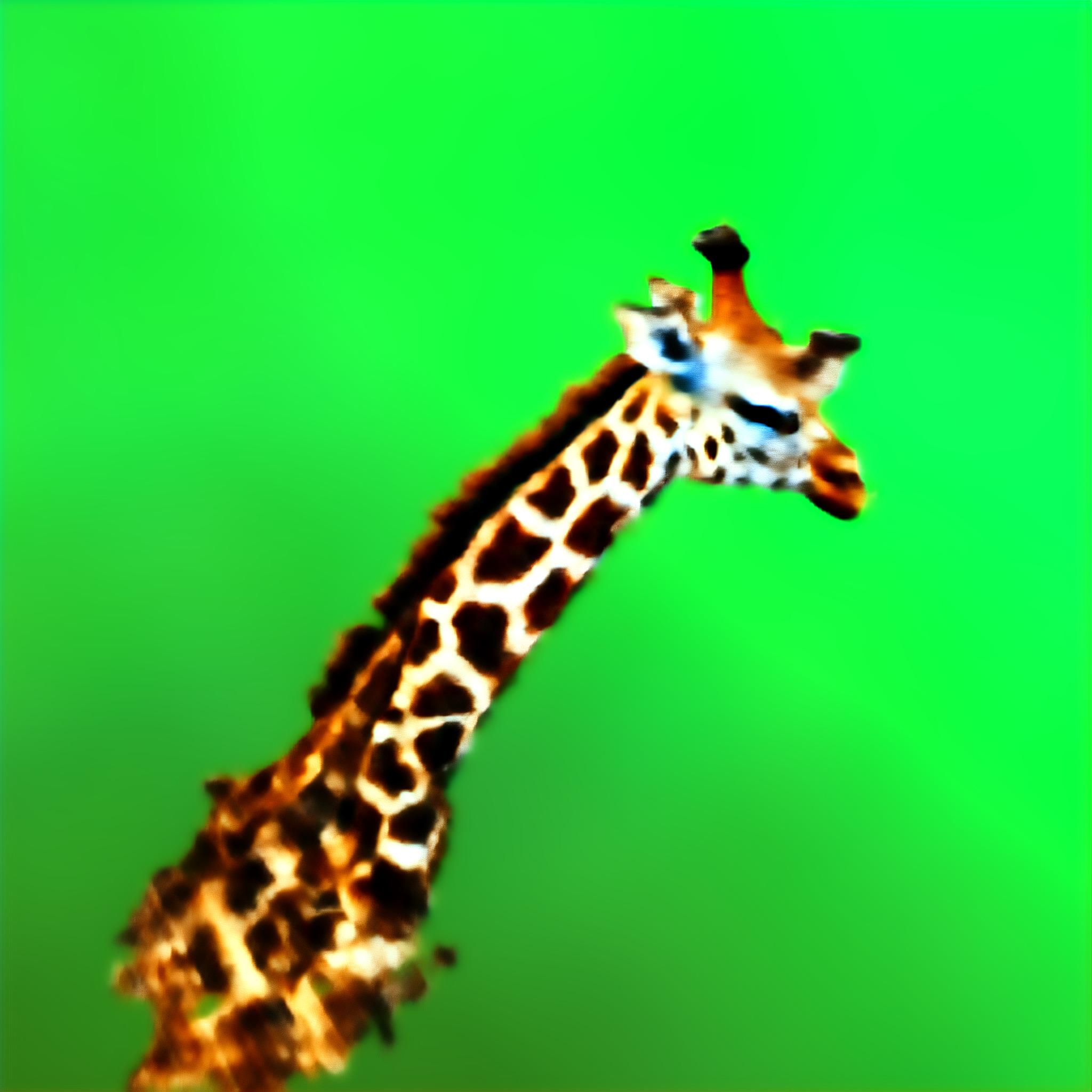} & 
        \includegraphics[width=0.19\linewidth]{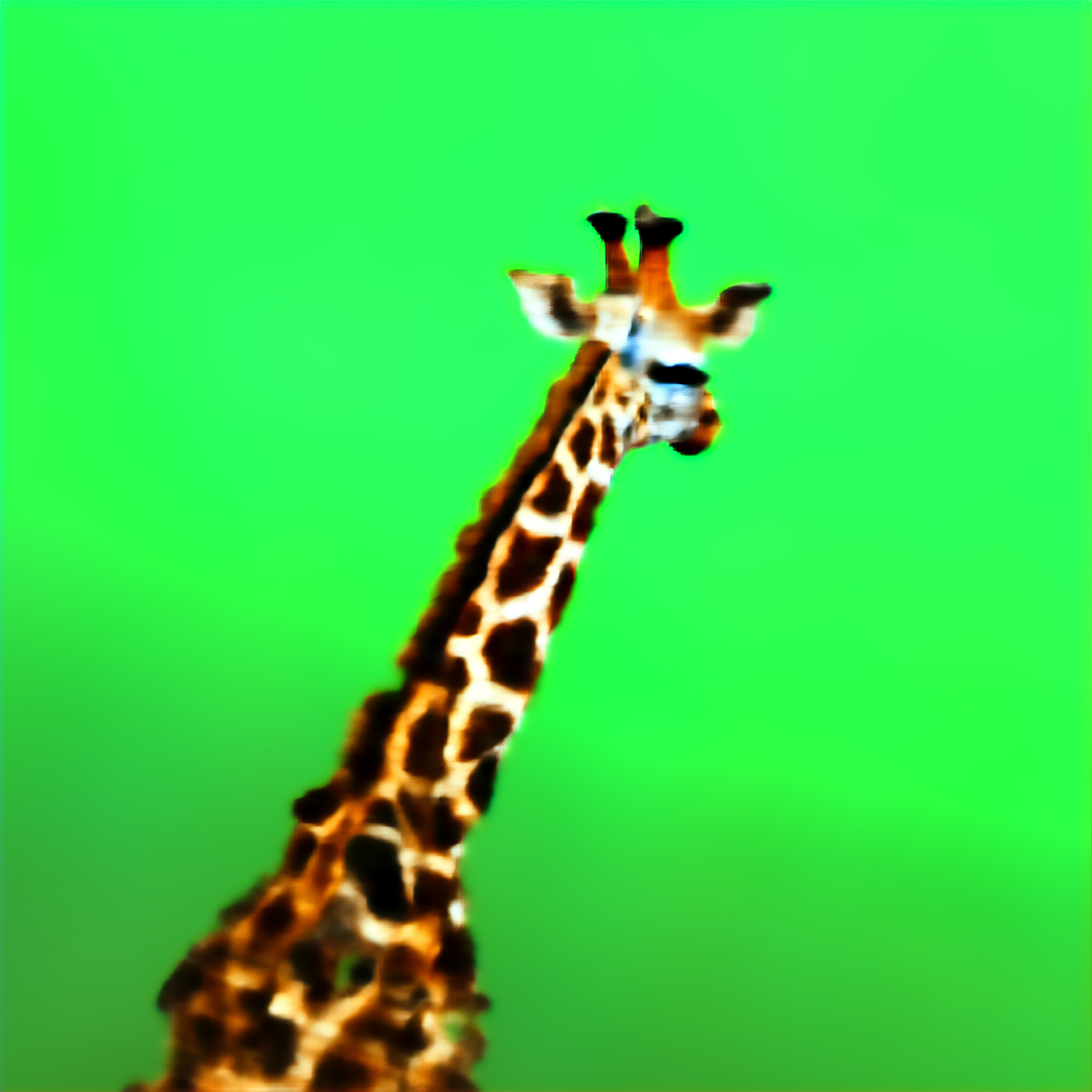} & 
        \includegraphics[width=0.19\linewidth]{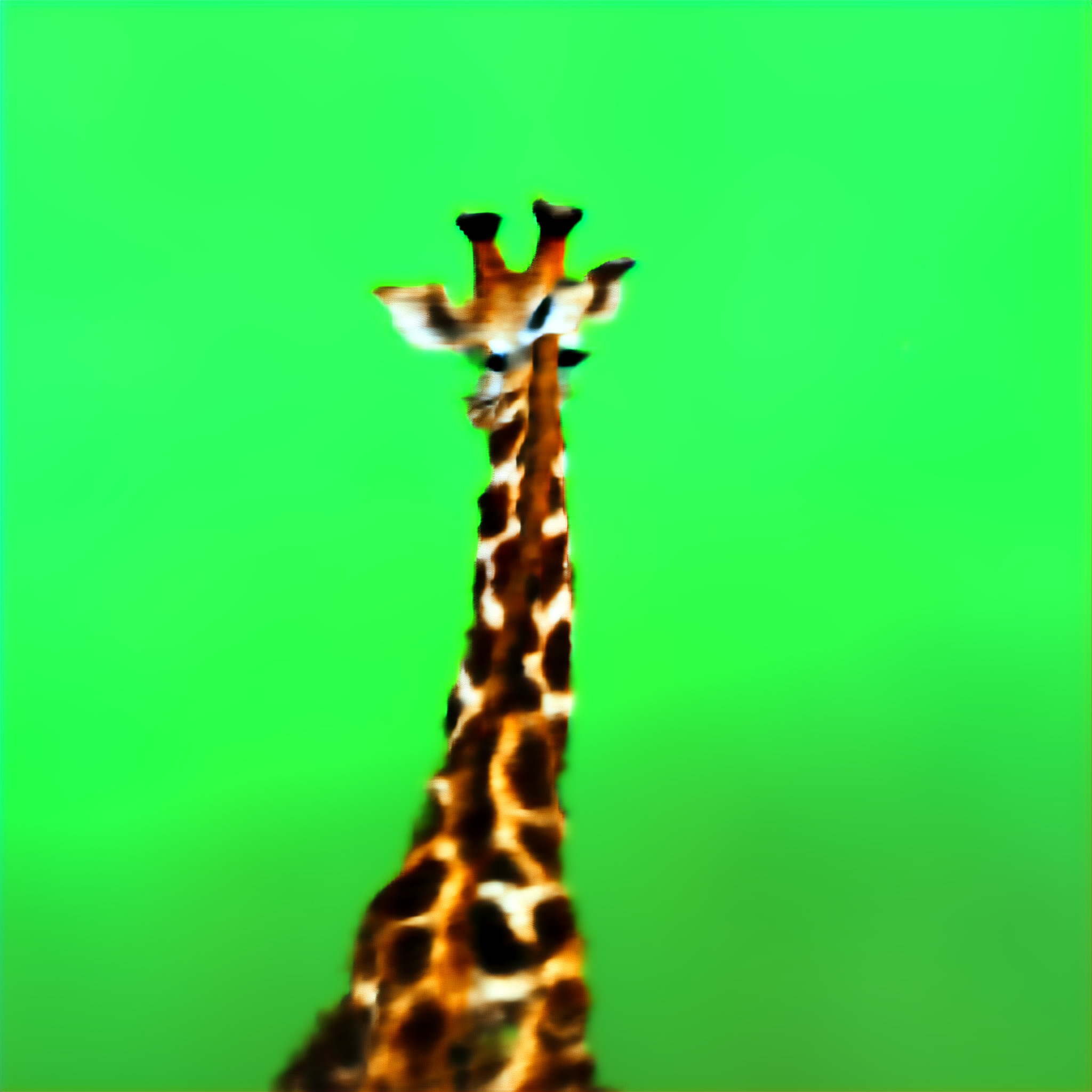} \\
        \multicolumn{5}{c}{``A photo of a giraffe''} \\
        \includegraphics[width=0.19\linewidth]{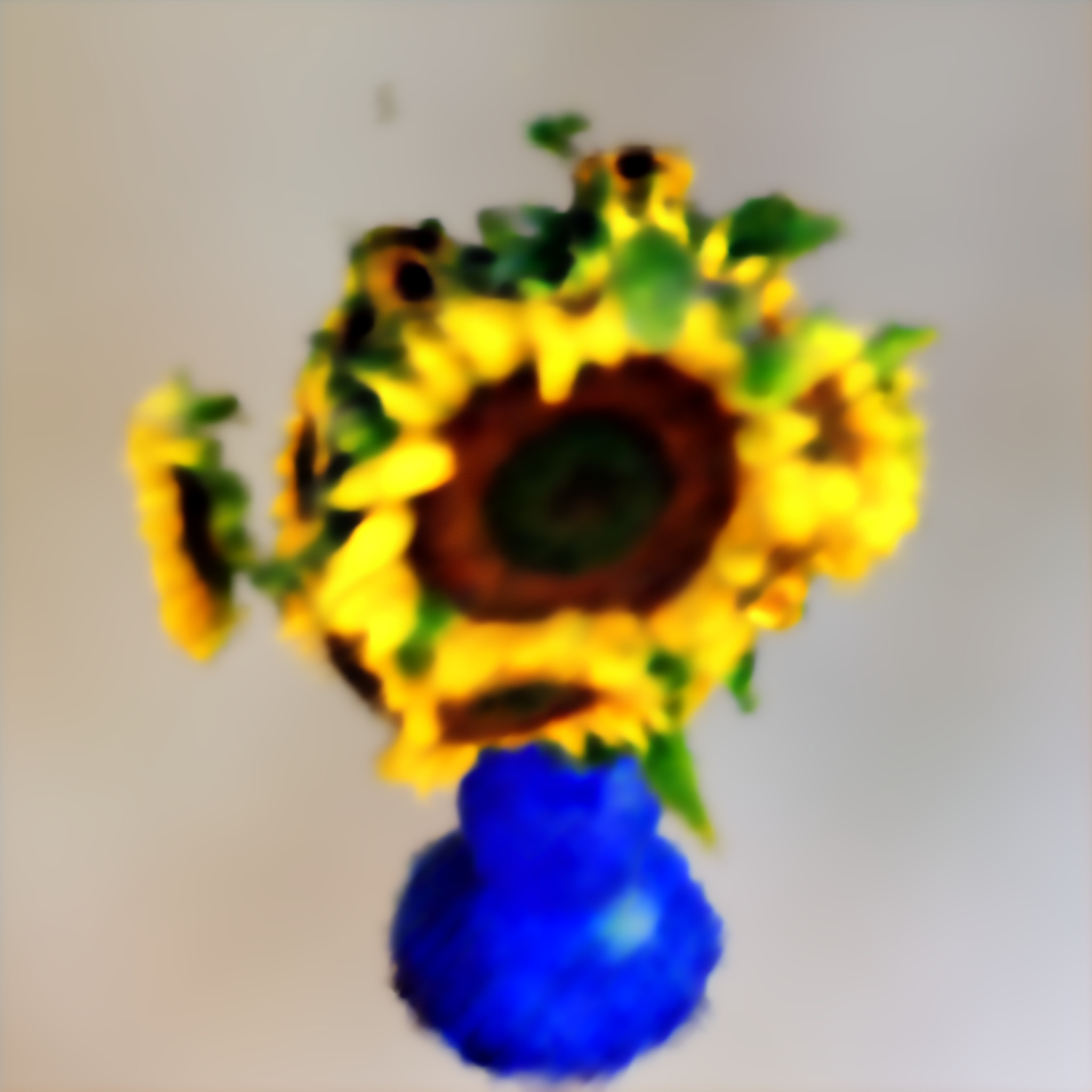} & 
        \includegraphics[width=0.19\linewidth]{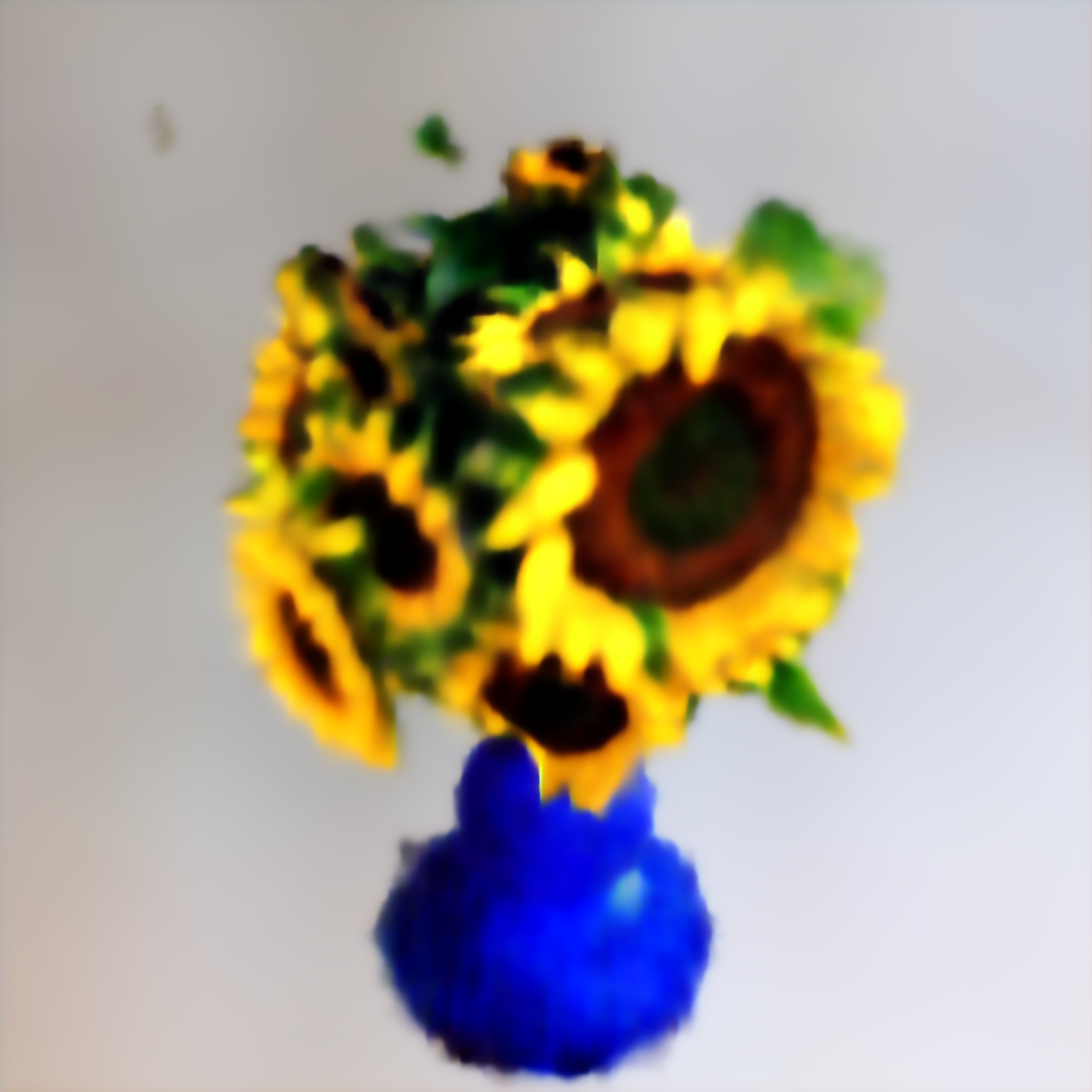} & 
        \includegraphics[width=0.19\linewidth]{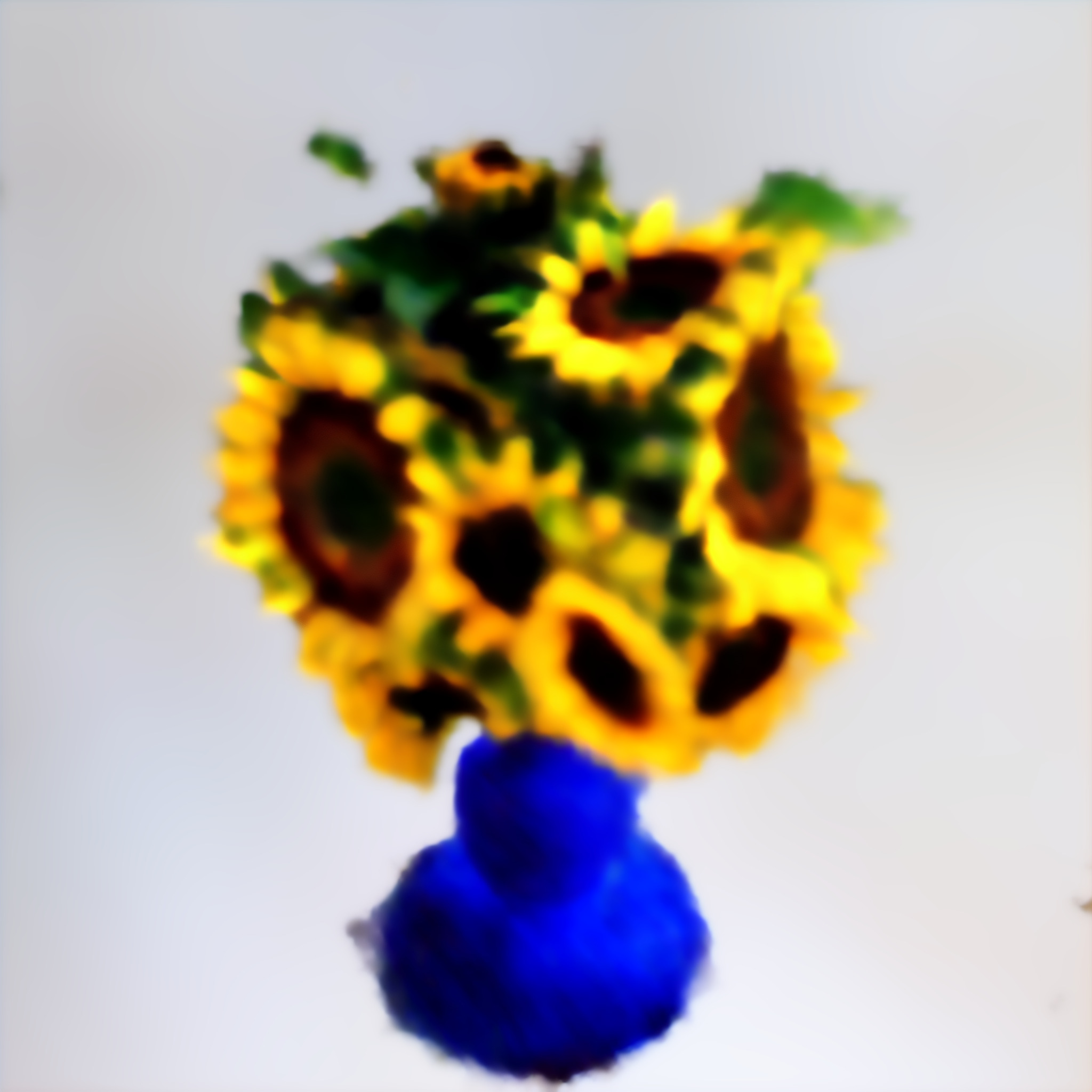} & 
        \includegraphics[width=0.19\linewidth]{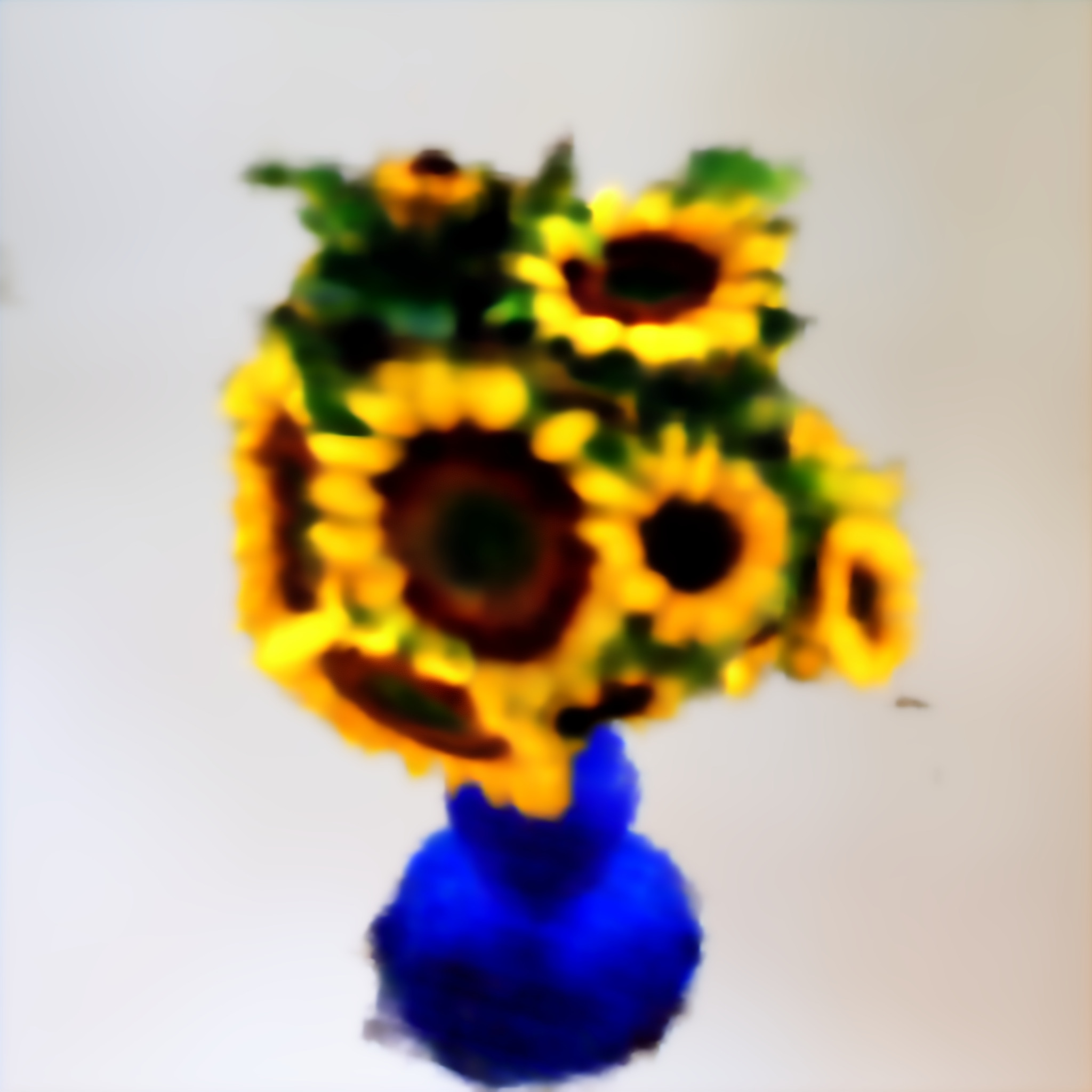} & 
        \includegraphics[width=0.19\linewidth]{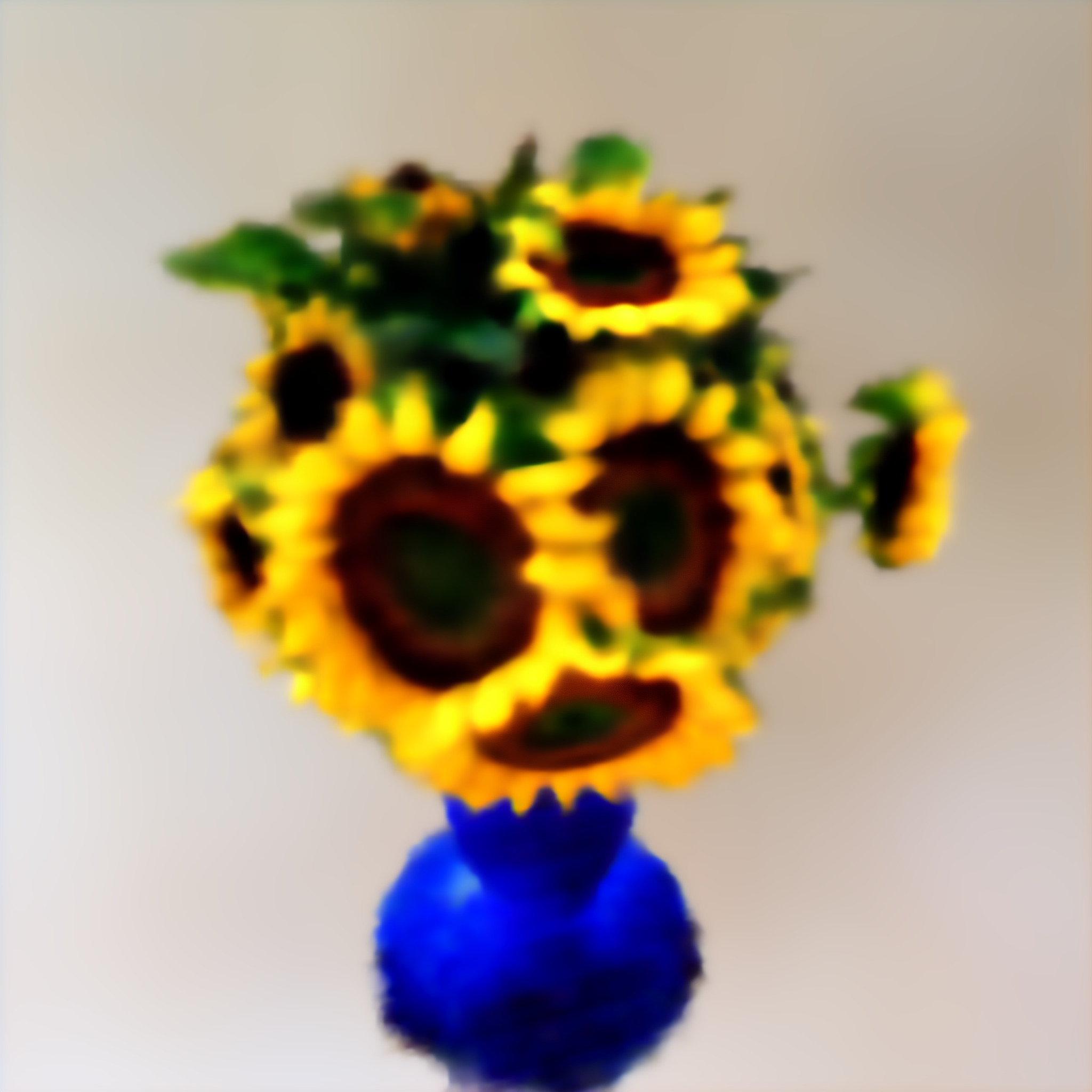} \\
        \multicolumn{5}{c}{``A photo of a vase with sunflowers''} \\
        \includegraphics[width=0.19\linewidth]{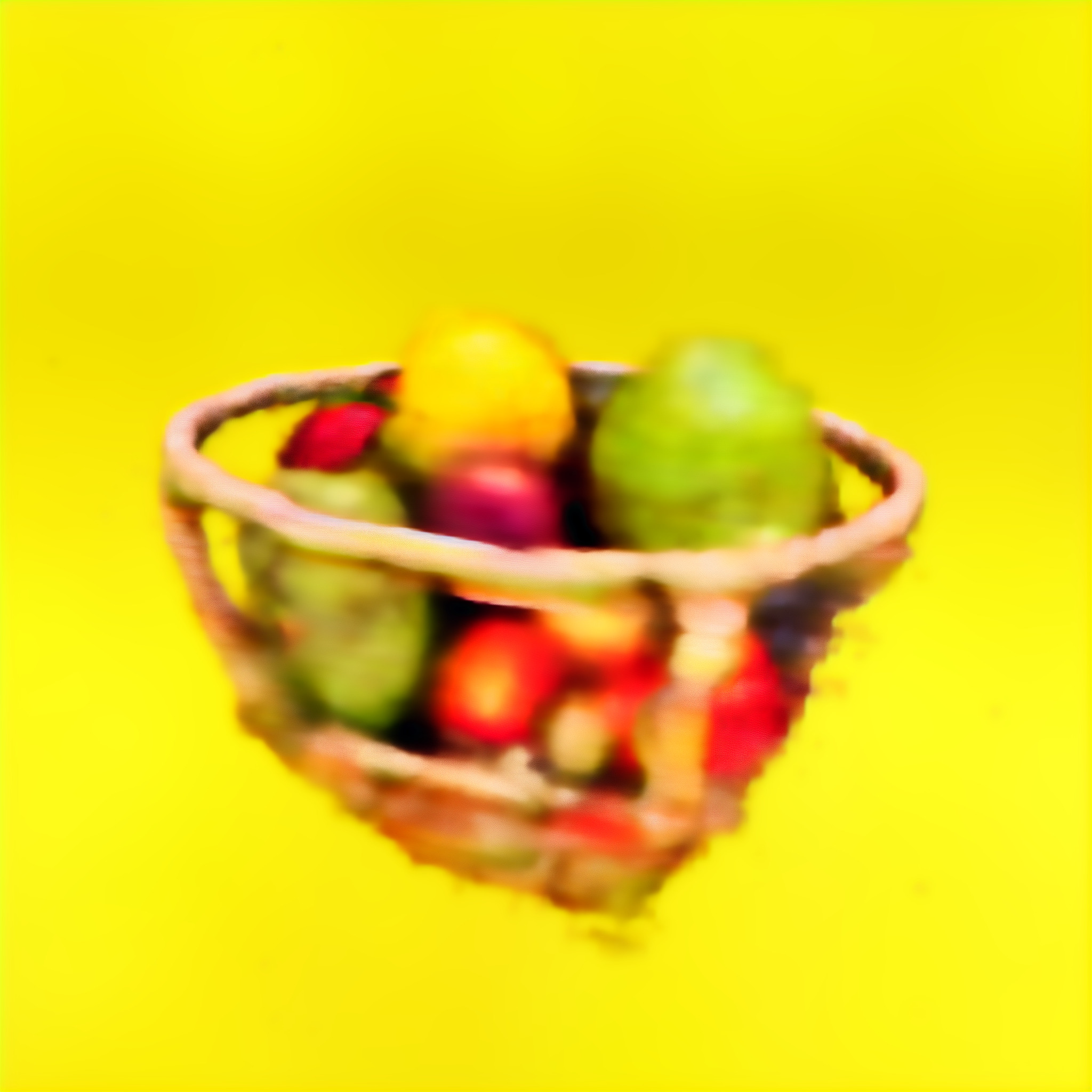} & 
        \includegraphics[width=0.19\linewidth]{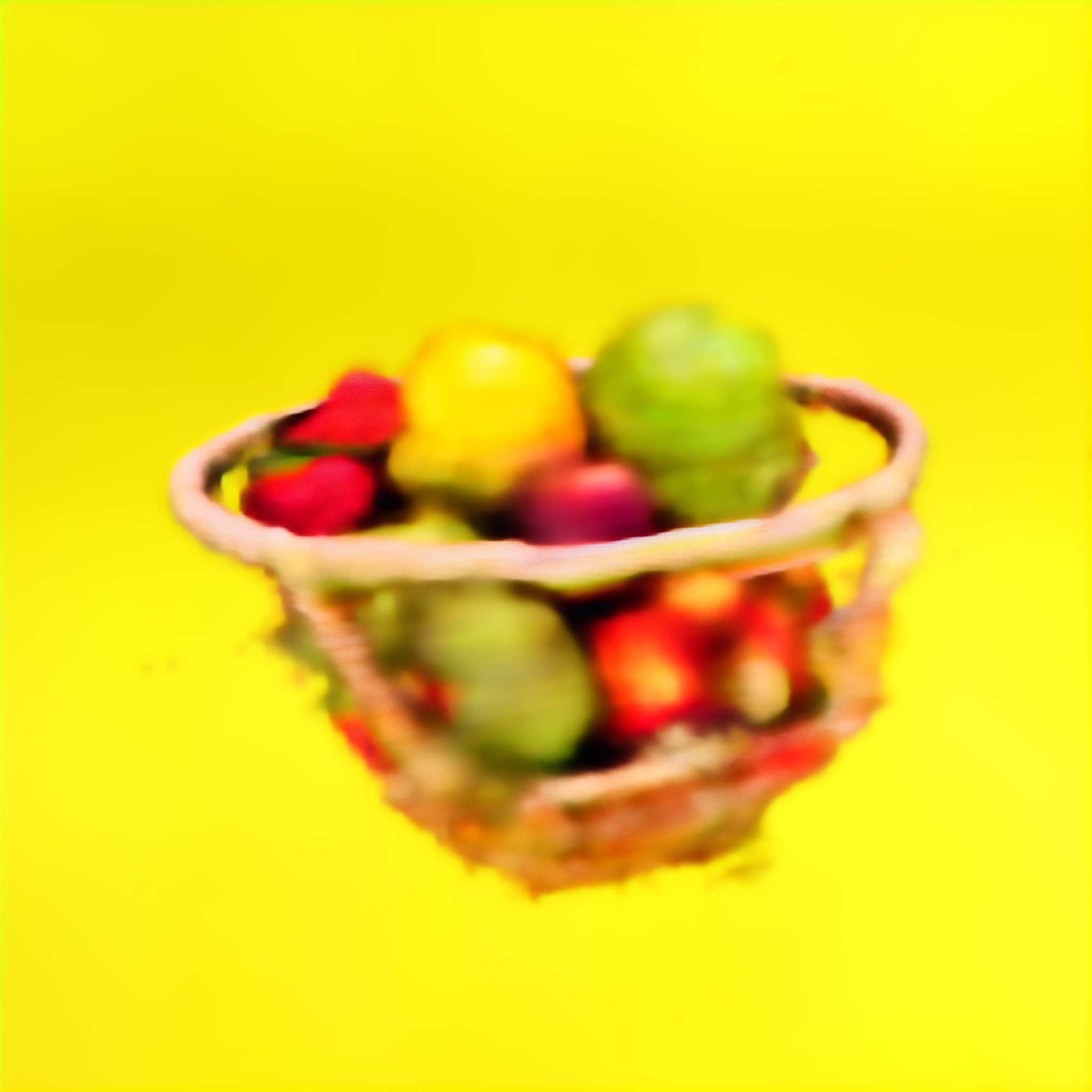} & 
        \includegraphics[width=0.19\linewidth]{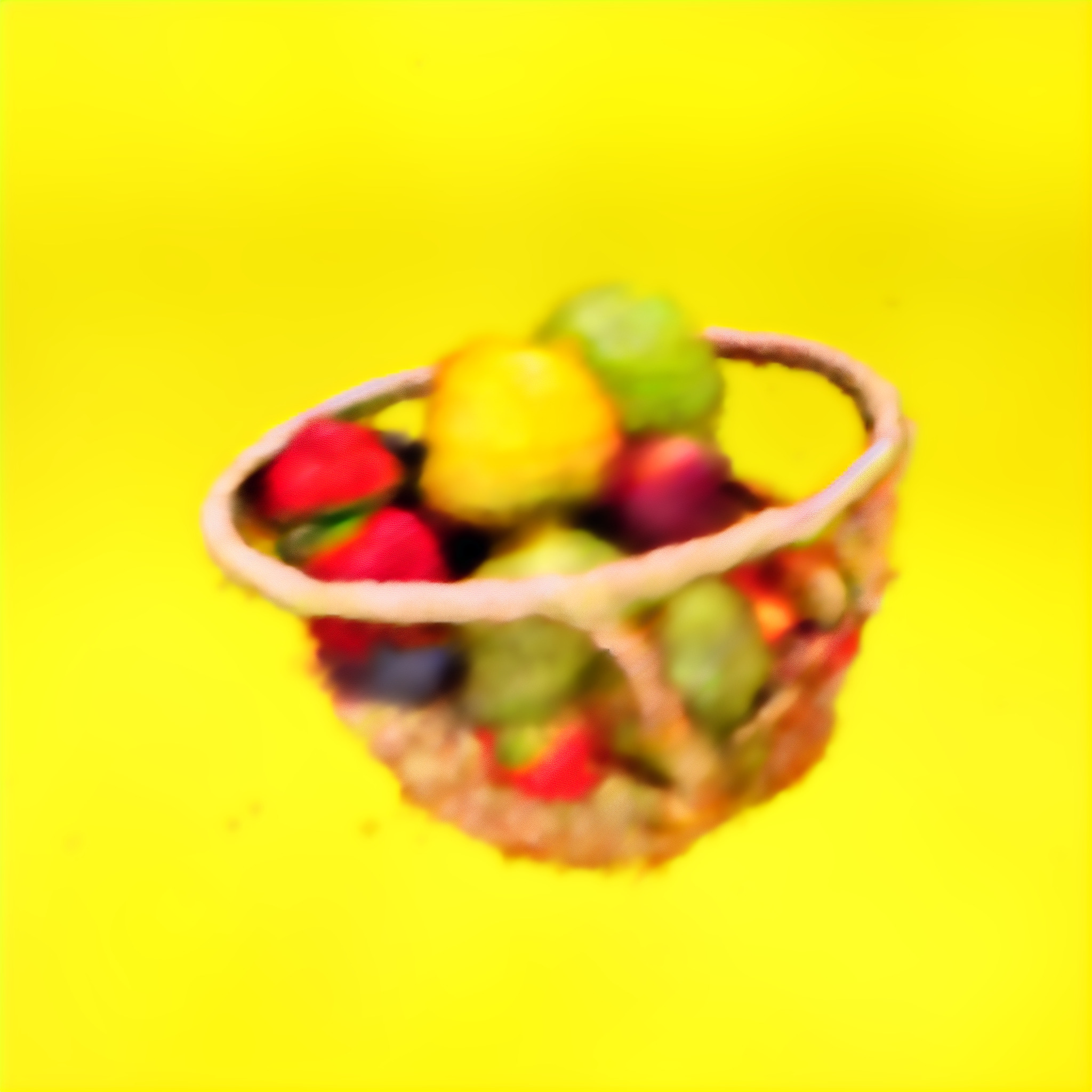} & 
        \includegraphics[width=0.19\linewidth]{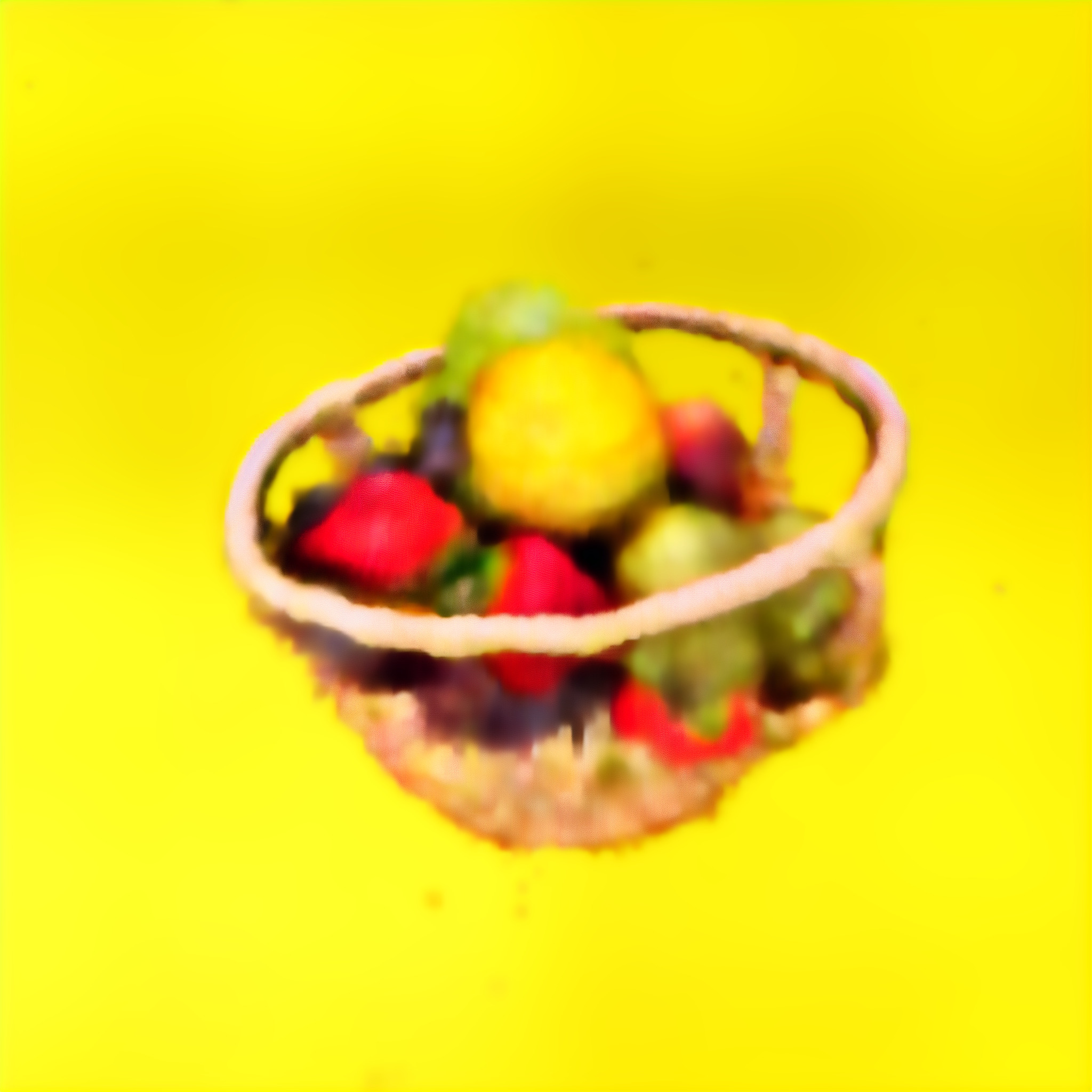} & 
        \includegraphics[width=0.19\linewidth]{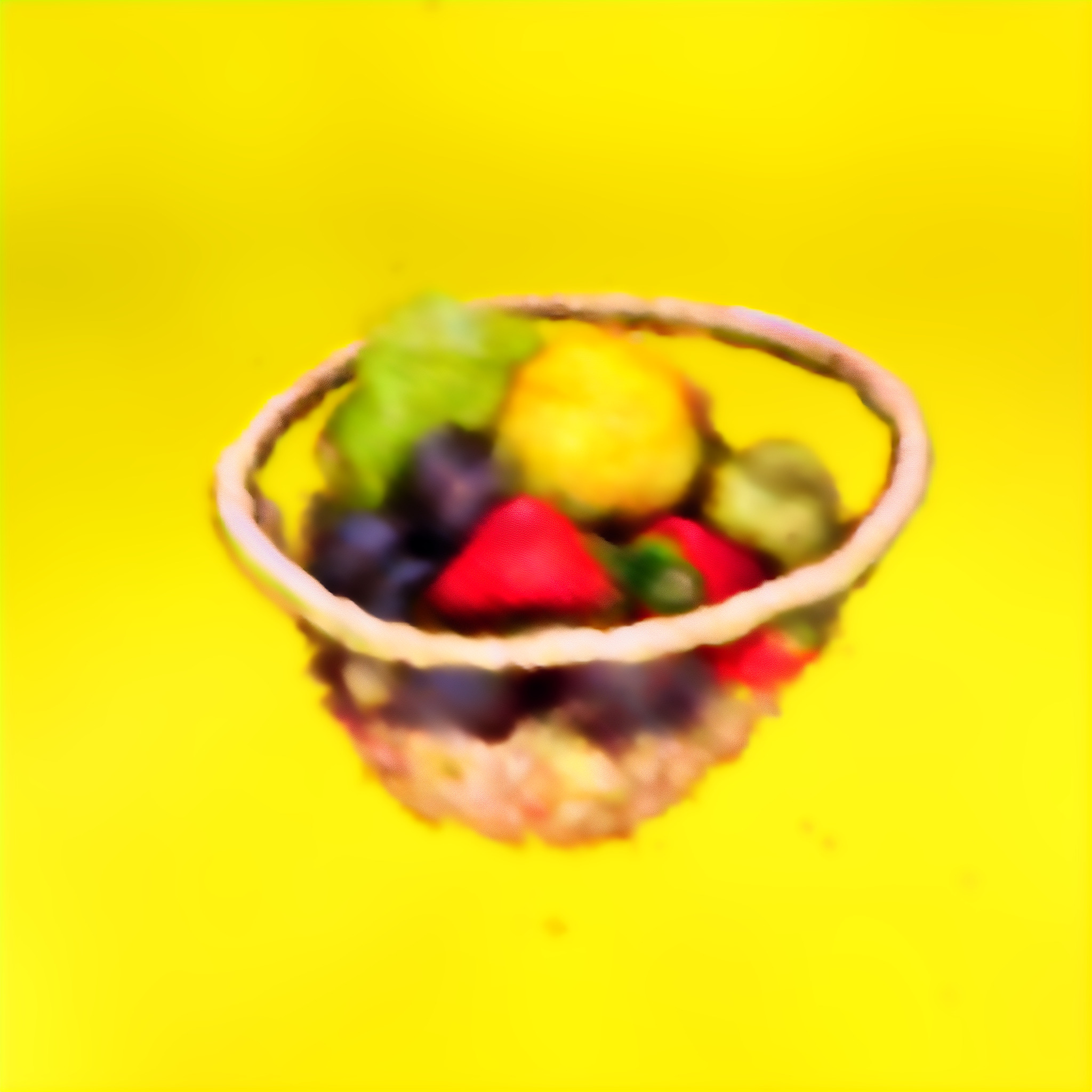}  \\
        \multicolumn{5}{c}{``A photo of a basket with fruits''}

    \end{tabular}}
    \caption{Latent-NeRF results from different viewpoints.}
    \label{fig:text2nerf_directions}
\end{figure} 

%% file: figures/4_exp/latent_nerf/text2nerf_comparison/fig.tex
\begin{figure}
    \centering
    \setlength{\tabcolsep}{1pt}
    {\scriptsize
    \begin{tabular}{c c}
        \raisebox{0.052\textwidth}{\rotatebox[origin=t]{90}{\scalebox{0.9}{DreamFields}}}& 
        \includegraphics[width=0.9\linewidth]{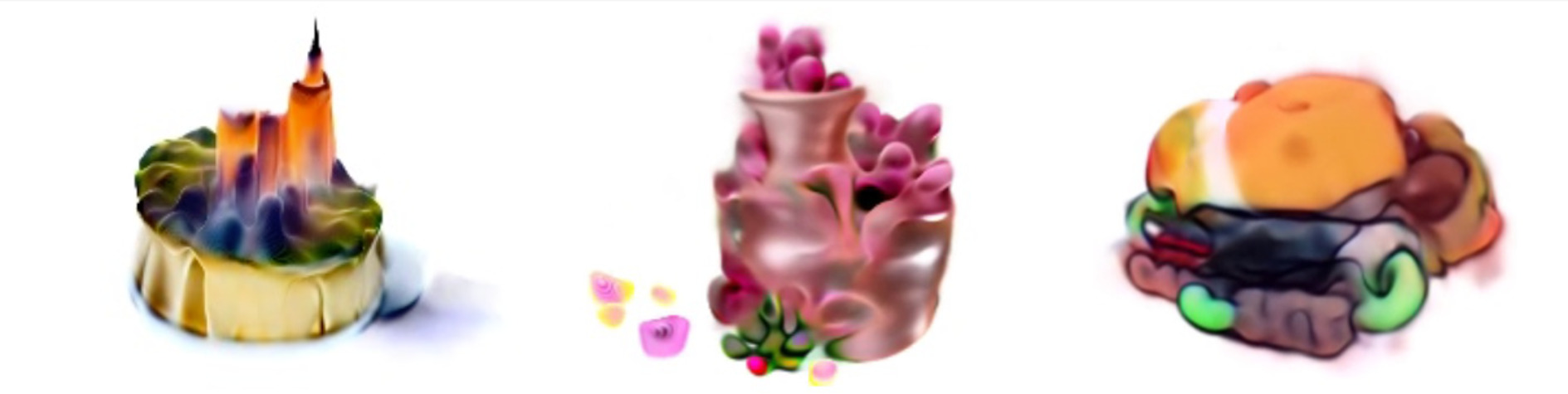} \\
        \raisebox{0.052\textwidth}{\rotatebox[origin=t]{90}{\scalebox{0.9}{DreamFields reimpl.}}} & 
        \includegraphics[width=0.9\linewidth]{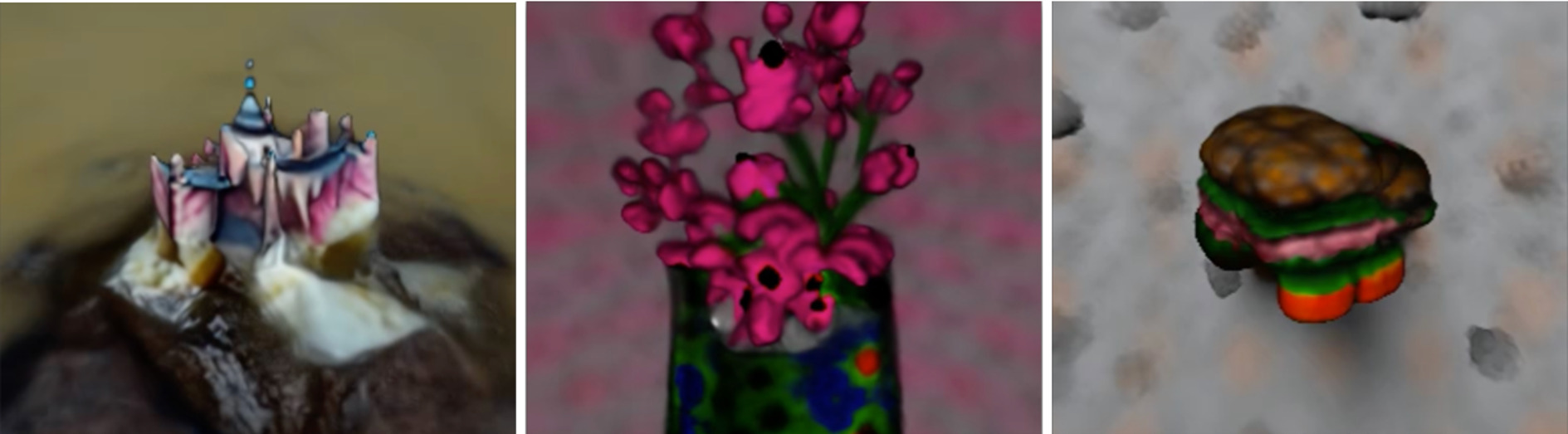} \\
        \raisebox{0.052\textwidth}{\rotatebox[origin=t]{90}{\scalebox{0.9}{CLIPMesh}}} & 
        \includegraphics[width=0.9\linewidth]{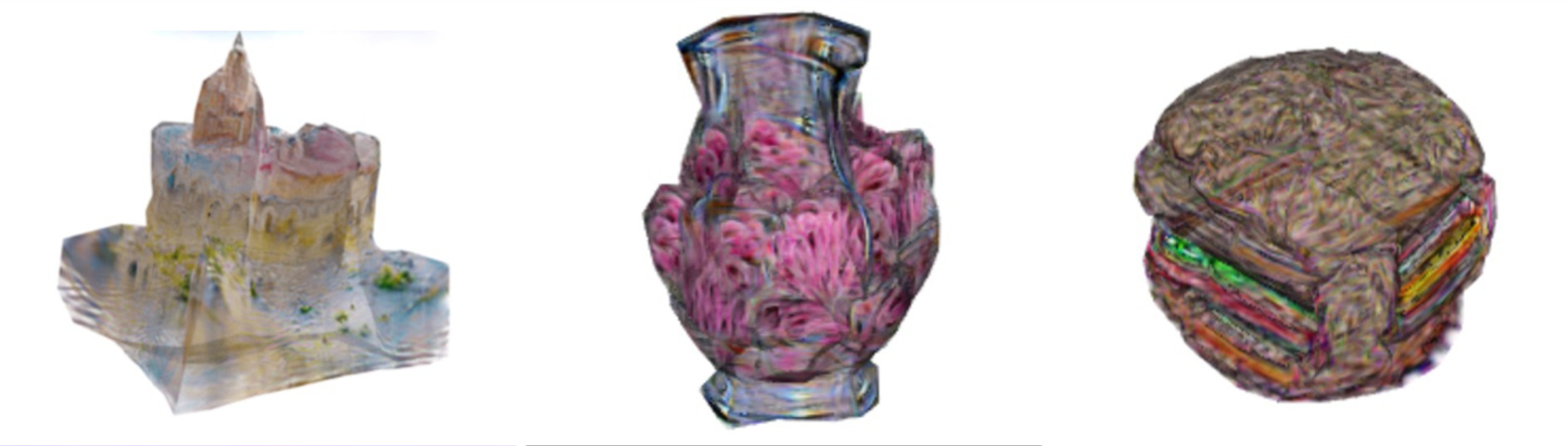} \\
        \raisebox{0.052\textwidth}{\rotatebox[origin=t]{90}{\scalebox{0.9}{DreamFusion}}} & 
        \includegraphics[width=0.9\linewidth]{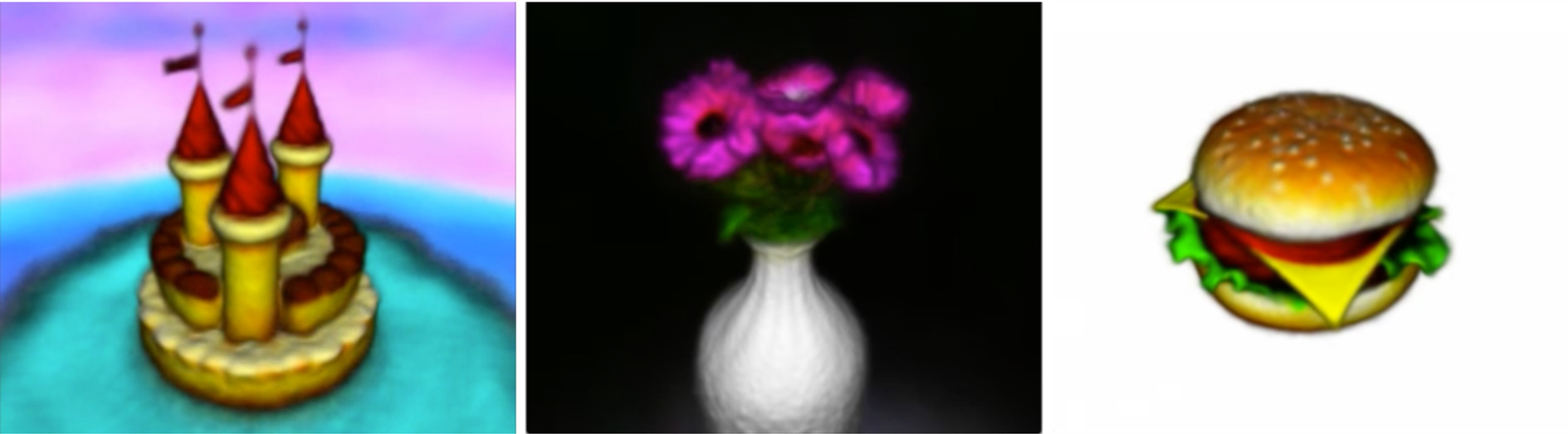} \\
        \raisebox{0.051\textwidth}{\rotatebox[origin=t]{90}{\scalebox{0.9}{Latent-NeRF (Ours)}}} & 
        \includegraphics[width=0.9\linewidth]{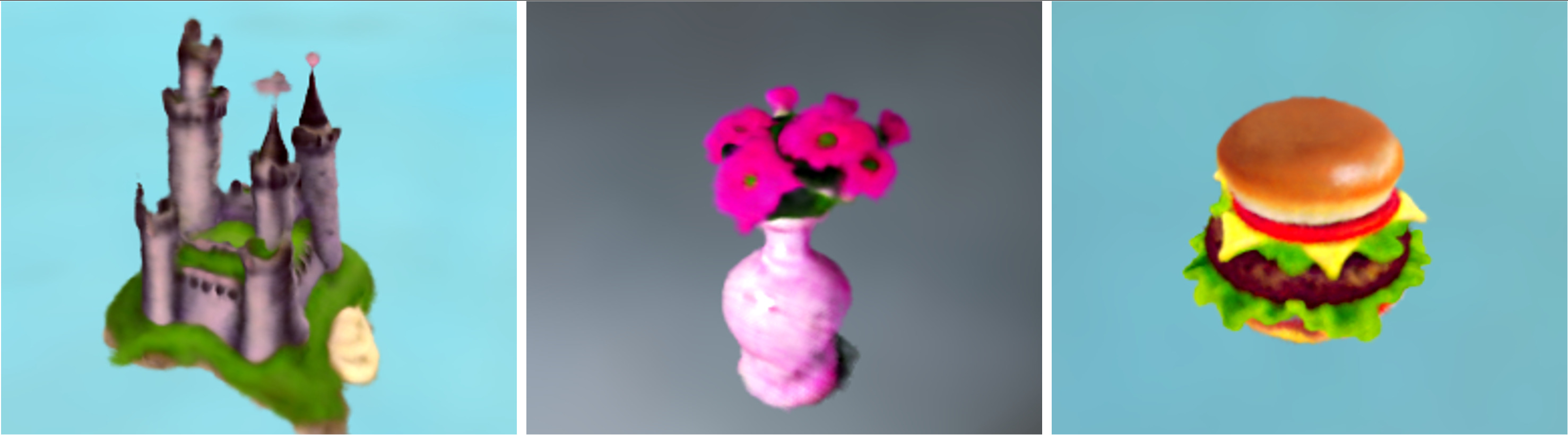} \\
        & 
        \begin{tabular*}{0.9\linewidth}{P{0.28\linewidth}P{0.31\linewidth}P{0.28\linewidth}@{}}
        \centering
        ``a Matte painting of a castle made of cheesecake surrounded by a moat made of ice cream''&
        ``A vase with pink flowers'' & ``A hamburger''
        \end{tabular*}

    \end{tabular}
    }
    
    \caption{
    Qualitative comparison with other text-to-3D methods.
    }
    \label{fig:text2nerf_comparison}
\end{figure}

%% file: figures/4_exp/rgb_finetune/fig.tex
\begin{figure}
    \centering
    \setlength{\tabcolsep}{1pt}
    {\scriptsize
    \begin{tabular}{c c c c c}
        \raisebox{0.052\textwidth}{\rotatebox[origin=t]{90}{\scalebox{0.9}{Latent}}} & 
        \includegraphics[width=0.24\linewidth]{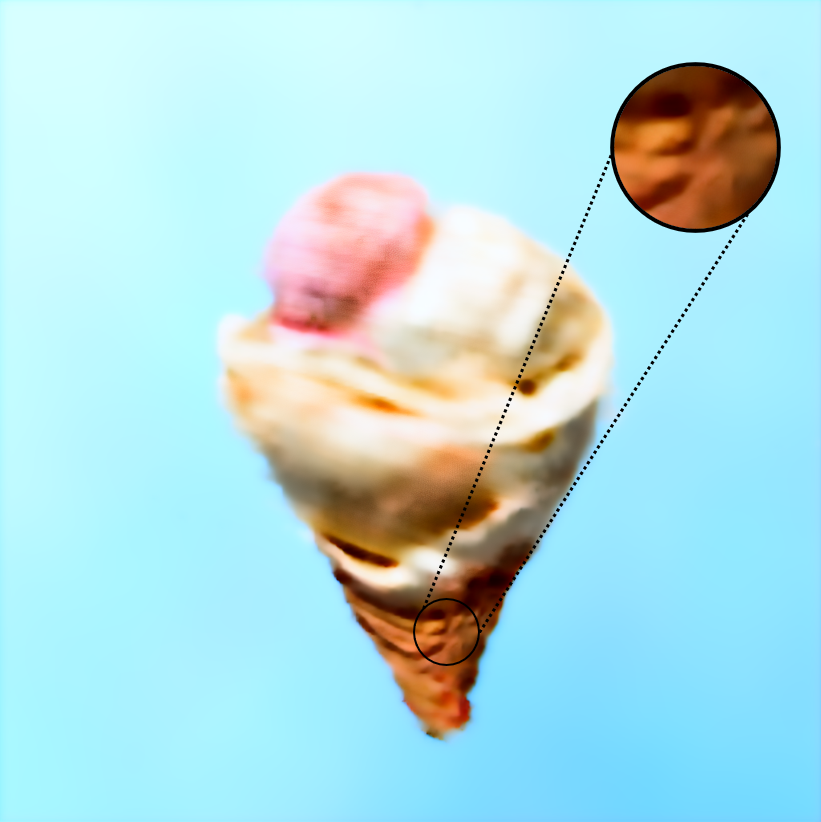} & 
        \includegraphics[width=0.24\linewidth]{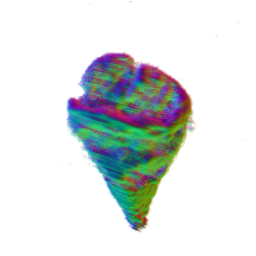} & 
        \includegraphics[width=0.24\linewidth]{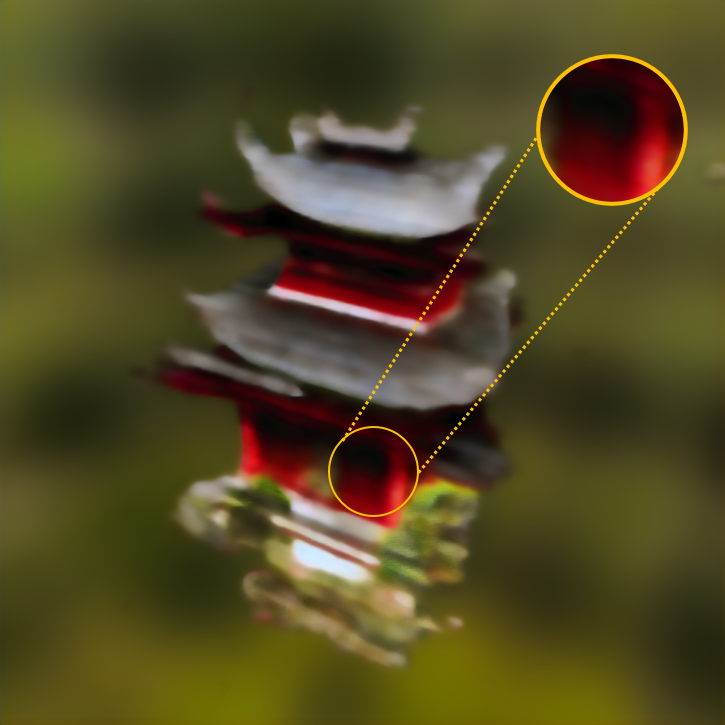} & 
        \includegraphics[width=0.24\linewidth]{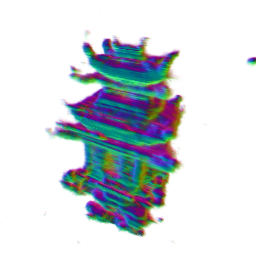} \\
        \raisebox{0.048\textwidth}{\rotatebox[origin=t]{90}{\scalebox{0.9}{RGB Refinement}}} & 
        \includegraphics[width=0.24\linewidth]{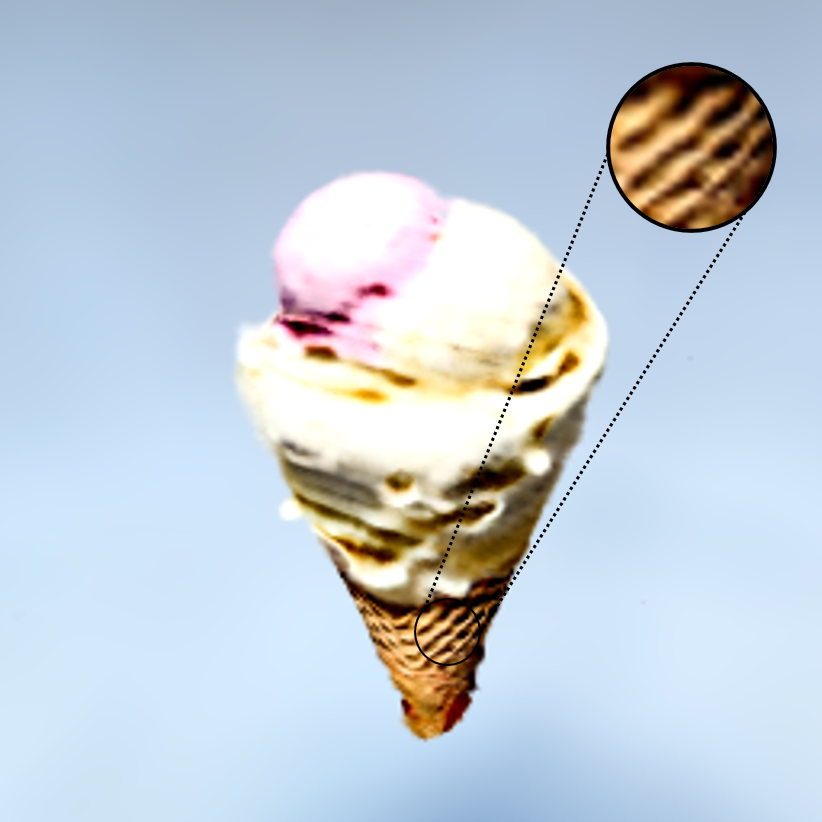} & 
        \includegraphics[width=0.24\linewidth]{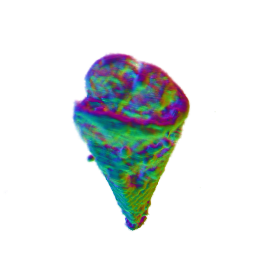} & 
        \includegraphics[width=0.24\linewidth]{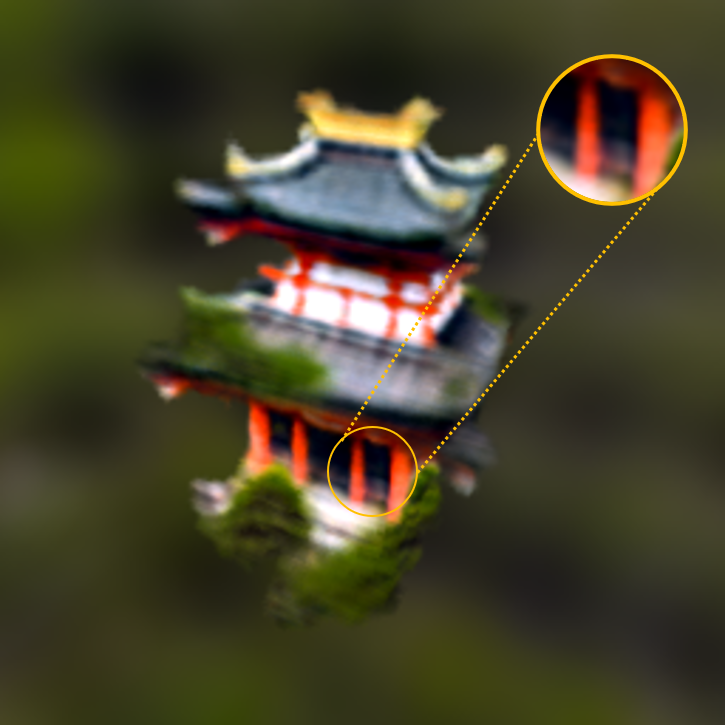} & 
        \includegraphics[width=0.24\linewidth]{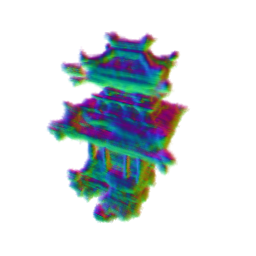}
        \\
        \raisebox{0.052\textwidth}{\rotatebox[origin=t]{90}{\scalebox{0.9}{Latent}}} & 
        \includegraphics[width=0.24\linewidth]{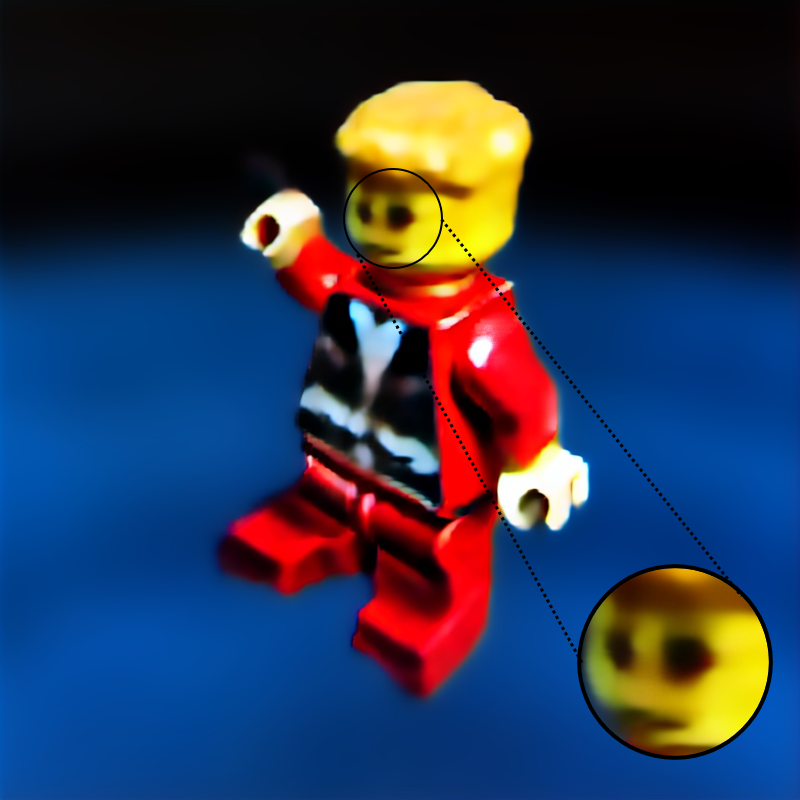} & 
        \includegraphics[width=0.24\linewidth]{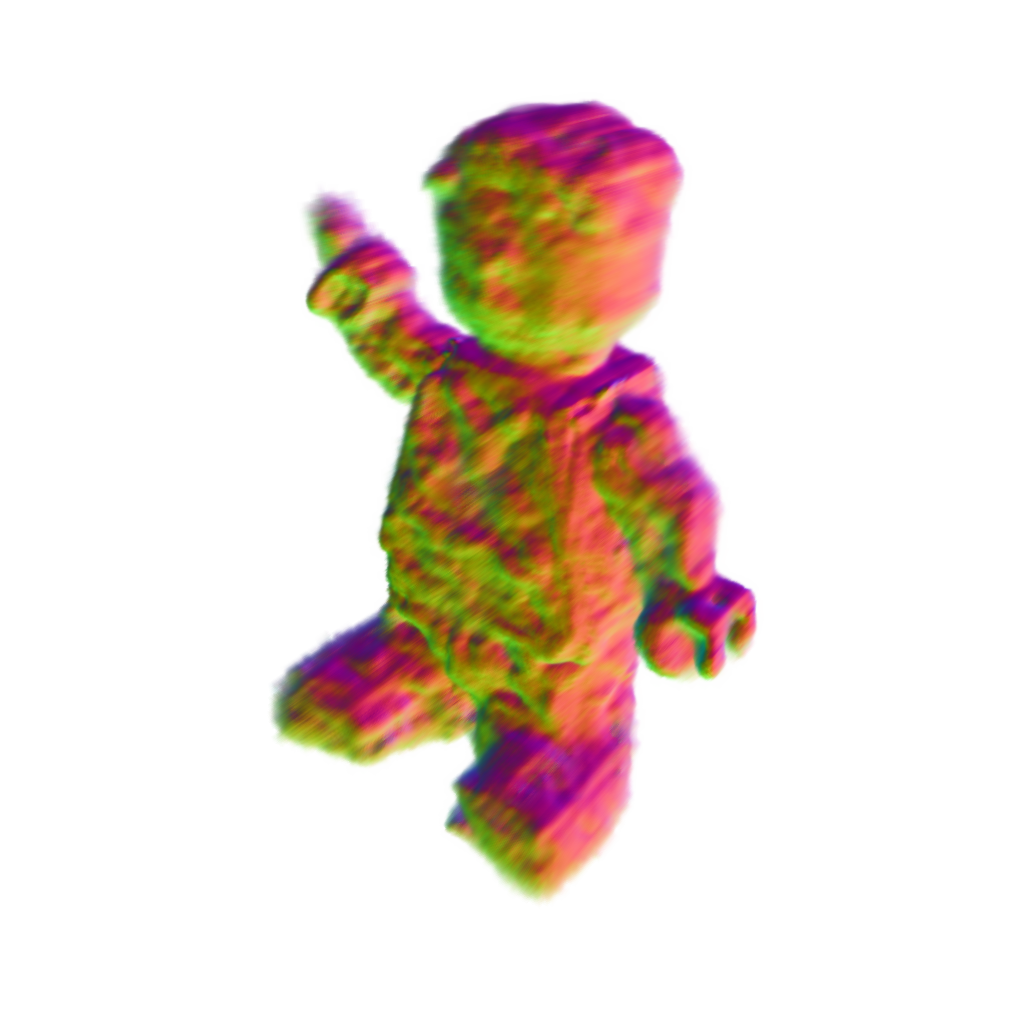} & 
        \includegraphics[width=0.24\linewidth]{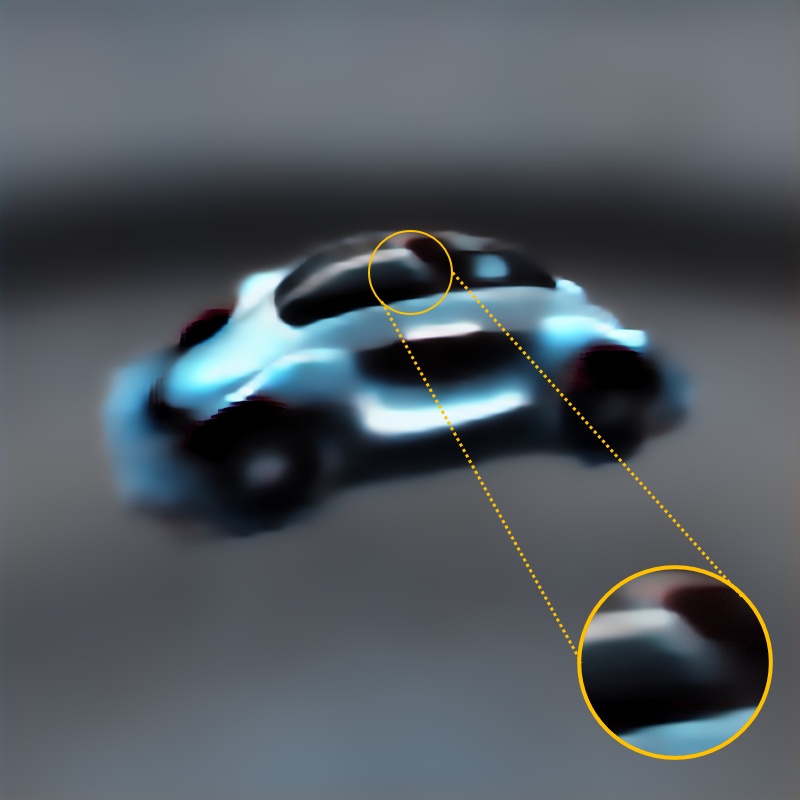} & 
        \includegraphics[width=0.24\linewidth]{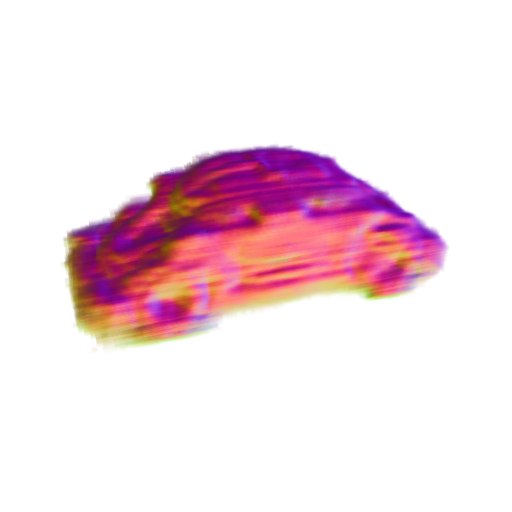} \\
        \raisebox{0.048\textwidth}{\rotatebox[origin=t]{90}{\scalebox{0.9}{RGB Refinement}}} & 
        \includegraphics[width=0.24\linewidth]{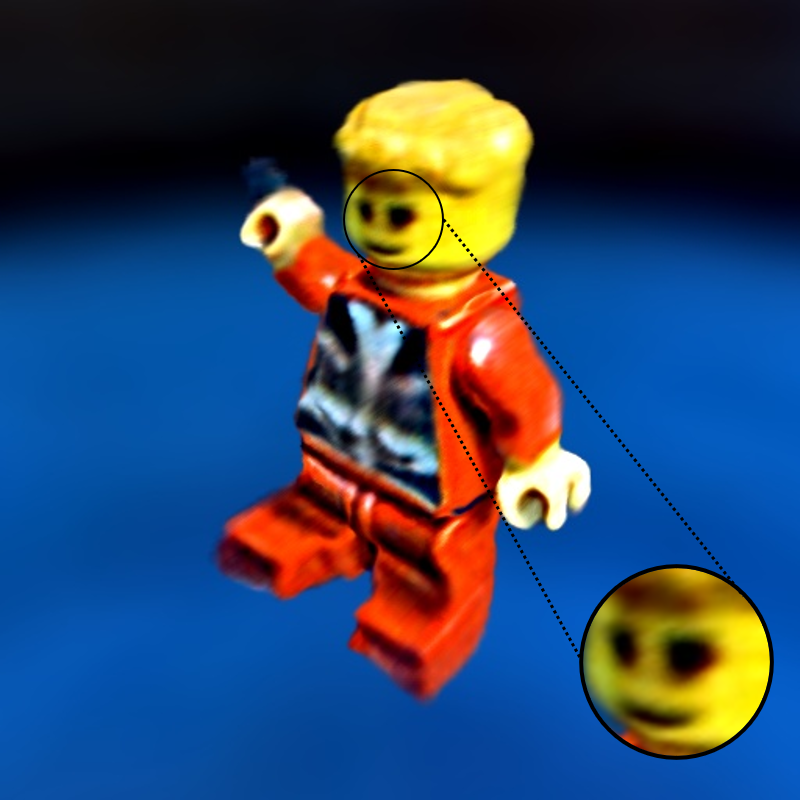} & 
        \includegraphics[width=0.24\linewidth]{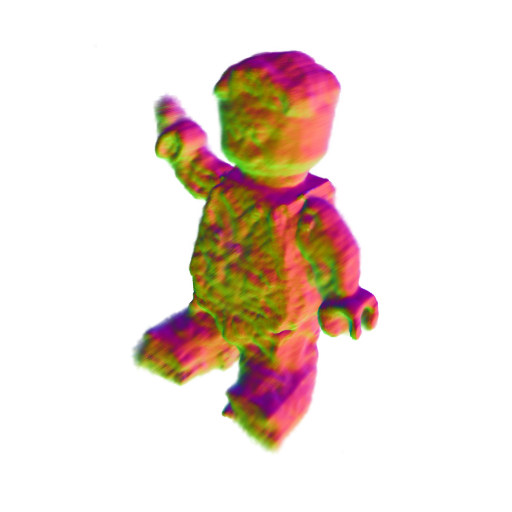} & 
        \includegraphics[width=0.24\linewidth]{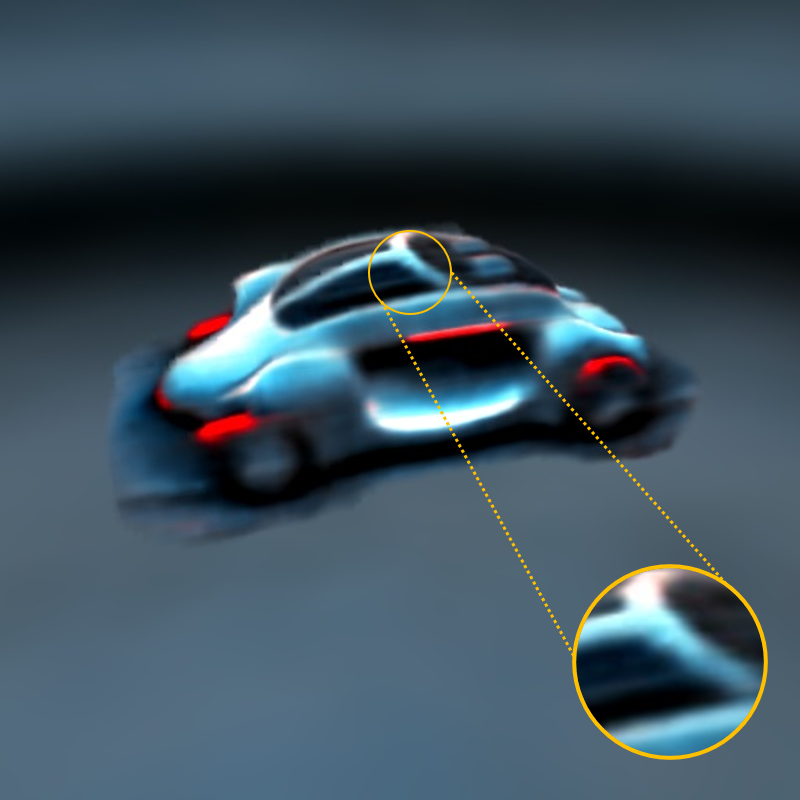} & 
        \includegraphics[width=0.24\linewidth]{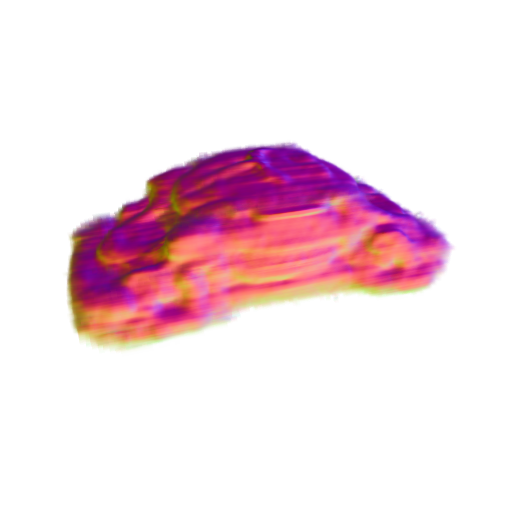} \\
        
    \end{tabular}}
    \caption{RGB refinement combined with (i) Latent-NeRF (ice-cream and temple), and (ii) \meshsketch \  (lego and car). For each shape we also show the normals direction.}
    \label{fig:rgb_finetune}
\end{figure} 

%% file: figures/4_exp/textual_inversion/cat.tex
\begin{figure}
    \centering
    \setlength{\tabcolsep}{0.5pt}
    {\scriptsize

  \begin{tabular}{ c c }
    \raisebox{-.5\height}{
    \includegraphics[width=0.24\linewidth]{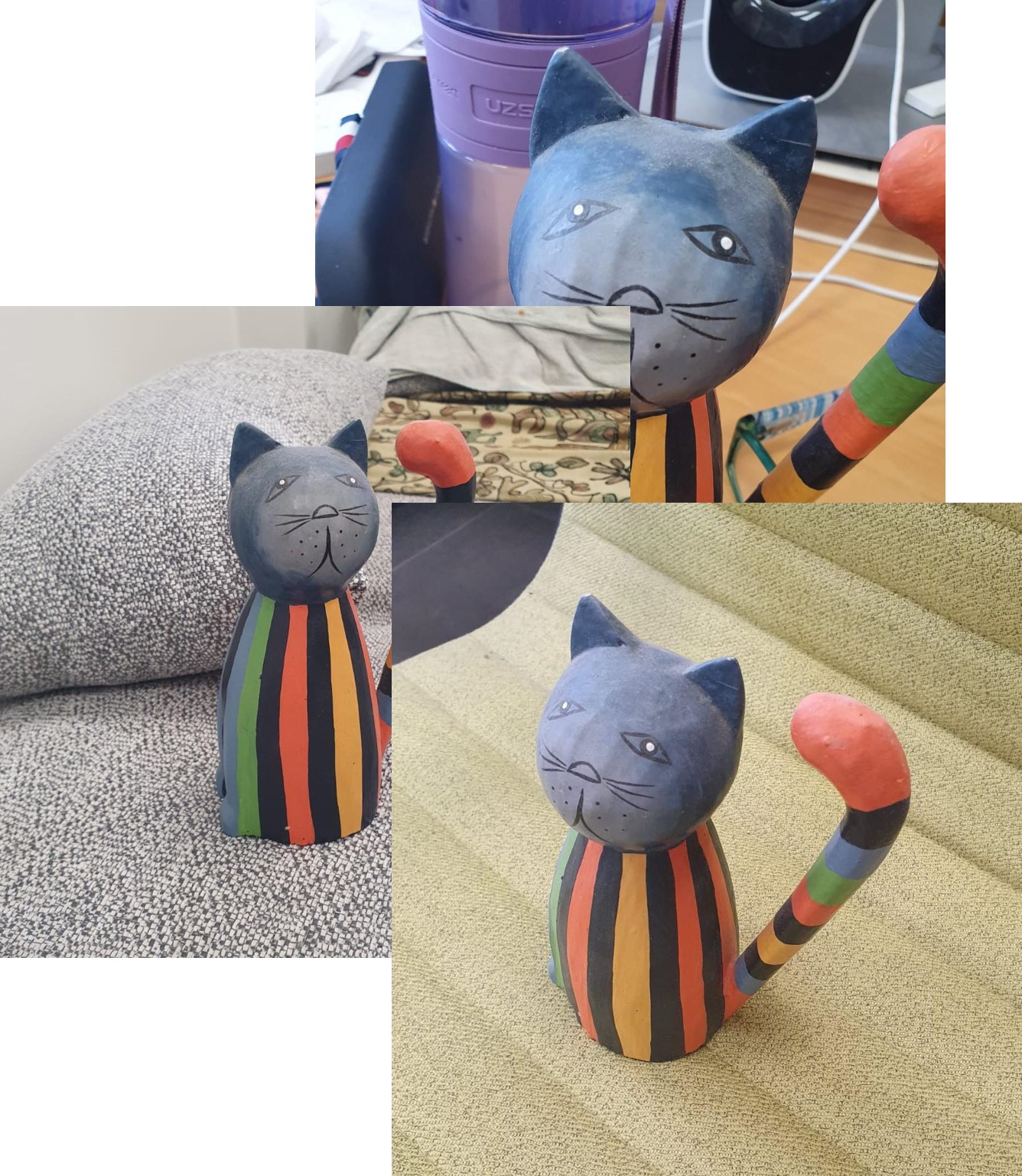}} \quad & %
    \begin{tabular}{ c c c }
    \includegraphics[width=0.24\linewidth]{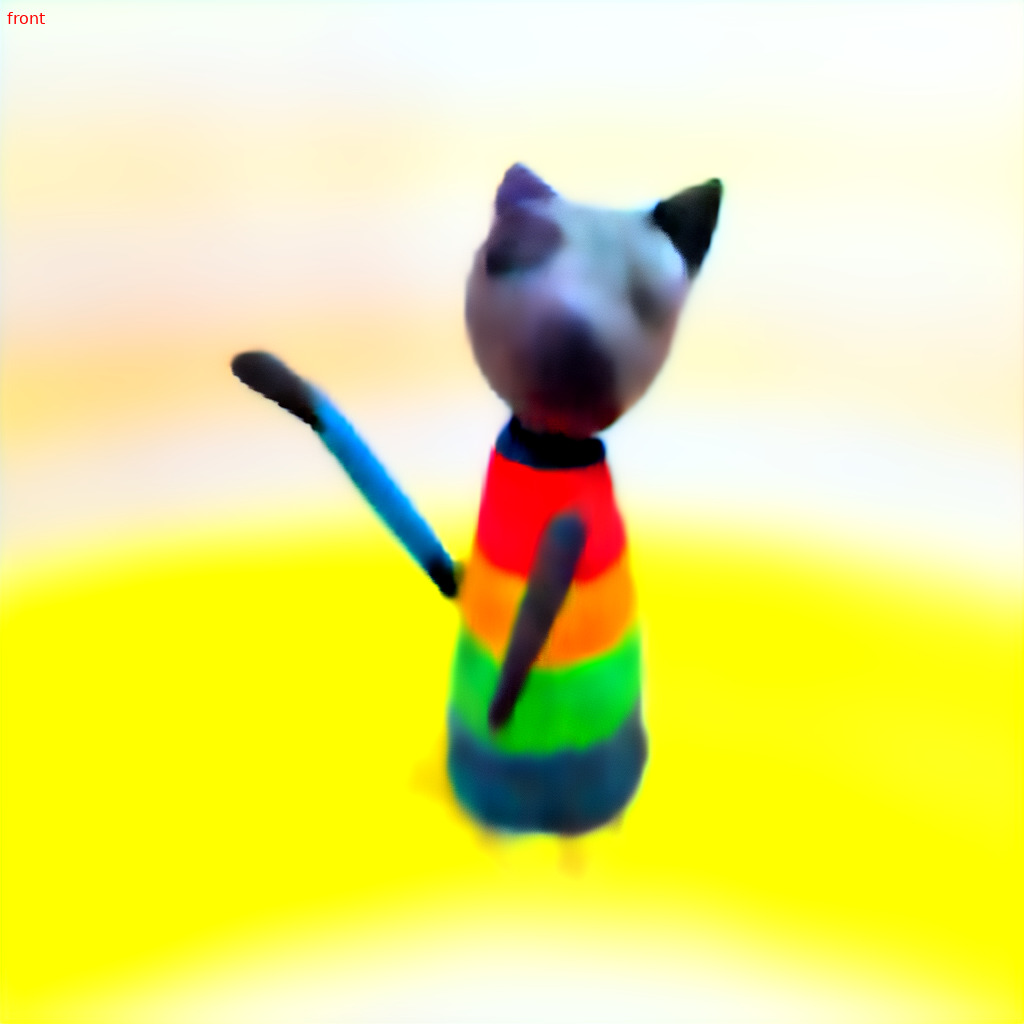} & 
    \includegraphics[width=0.24\linewidth]{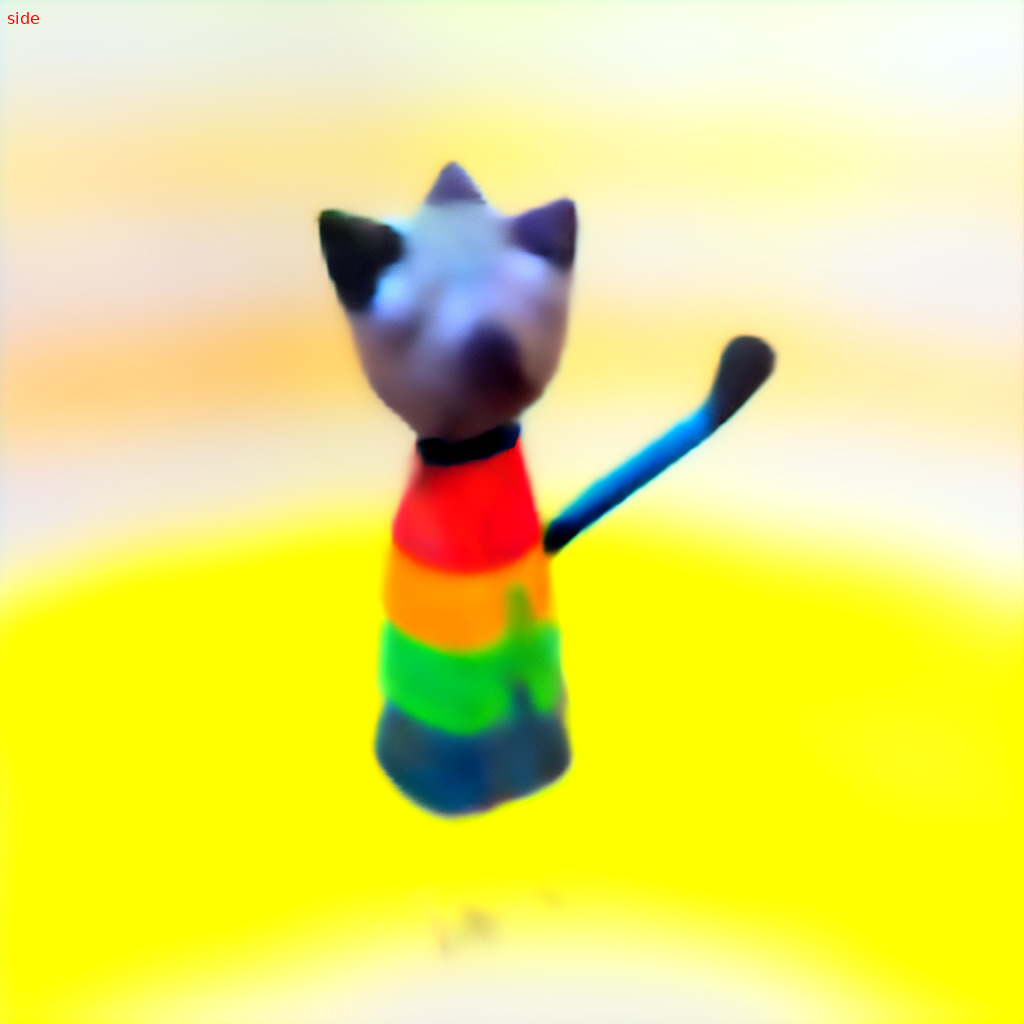} & 
    \includegraphics[width=0.24\linewidth]{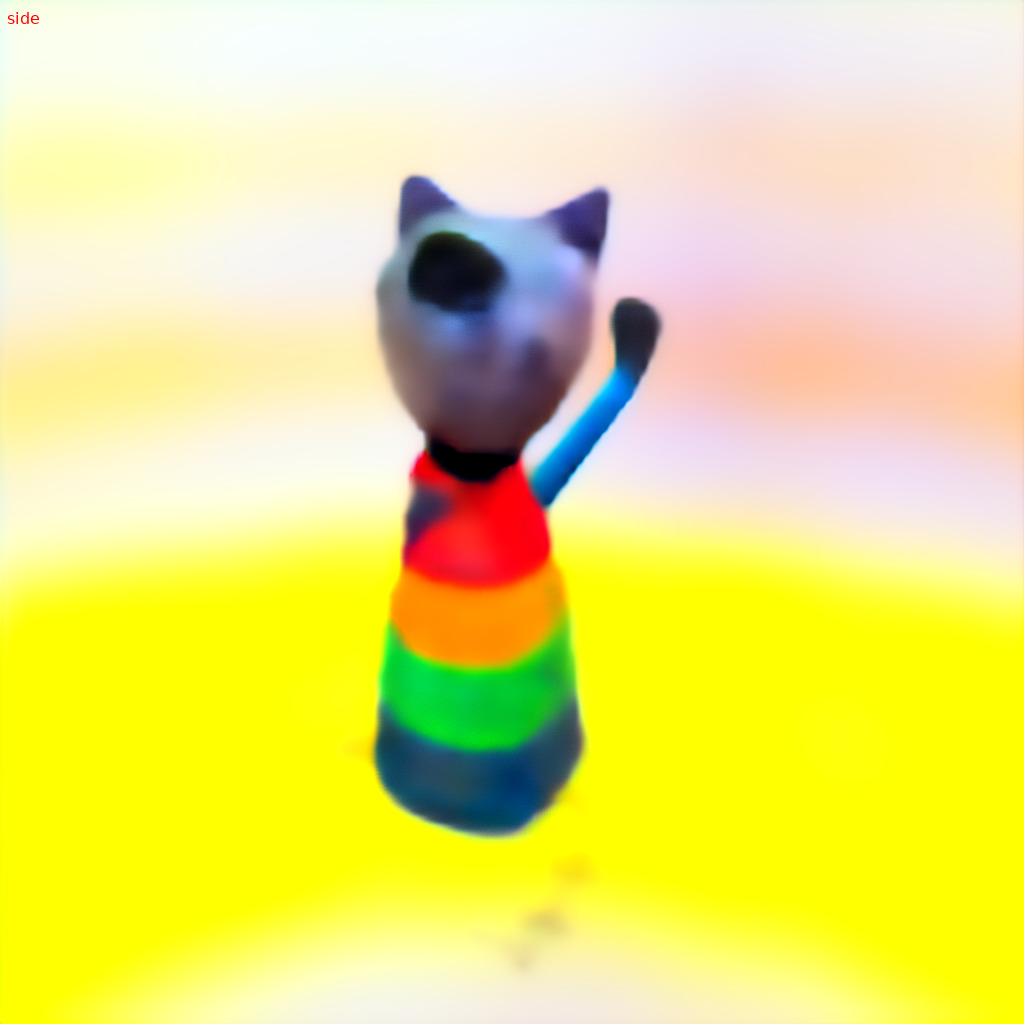}\tabularnewline
    \multicolumn{3}{ c }{a  * sculpture}\tabularnewline
     \includegraphics[width=0.24\linewidth]{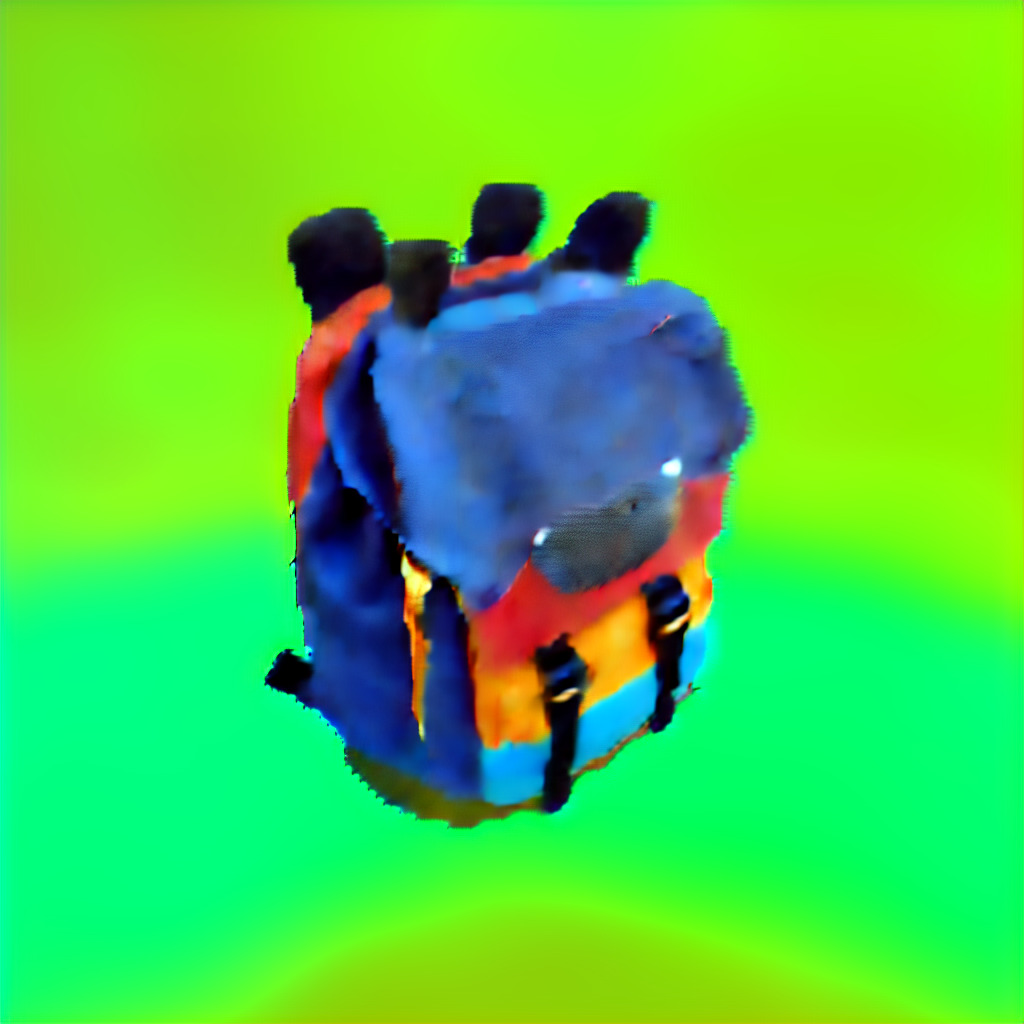} & 
    \includegraphics[width=0.24\linewidth]{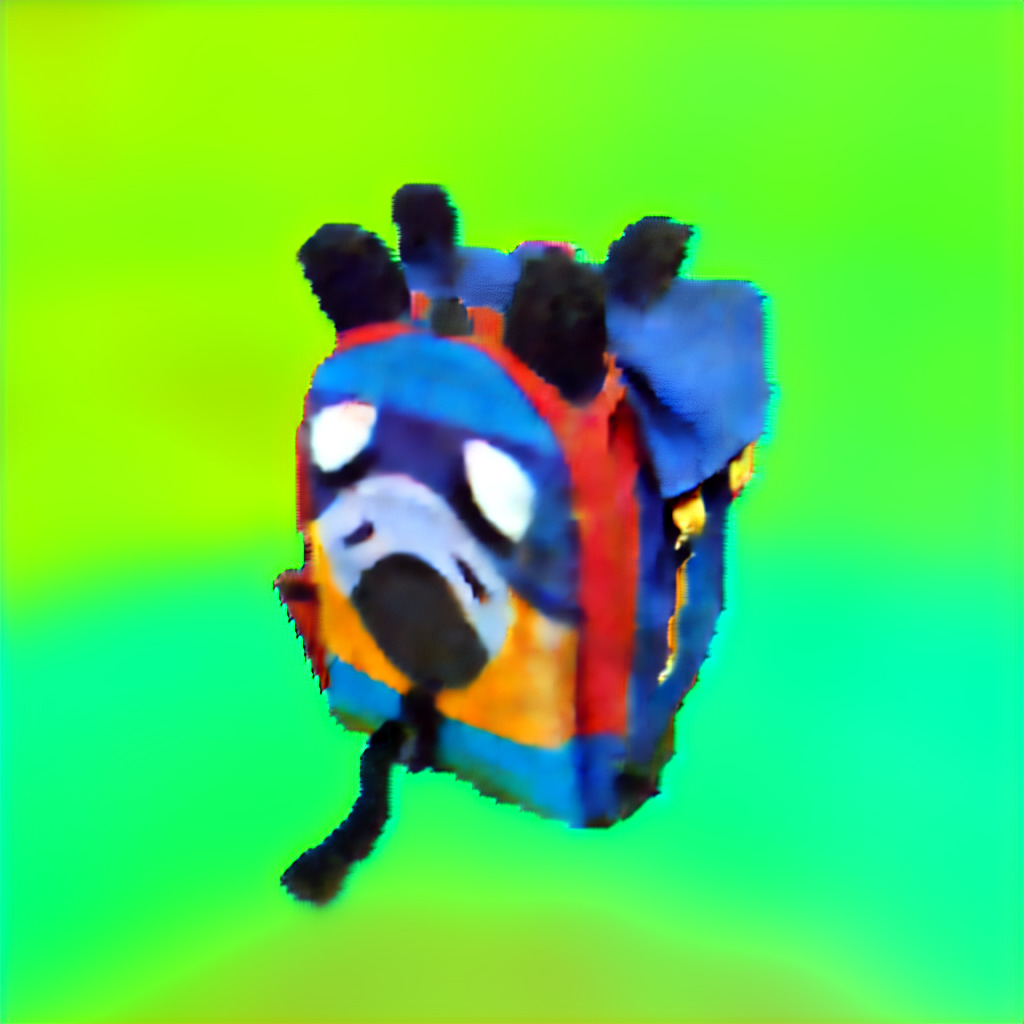} & 
    \includegraphics[width=0.24\linewidth]{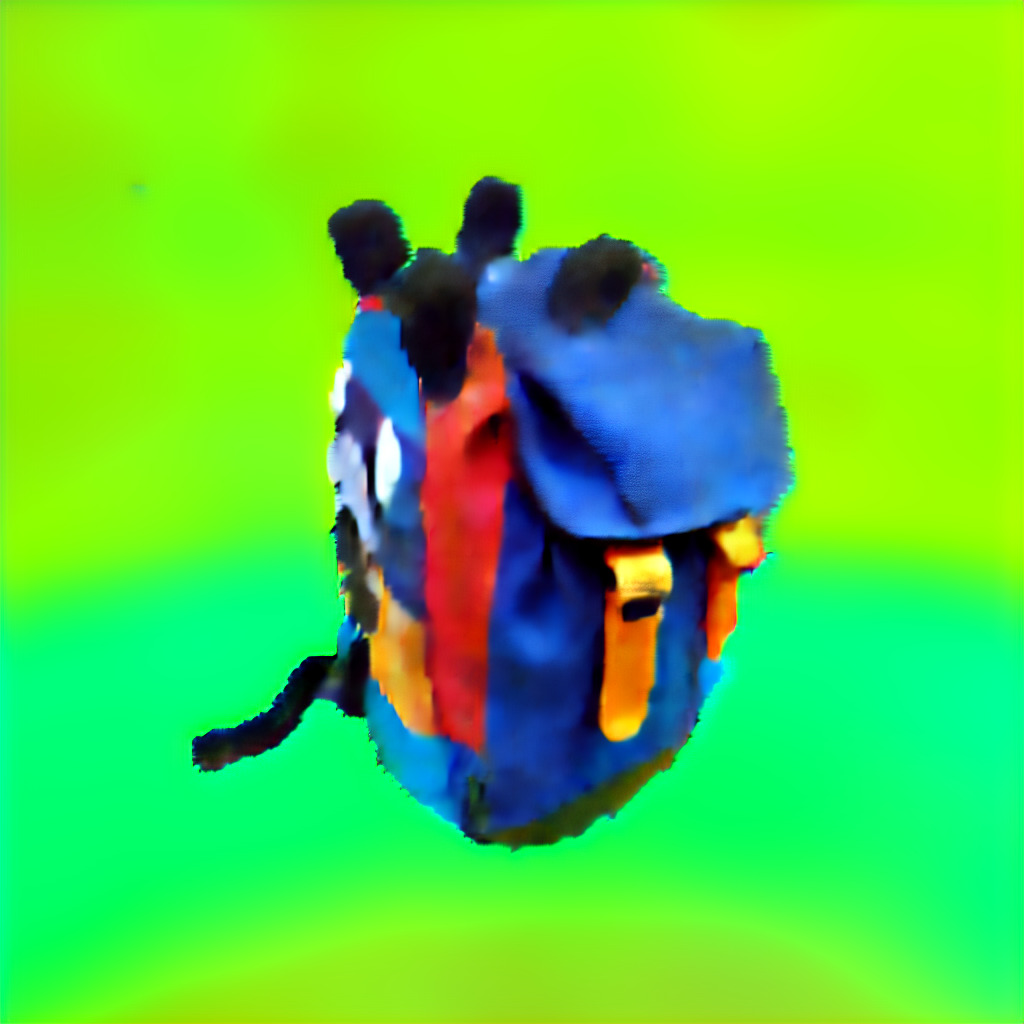} \tabularnewline
    \multicolumn{3}{ c }{a backpack that looks like *}\tabularnewline
    \end{tabular}\tabularnewline
    \end{tabular}
    }
    \caption{Results of Latent-NeRF using a token learned from the images on the left with Textual Inversion~\cite{gal2022textual_inversion}.}
    \label{fig:textual_inversion}
\end{figure}

%% file: figures/4_exp/sketch_mesh/animals/fig.tex
\begin{figure}[t]
    \centering
    \includegraphics[height=0.14\linewidth]{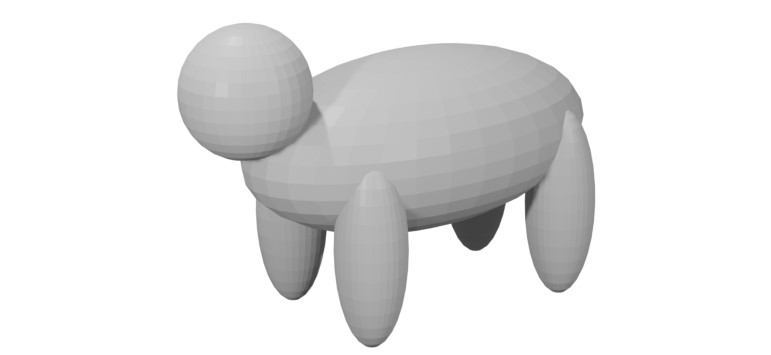} \\
    \scriptsize{Input}
    \setlength{\tabcolsep}{0.5pt}
    {\scriptsize
    \begin{tabular}{c c c c c}
        \raisebox{0.052\textwidth}{\rotatebox[origin=t]{90}{\scalebox{0.9}{``A Deer''}}} & 
        \includegraphics[width=0.235\linewidth]{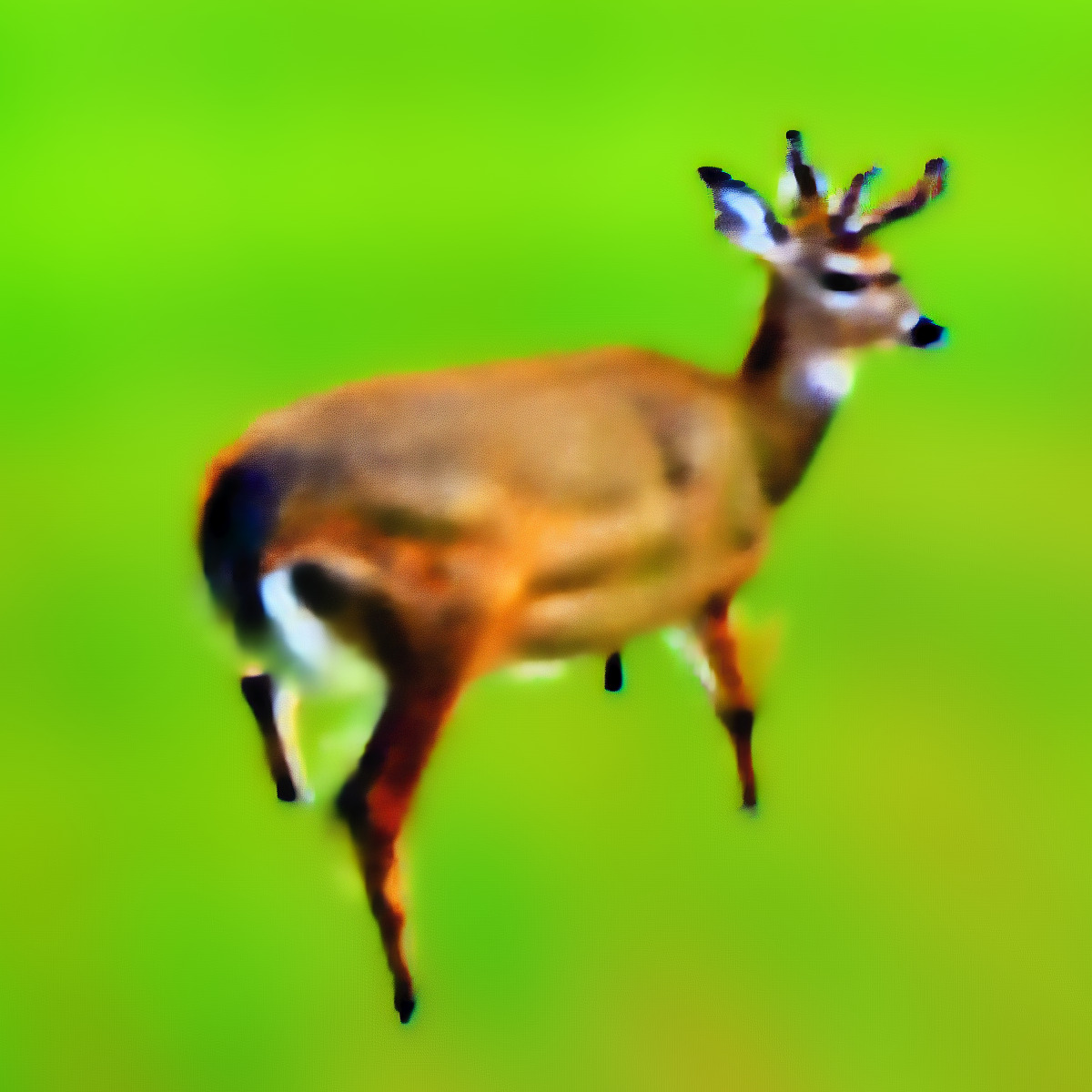} & 
        \includegraphics[width=0.235\linewidth]{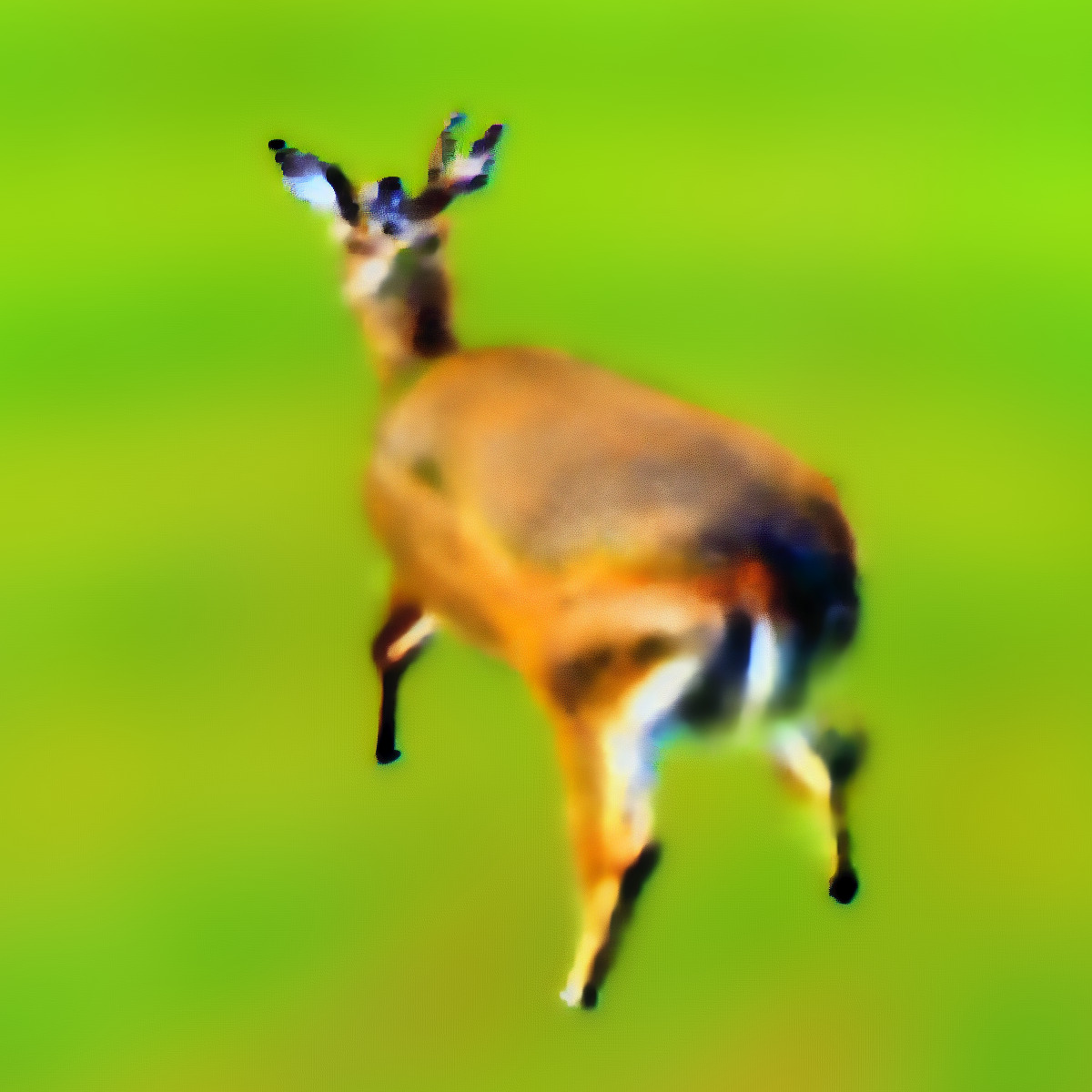} & 
        \includegraphics[width=0.235\linewidth]{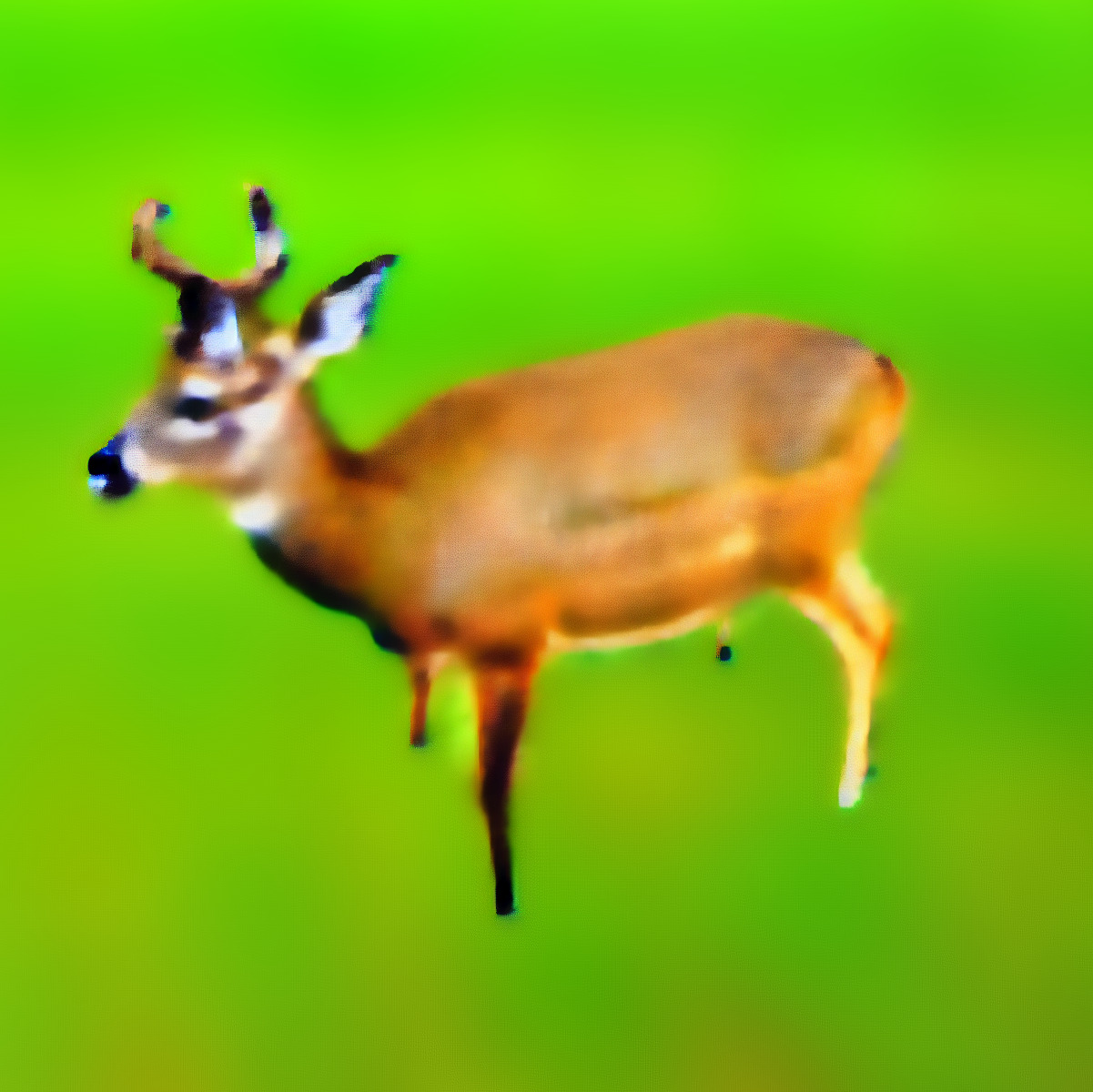} & 
        \includegraphics[width=0.235\linewidth]{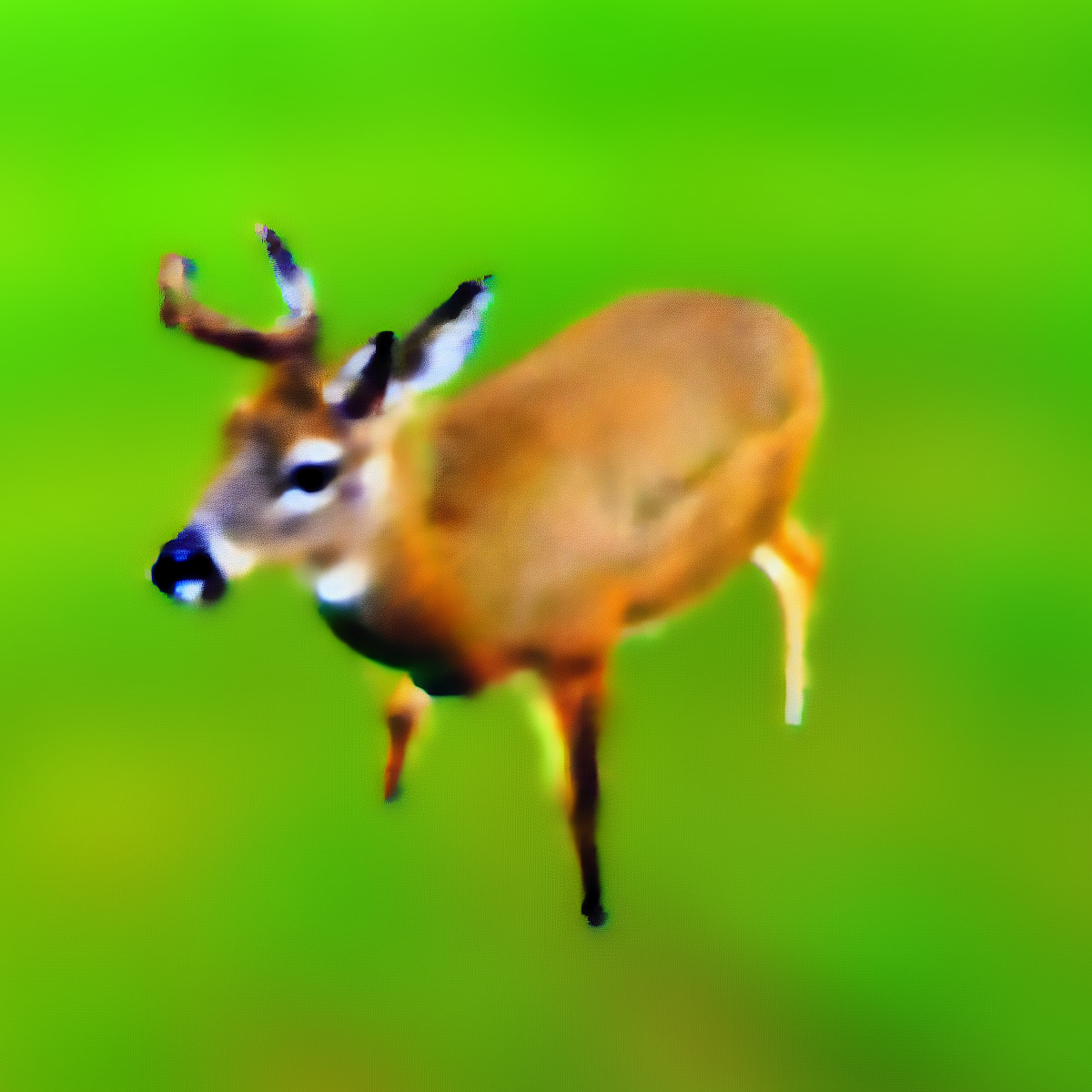} \\
        \raisebox{0.052\textwidth}{\rotatebox[origin=t]{90}{\scalebox{0.9}{``A German Sheperd''}}} & 
        \includegraphics[width=0.235\linewidth]{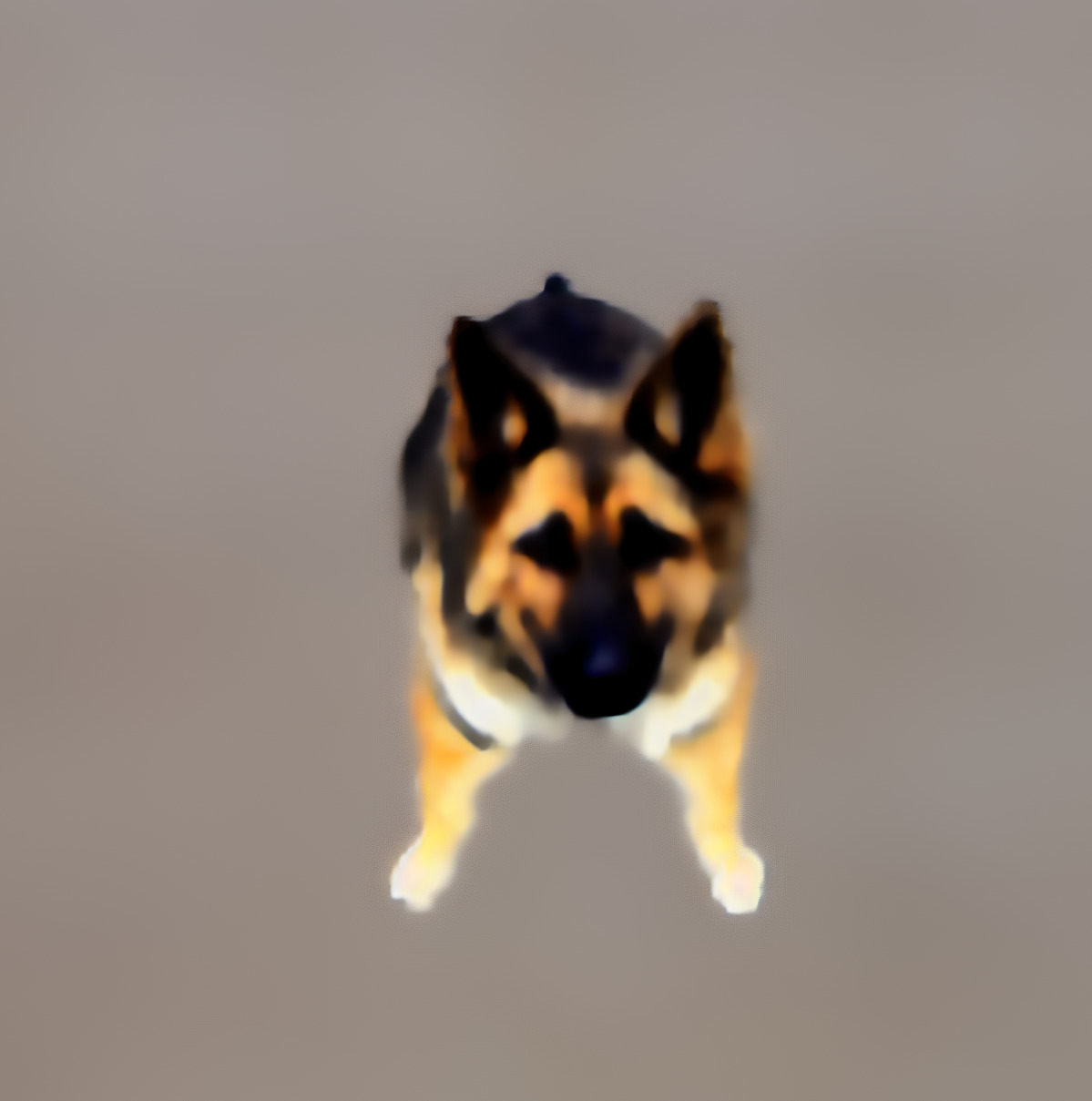} & 
        \includegraphics[width=0.235\linewidth]{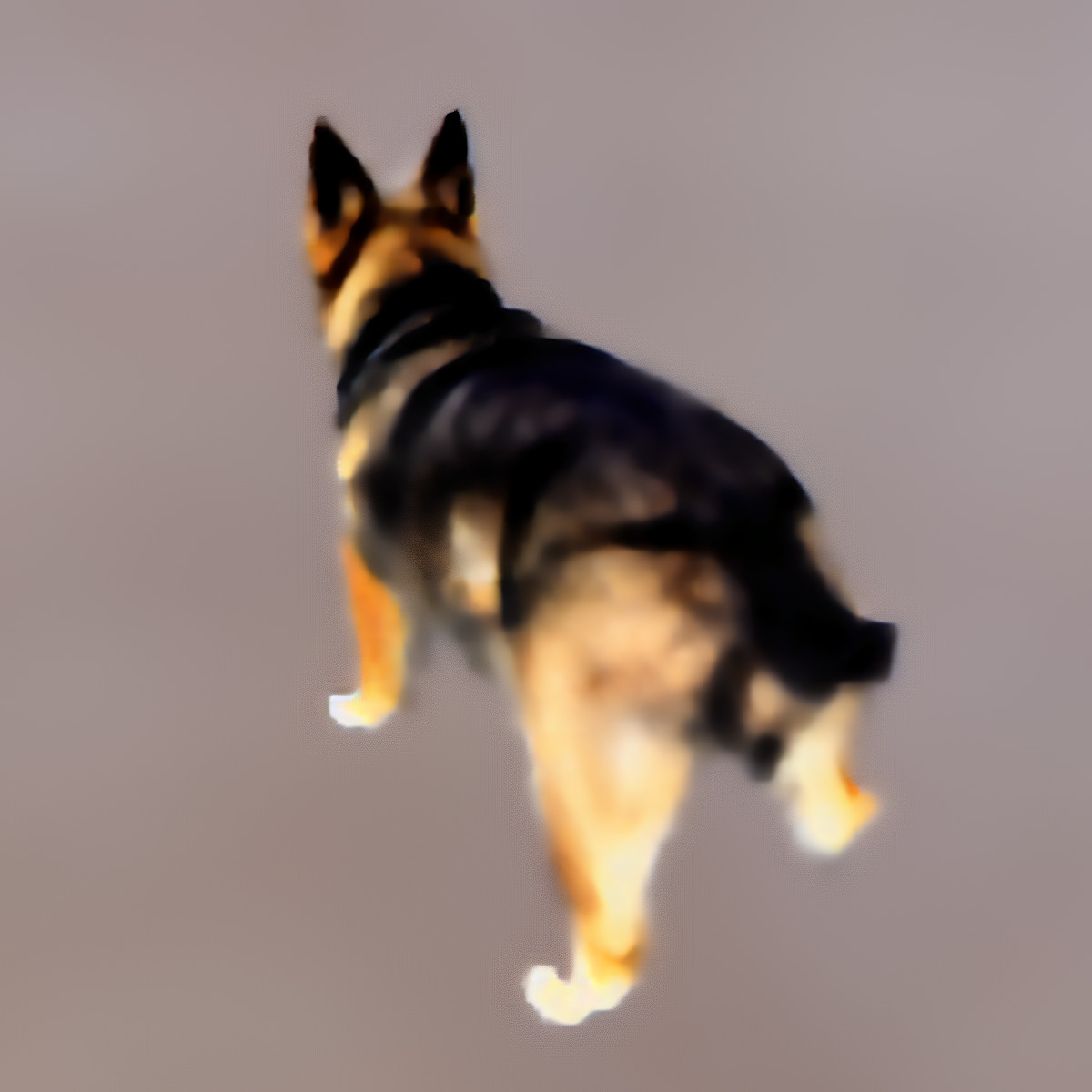} & 
        \includegraphics[width=0.235\linewidth]{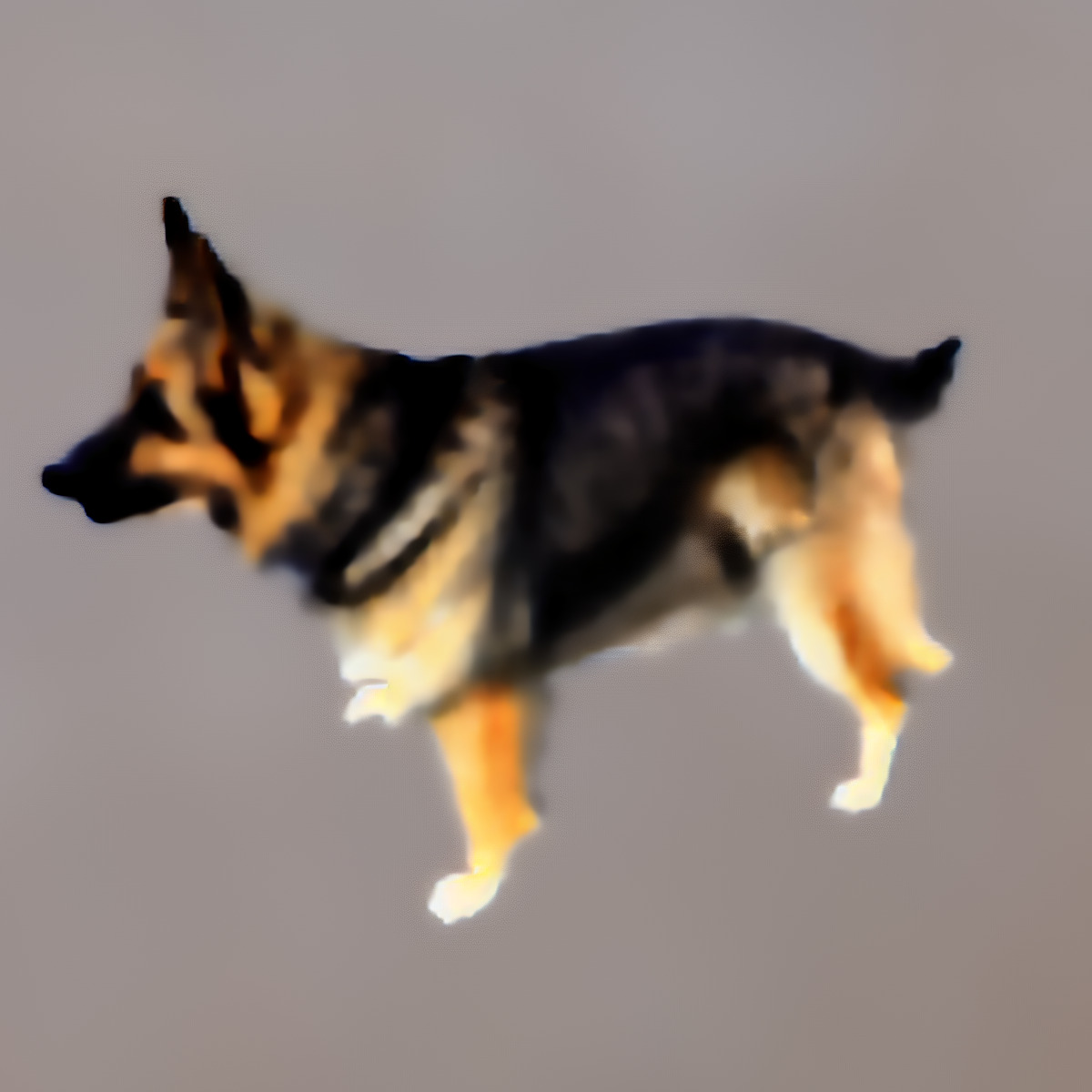} & 
        \includegraphics[width=0.235\linewidth]{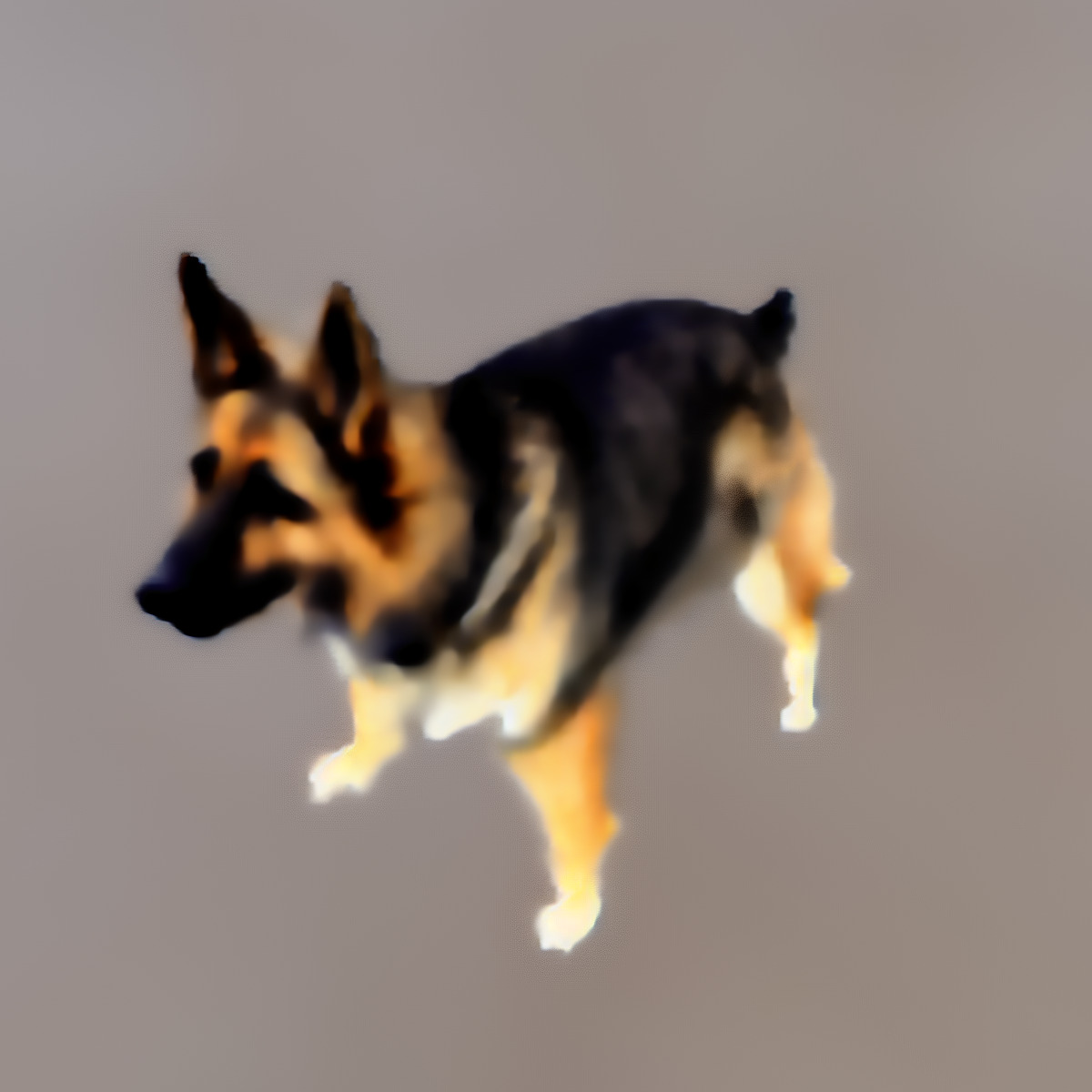} \\
        \raisebox{0.052\textwidth}{\rotatebox[origin=t]{90}{\scalebox{0.9}{``A Pig''}}} & 
        \includegraphics[width=0.235\linewidth]{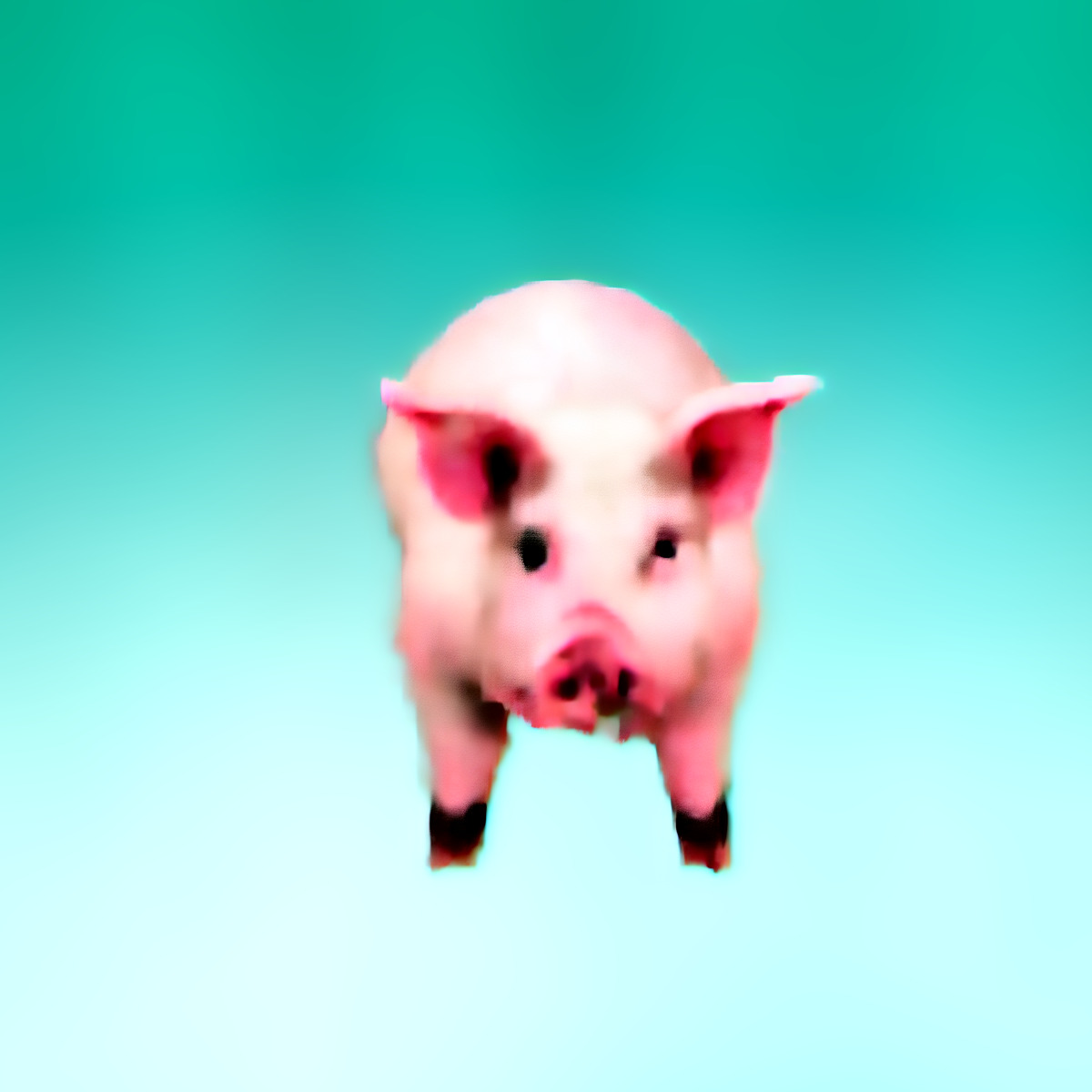} & 
        \includegraphics[width=0.235\linewidth]{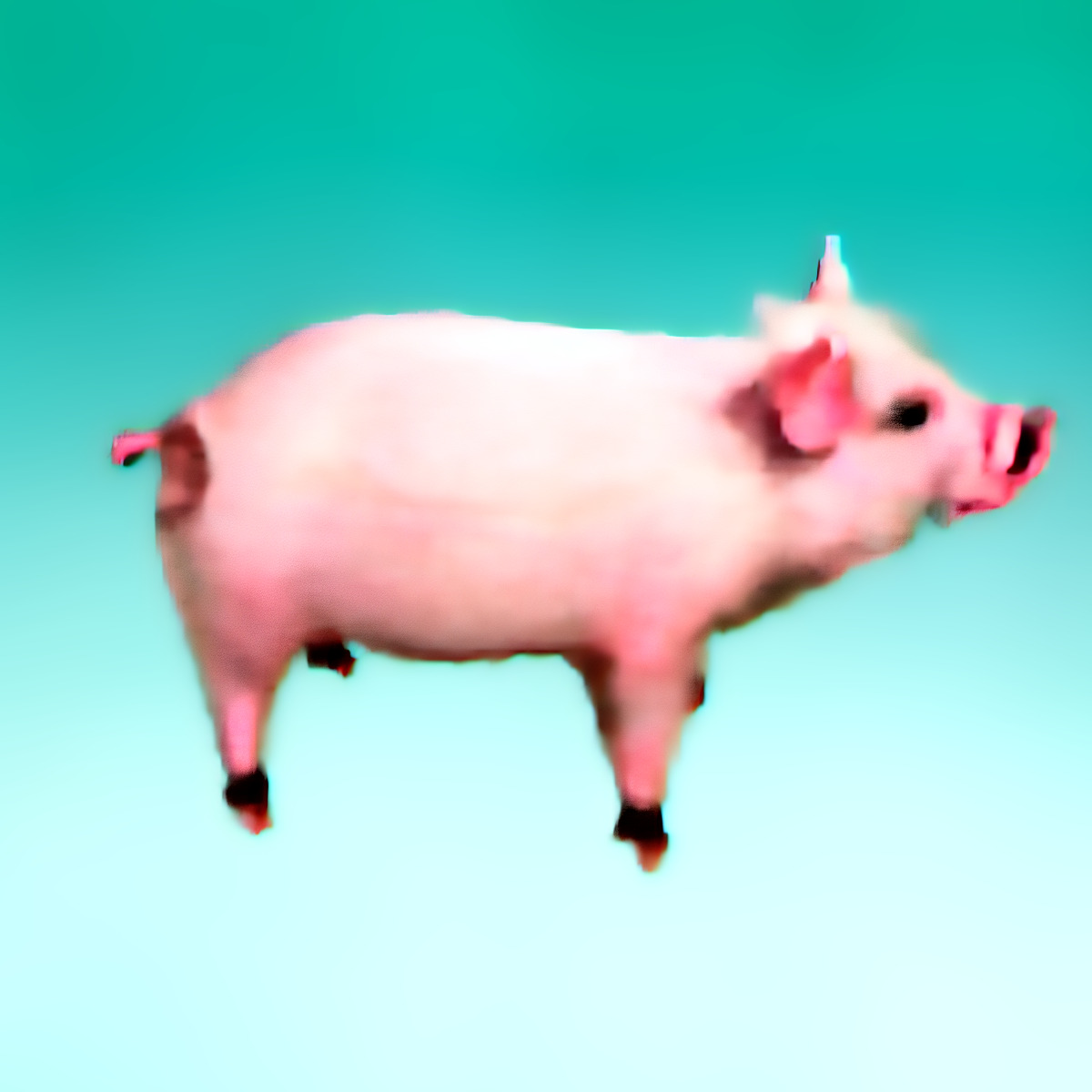} & 
        \includegraphics[width=0.235\linewidth]{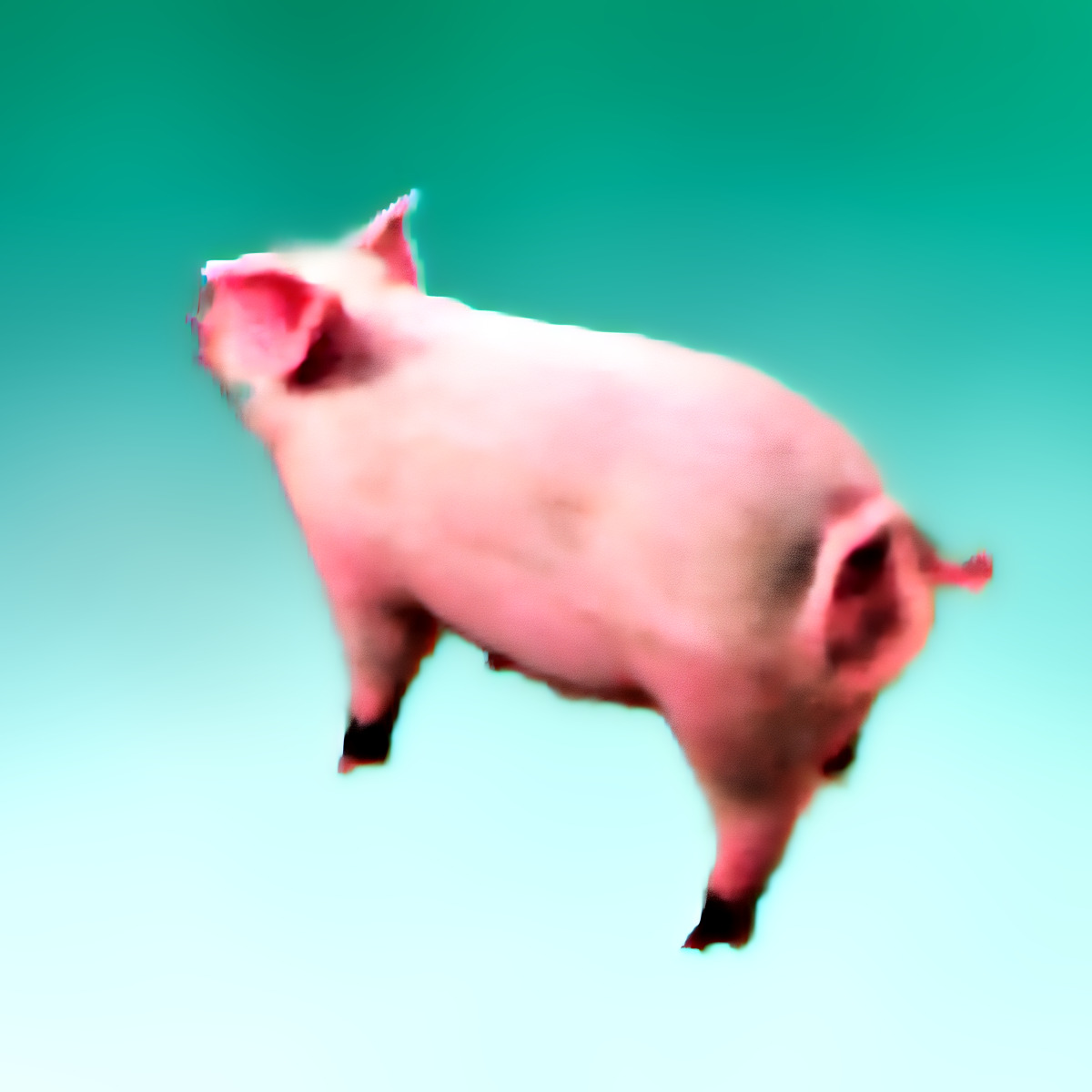} & 
        \includegraphics[width=0.235\linewidth]{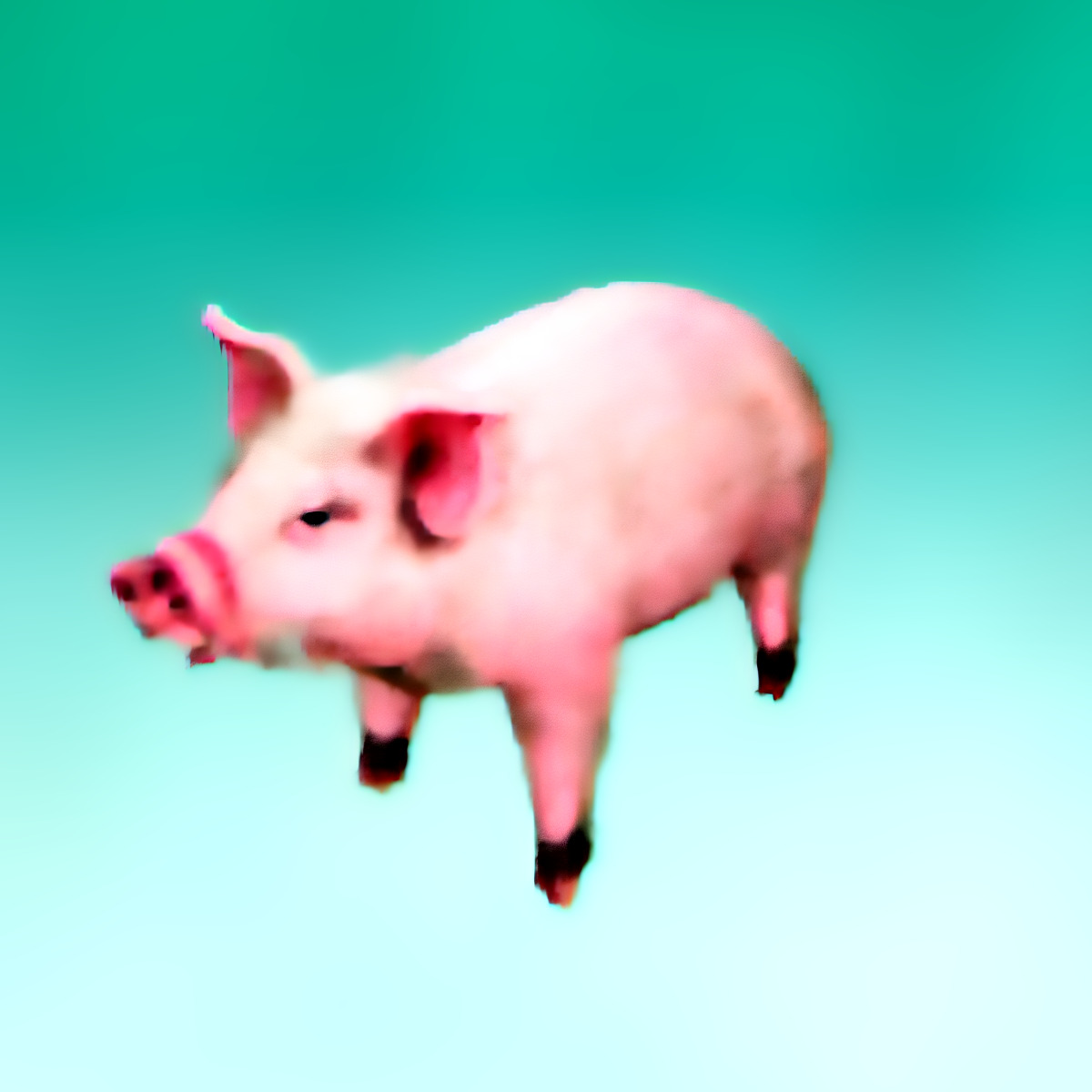} 
        
    \end{tabular}}
    \caption{\meshsketch{} results using different prompts. This examples shows how the same \meshsketch{} can influence different objects according to different text prompts.}
    \label{fig:skecth_mesh_animals}
\end{figure} 

%% file: figures/4_exp/sketch_mesh/house/fig.tex
\begin{figure}[h]
    \centering
    \includegraphics[height=0.16\linewidth]{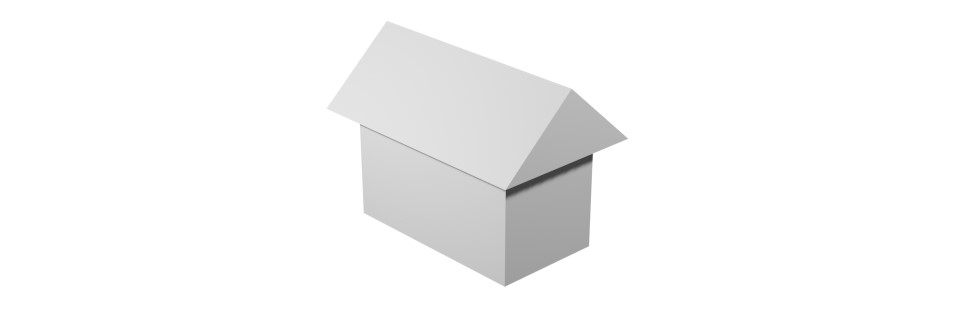} \\
    \scriptsize{Input}
    \setlength{\tabcolsep}{0.5pt}
    {\scriptsize
    \begin{tabular}{c c c c}
        \includegraphics[width=0.24\linewidth]{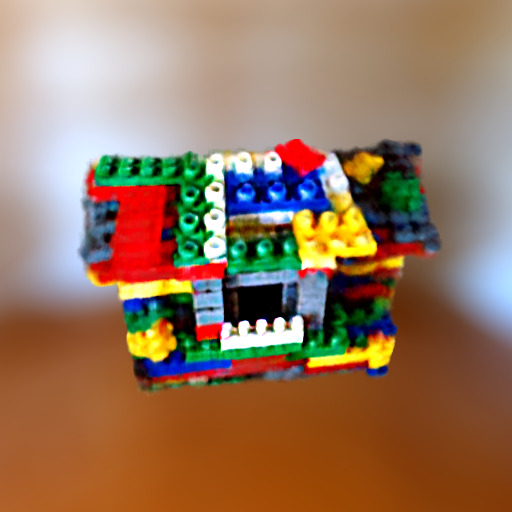} & 
        \includegraphics[width=0.24\linewidth]{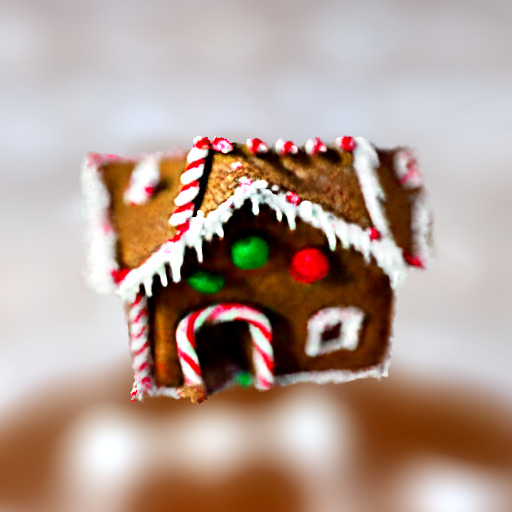} & 
        \includegraphics[width=0.24\linewidth]{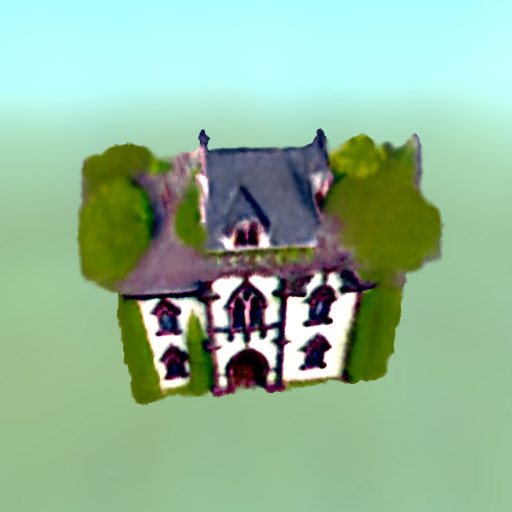} & 
        \includegraphics[width=0.24\linewidth]{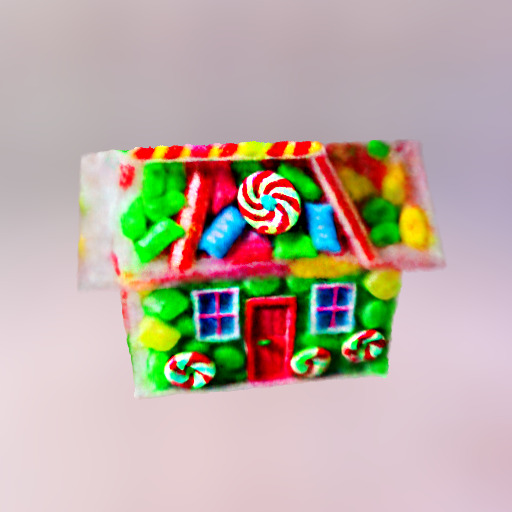} \\
        \includegraphics[width=0.24\linewidth]{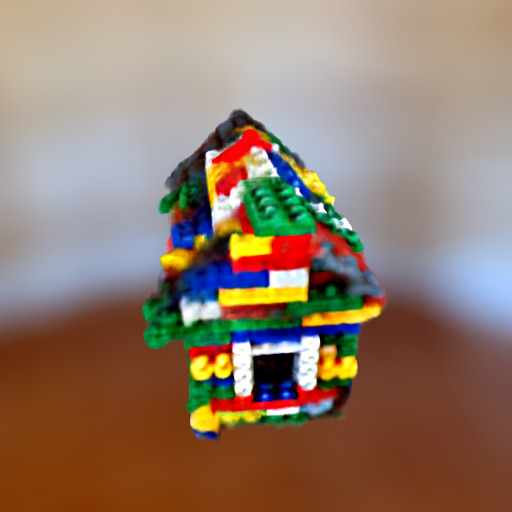} & 
        \includegraphics[width=0.24\linewidth]{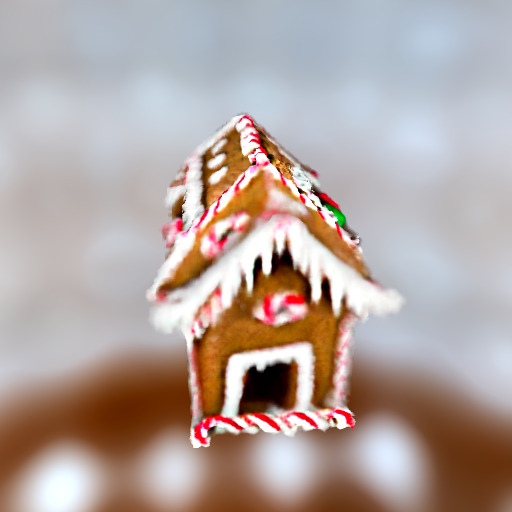} & 
        \includegraphics[width=0.24\linewidth]{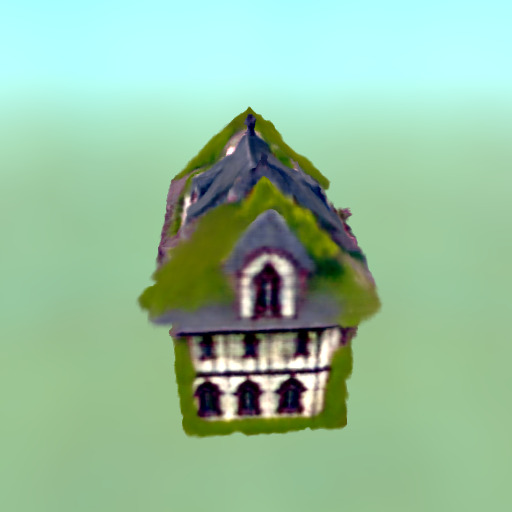} & 
        \includegraphics[width=0.24\linewidth]{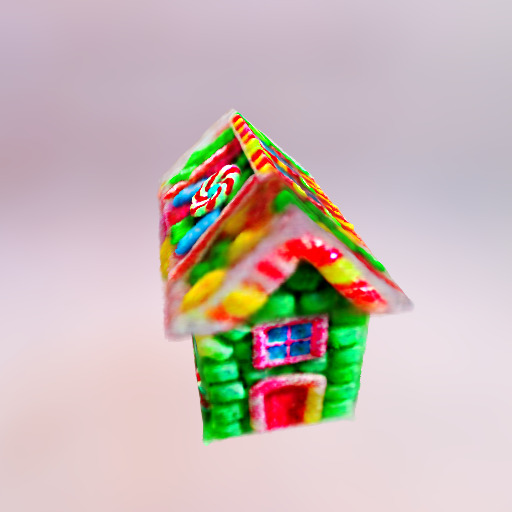} \\
        \includegraphics[width=0.24\linewidth]{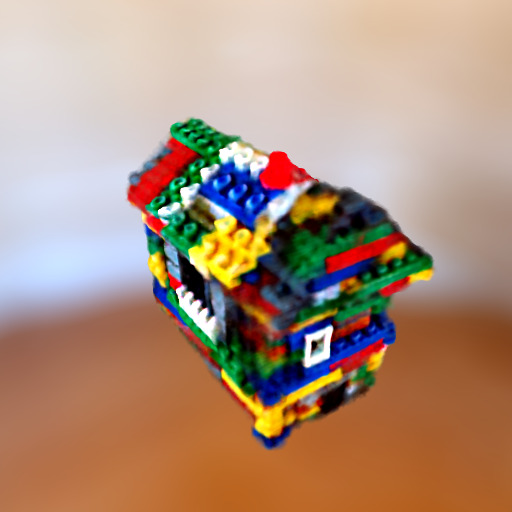} & 
        \includegraphics[width=0.24\linewidth]{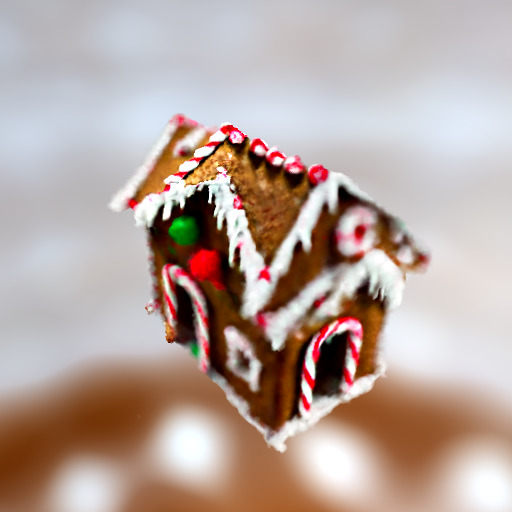} & 
        \includegraphics[width=0.24\linewidth]{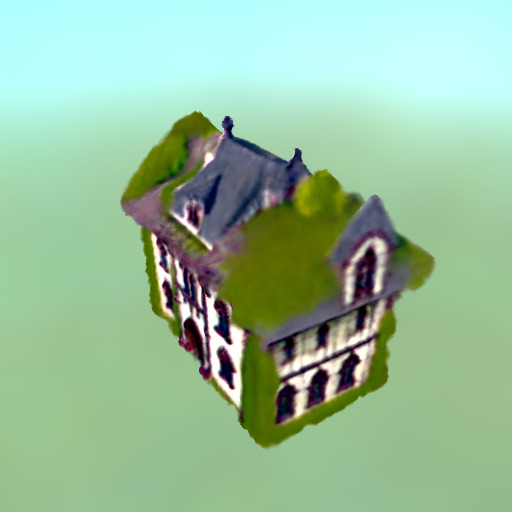} & 
        \includegraphics[width=0.24\linewidth]{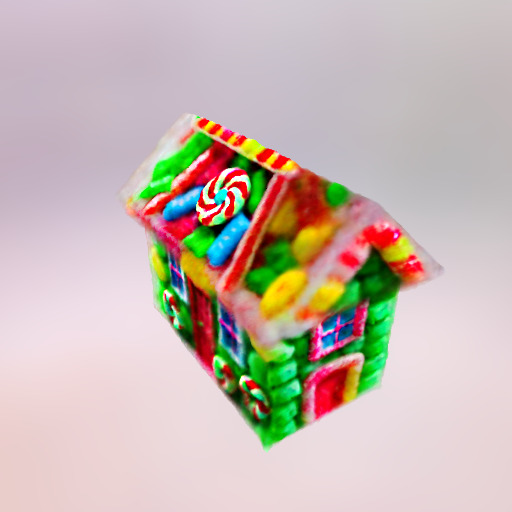} \\
        \includegraphics[width=0.24\linewidth]{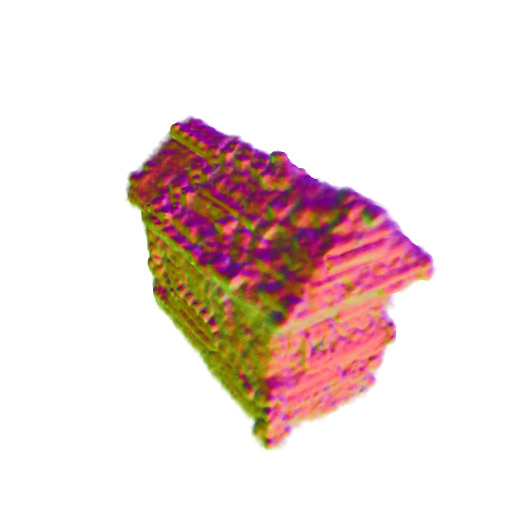} & 
        \includegraphics[width=0.24\linewidth]{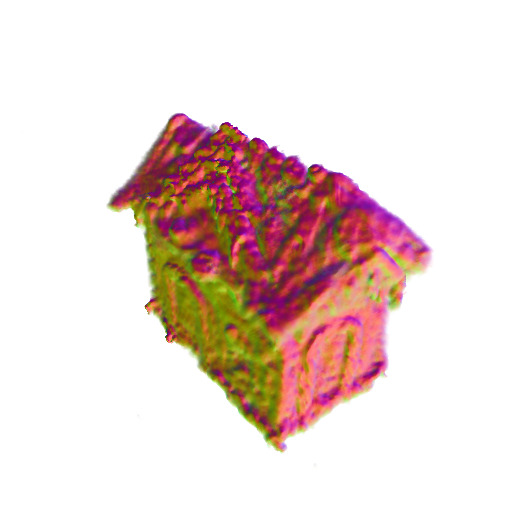} & 
        \includegraphics[width=0.24\linewidth]{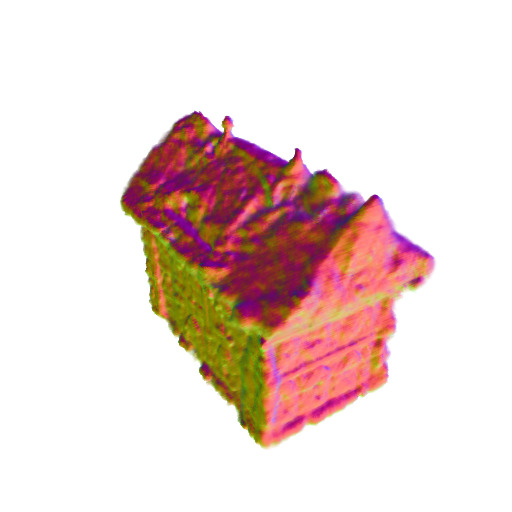} & 
        \includegraphics[width=0.24\linewidth]{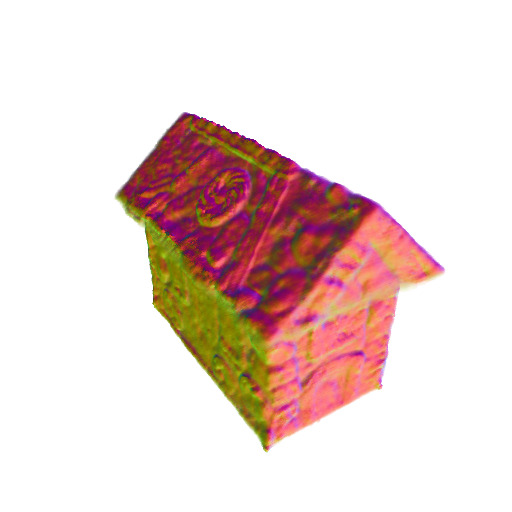} \\
        ``A house made  & ``A gingerbread  & ``A gothic & ``A house made  \\
        of lego'' & house'' & house'' & of candy''
        
    \end{tabular}}
    \caption{\meshsketch{} results, conditioned on a low-poly house shape. Our RGB Refinement was applied to improve detail quality.}
    \label{fig:skecth_mesh_house}
\end{figure} 

%% file: figures/4_exp/sketch_mesh/general/fig.tex
\begin{figure}[h]
    \centering
    \setlength{\tabcolsep}{0.5pt}
    {\scriptsize

    \begin{tabular}{ c c c c }
    \includegraphics[width=0.24\linewidth]{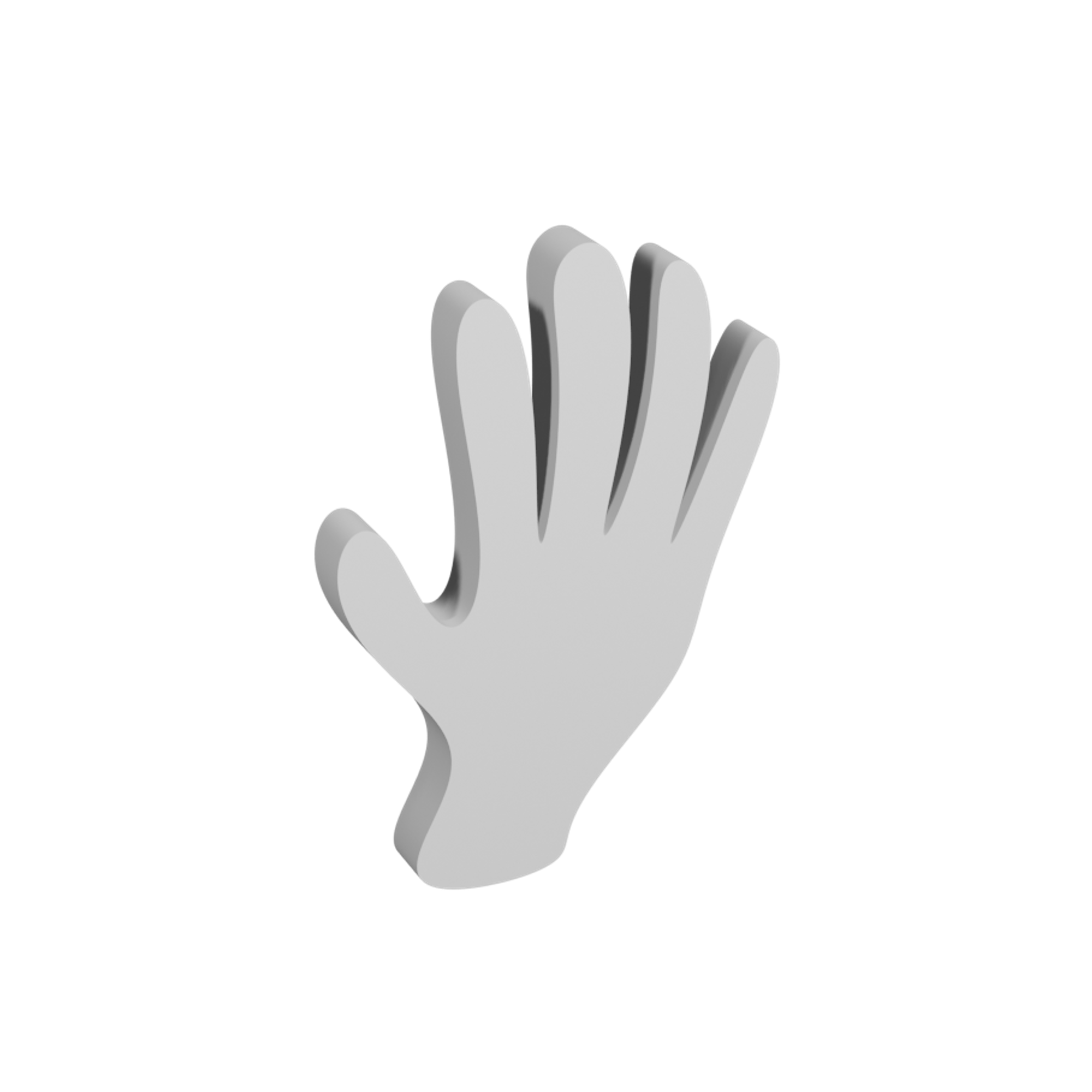} & 
    \includegraphics[width=0.24\linewidth]{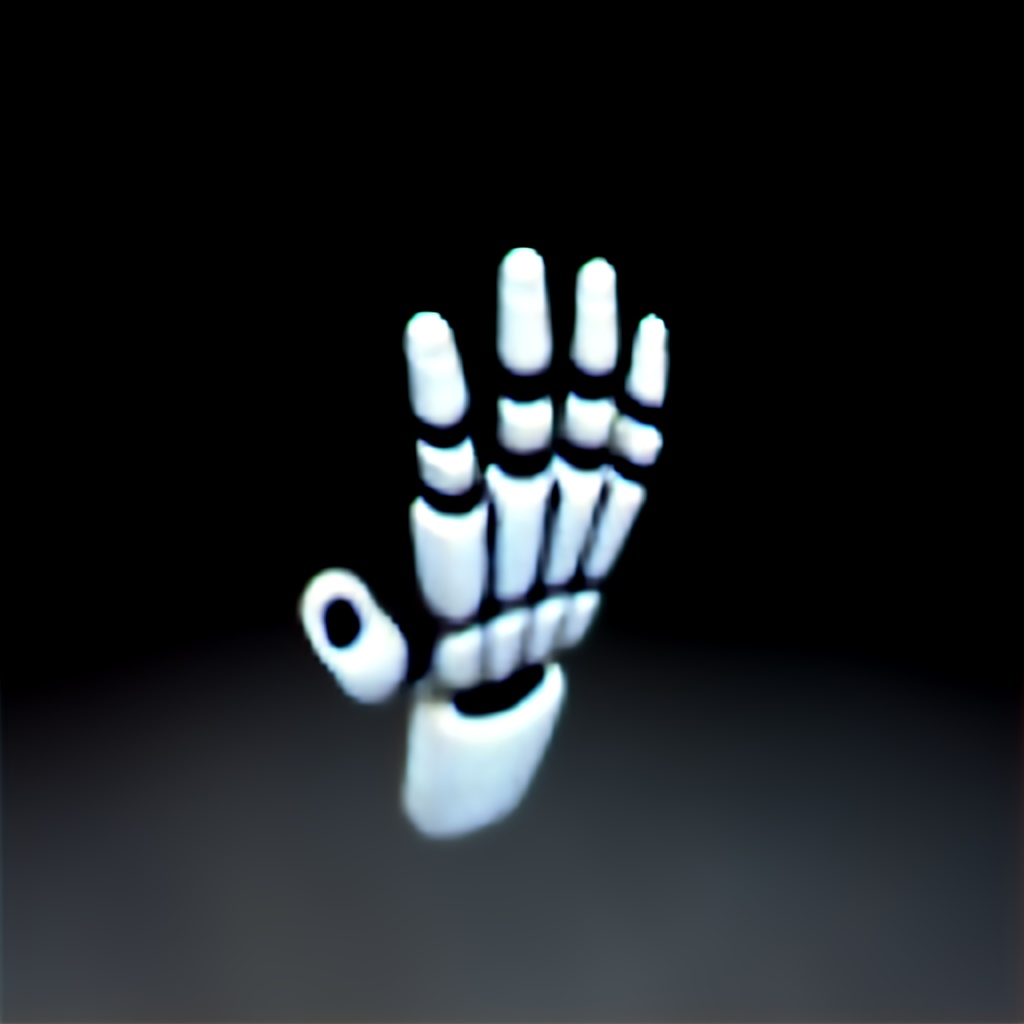} & \includegraphics[width=0.24\linewidth]{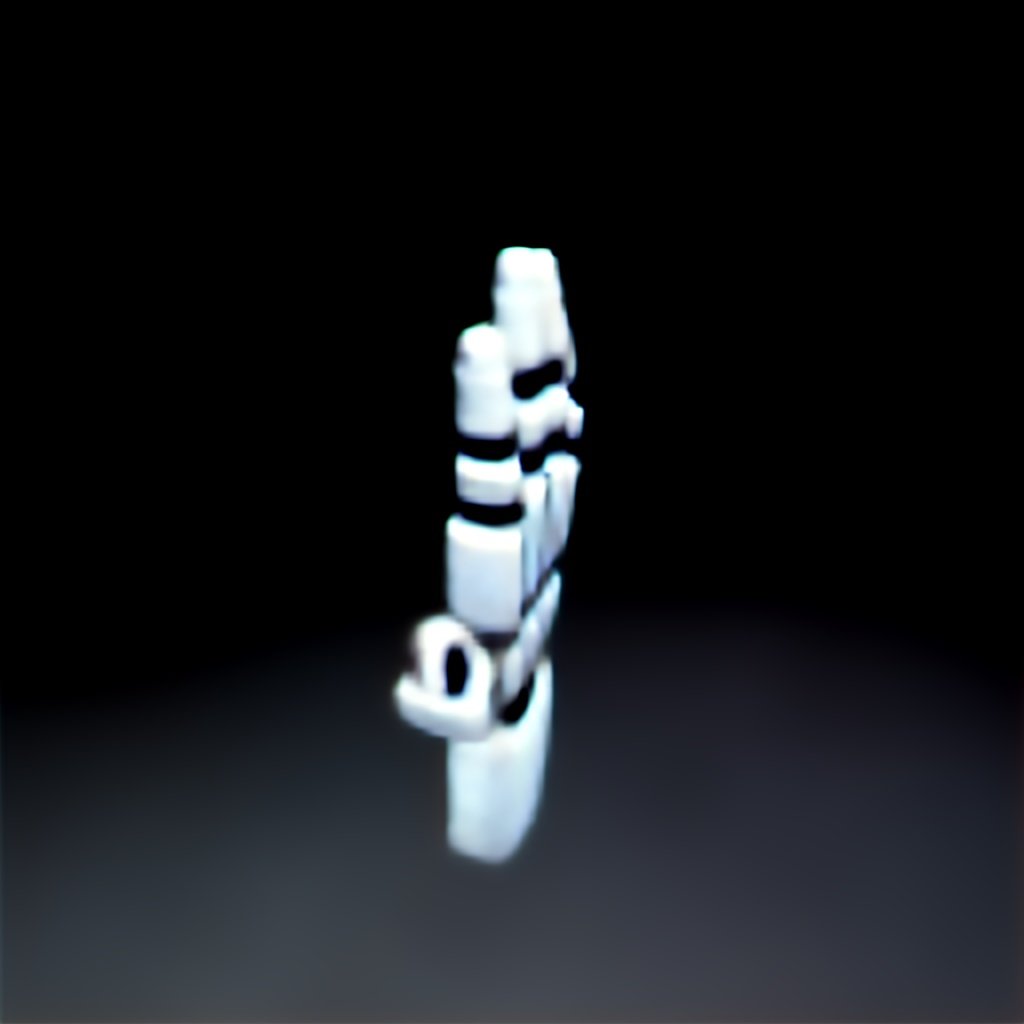} & 
    \includegraphics[width=0.24\linewidth]{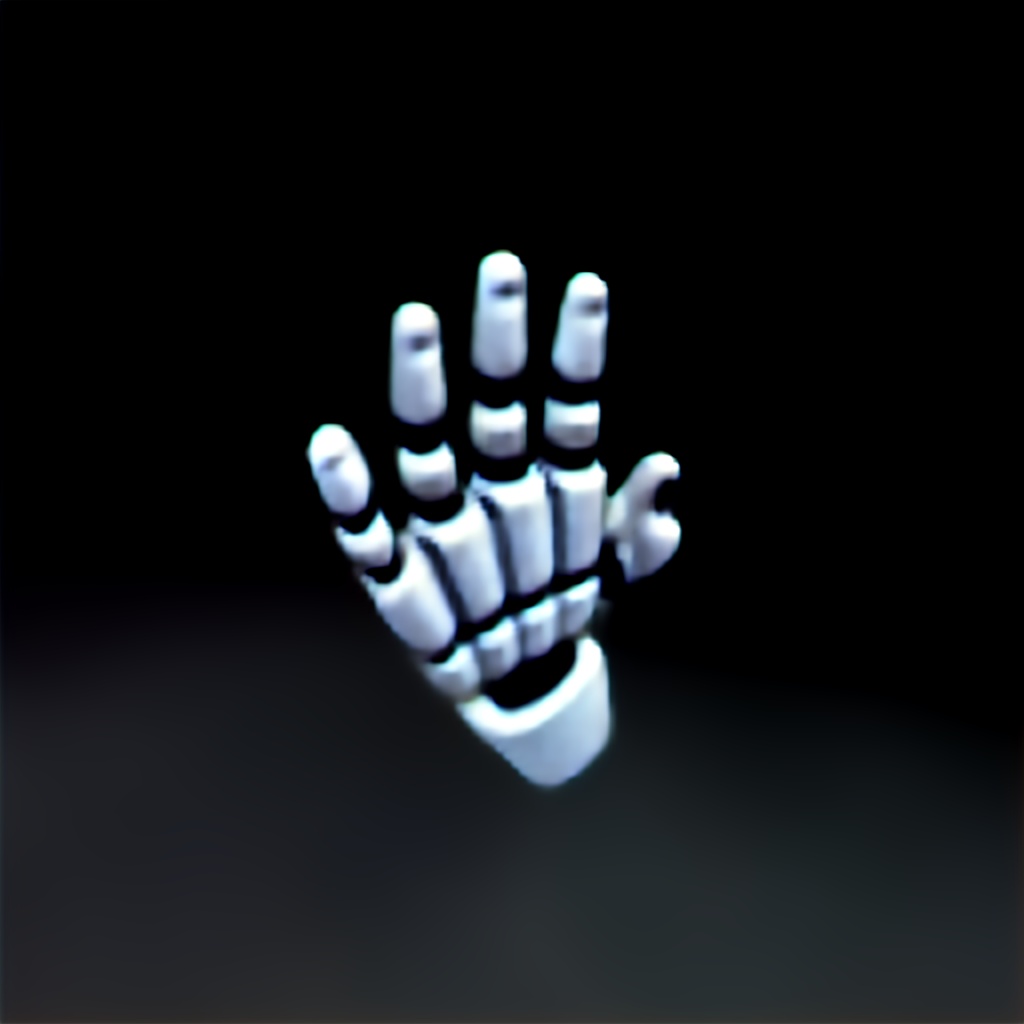}\tabularnewline
    &\multicolumn{3}{c}{``a robot hand, realistic''} \tabularnewline
    \includegraphics[width=0.24\linewidth]{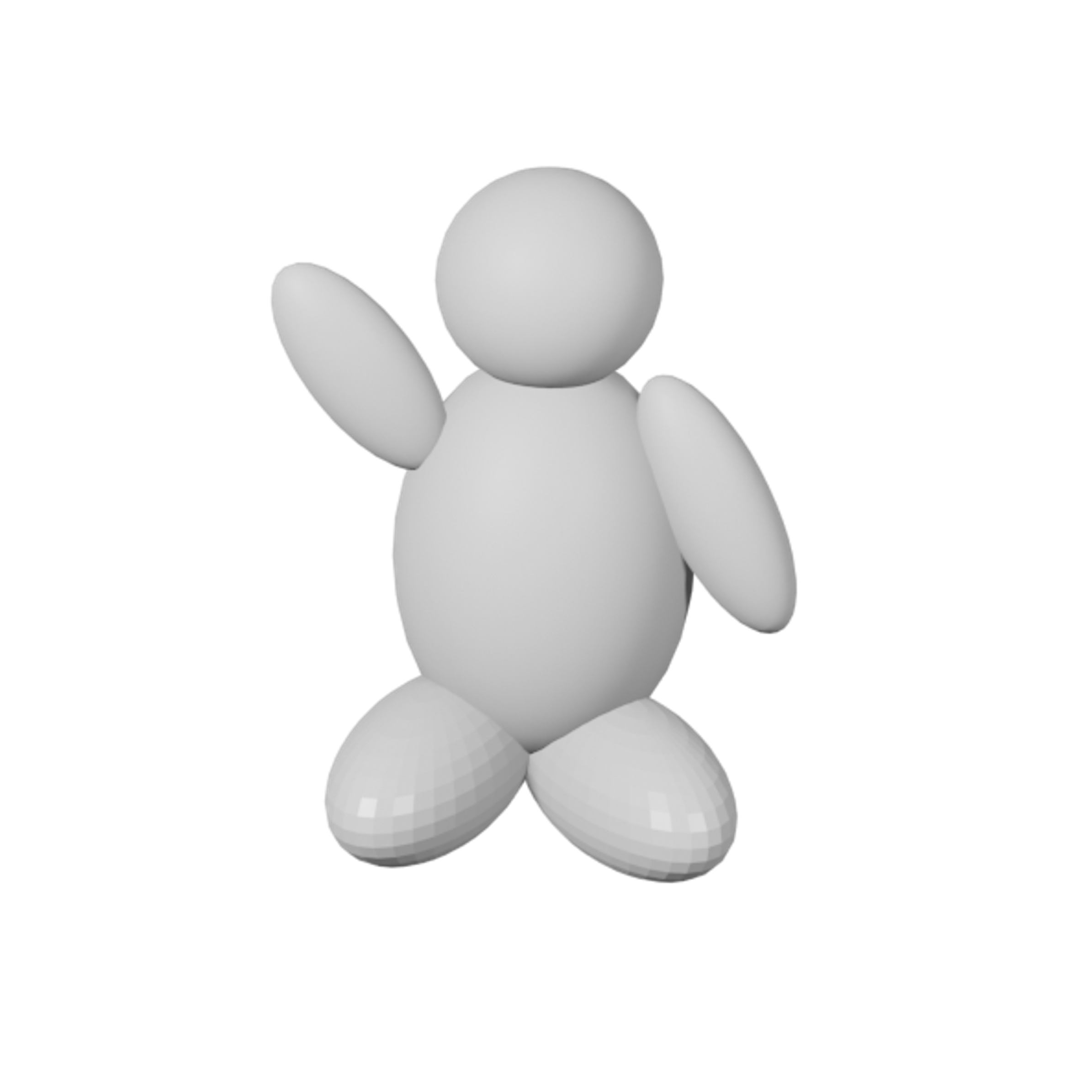} & 
    \includegraphics[width=0.24\linewidth]{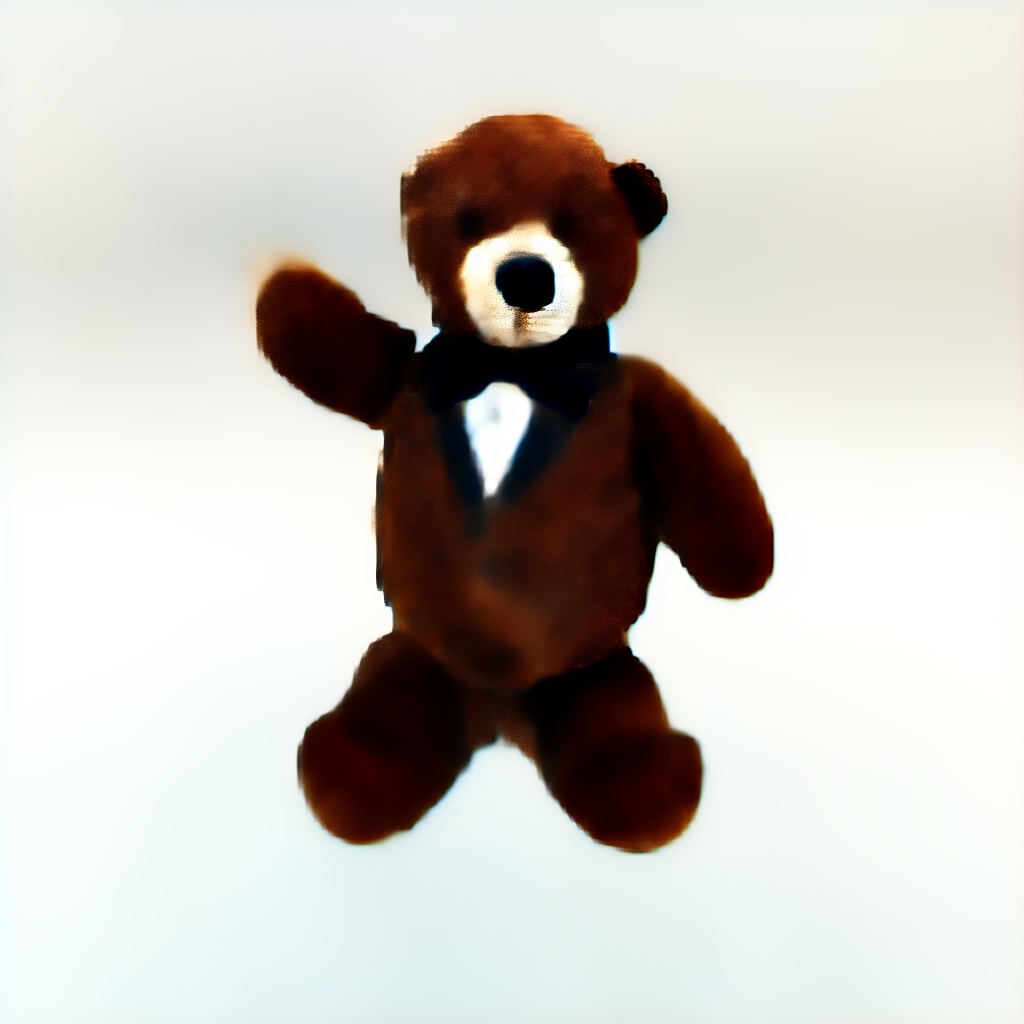} & 
    \includegraphics[width=0.24\linewidth]{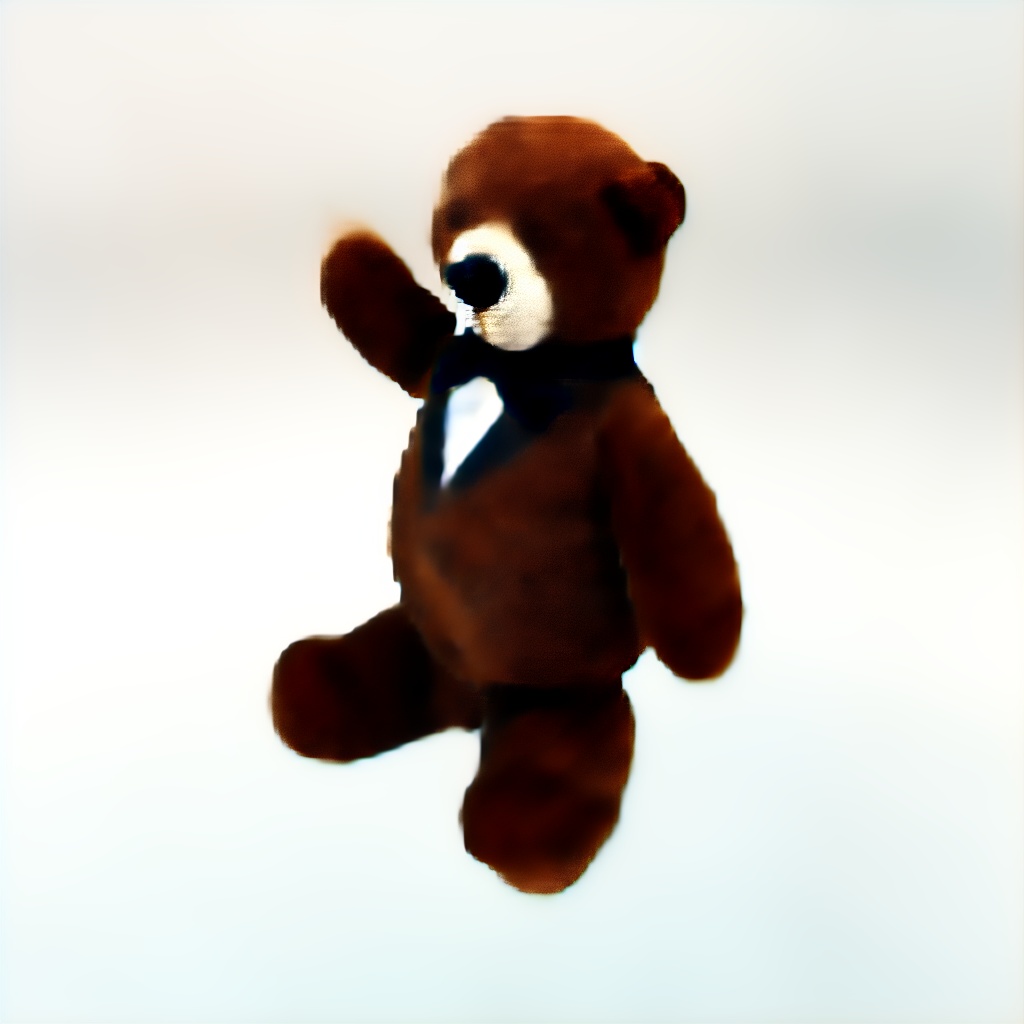} & 
    \includegraphics[width=0.24\linewidth]{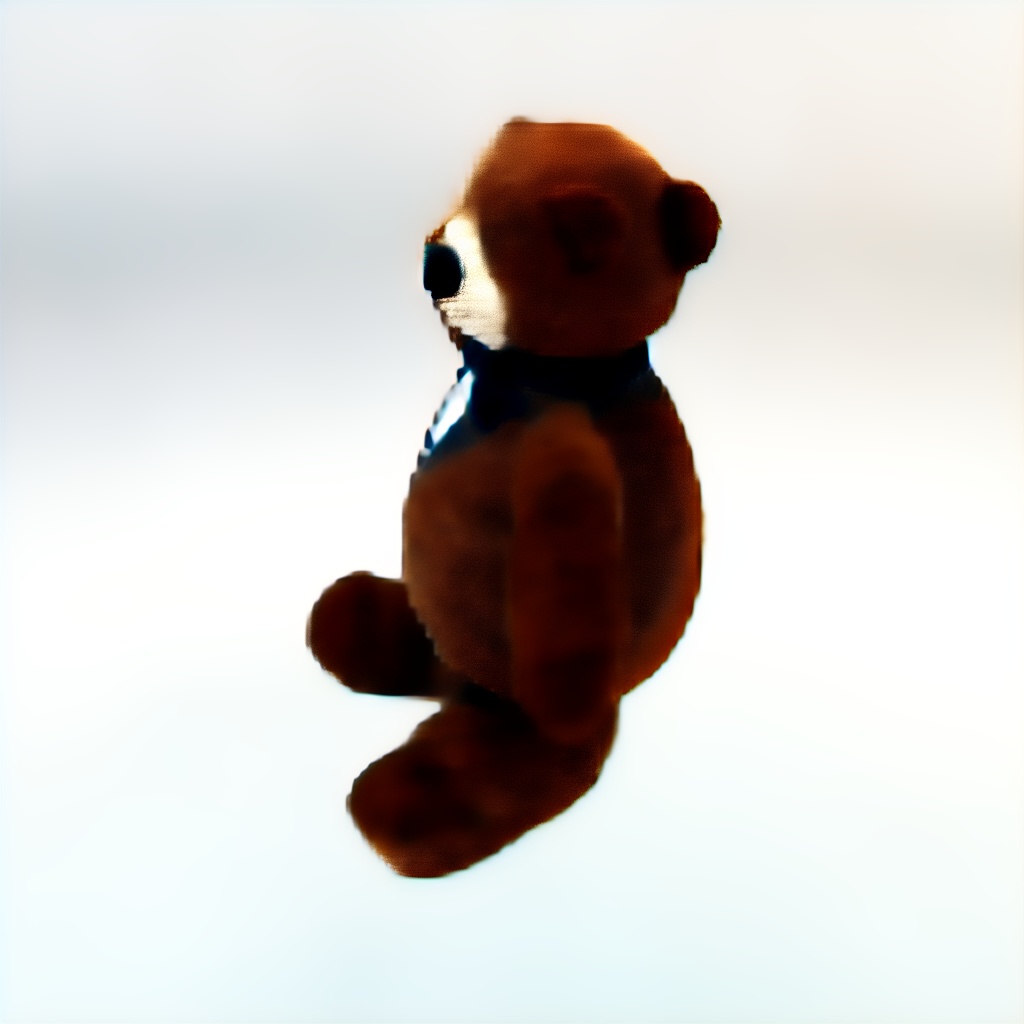}\tabularnewline
    &\multicolumn{3}{c}{``a teddy bear in a tuxedo''} \tabularnewline
    \includegraphics[width=0.24\linewidth]{figures/4_exp/sketch_mesh/general/images/teddy_sketch.png} & 
    \includegraphics[width=0.24\linewidth]{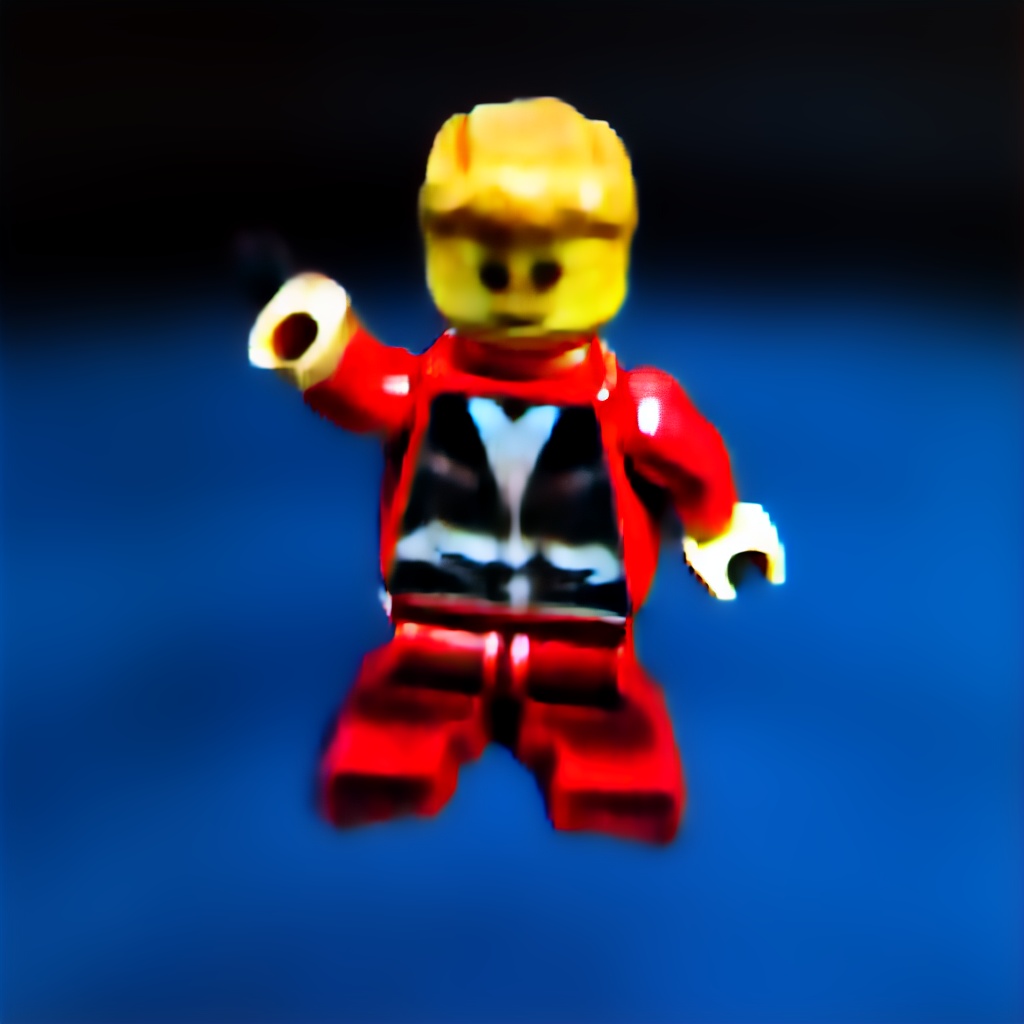} & 
    \includegraphics[width=0.24\linewidth]{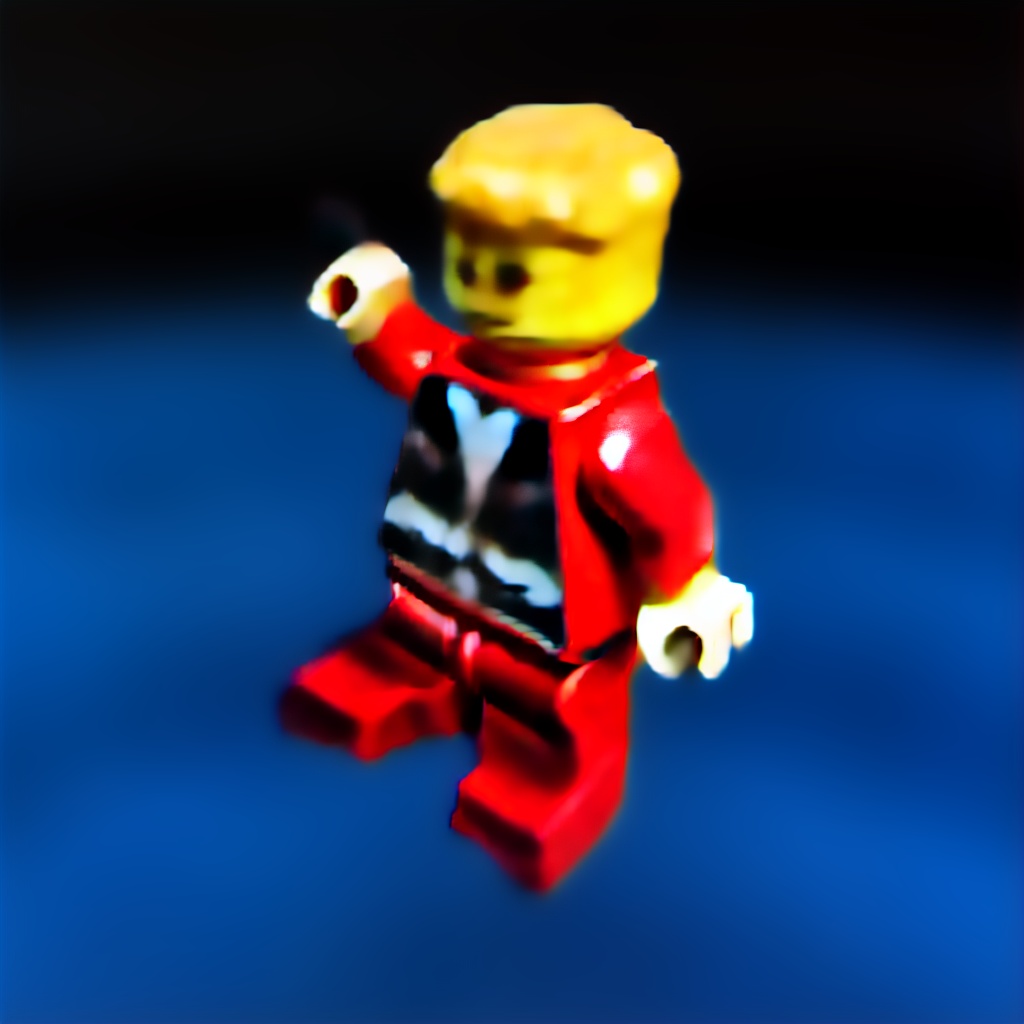} & 
    \includegraphics[width=0.24\linewidth]{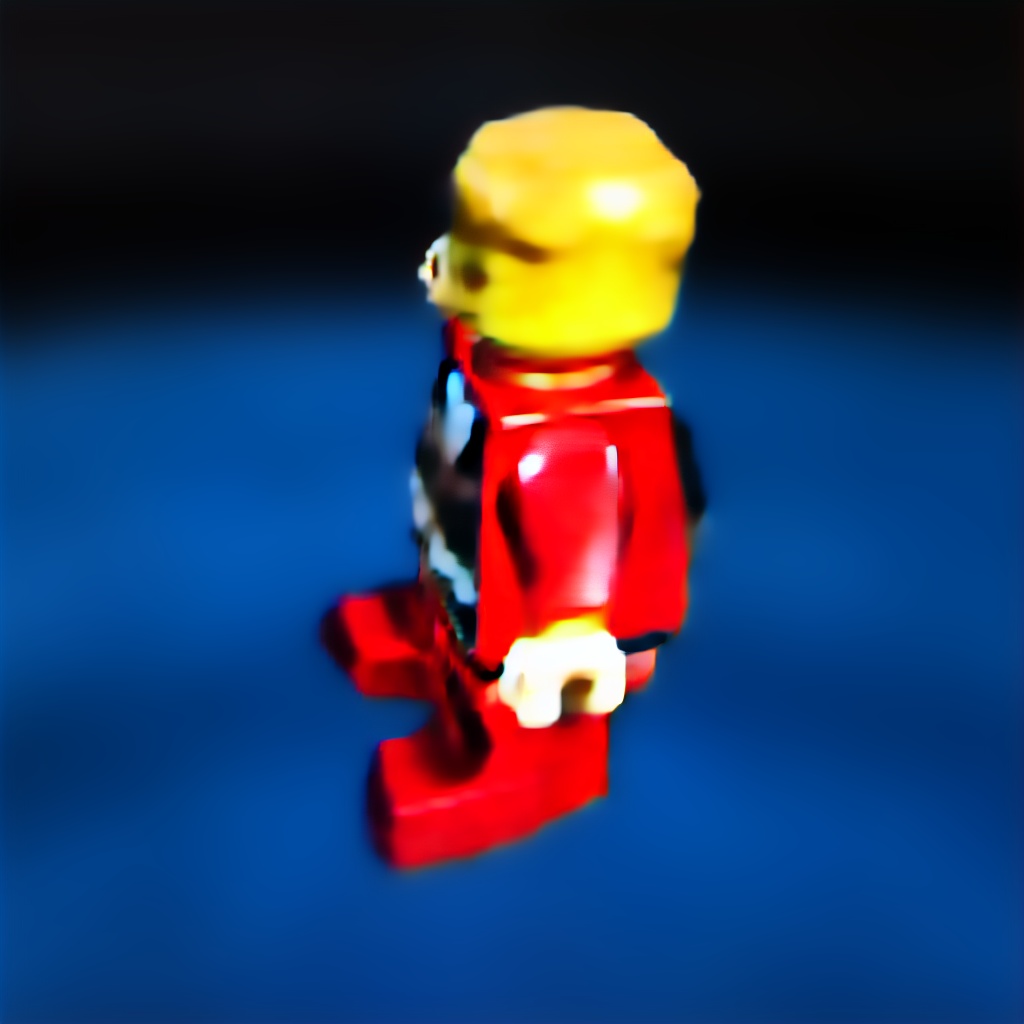}\tabularnewline
    &\multicolumn{3}{c}{``a lego man''} \tabularnewline
    \includegraphics[width=0.24\linewidth]{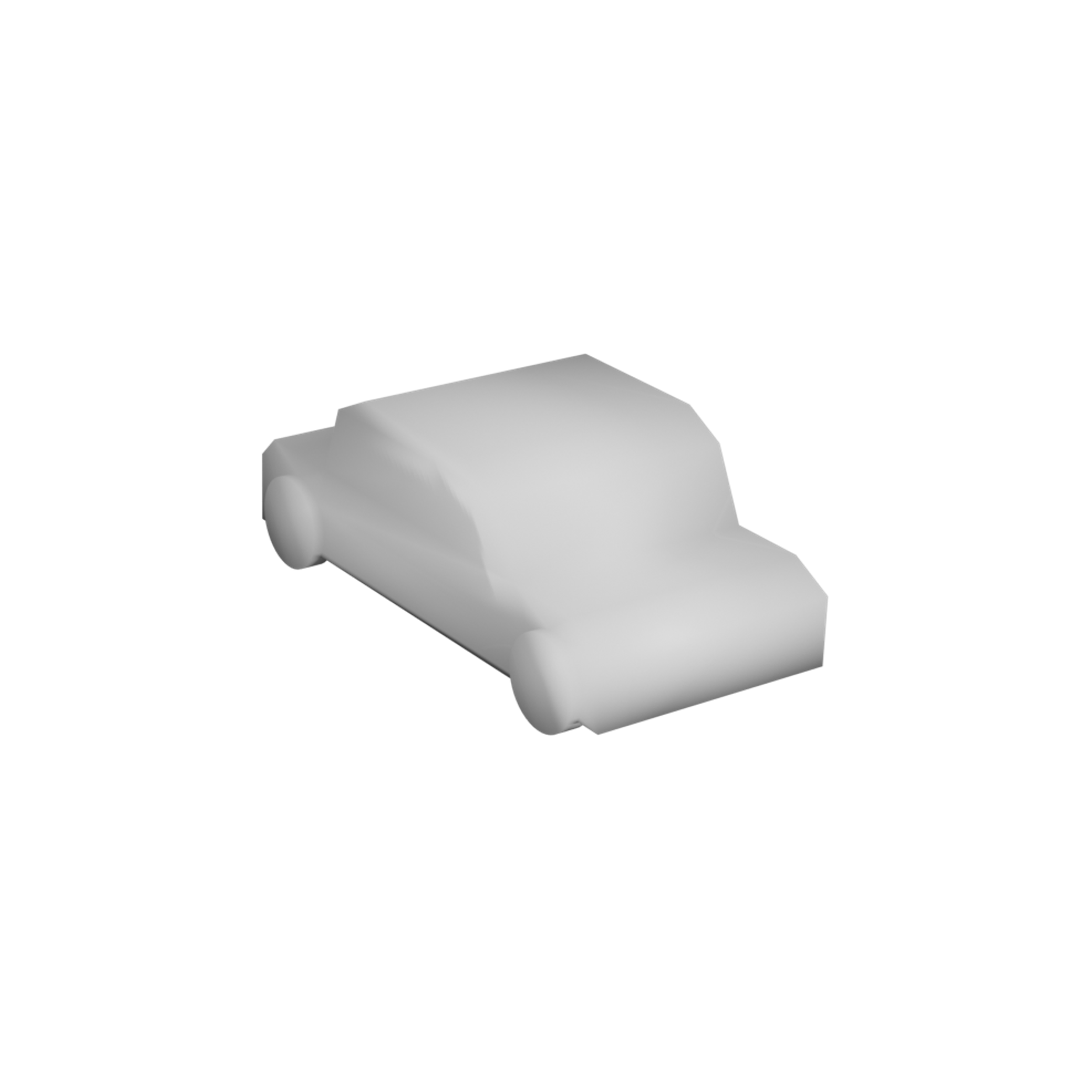} & 
    \includegraphics[width=0.24\linewidth]{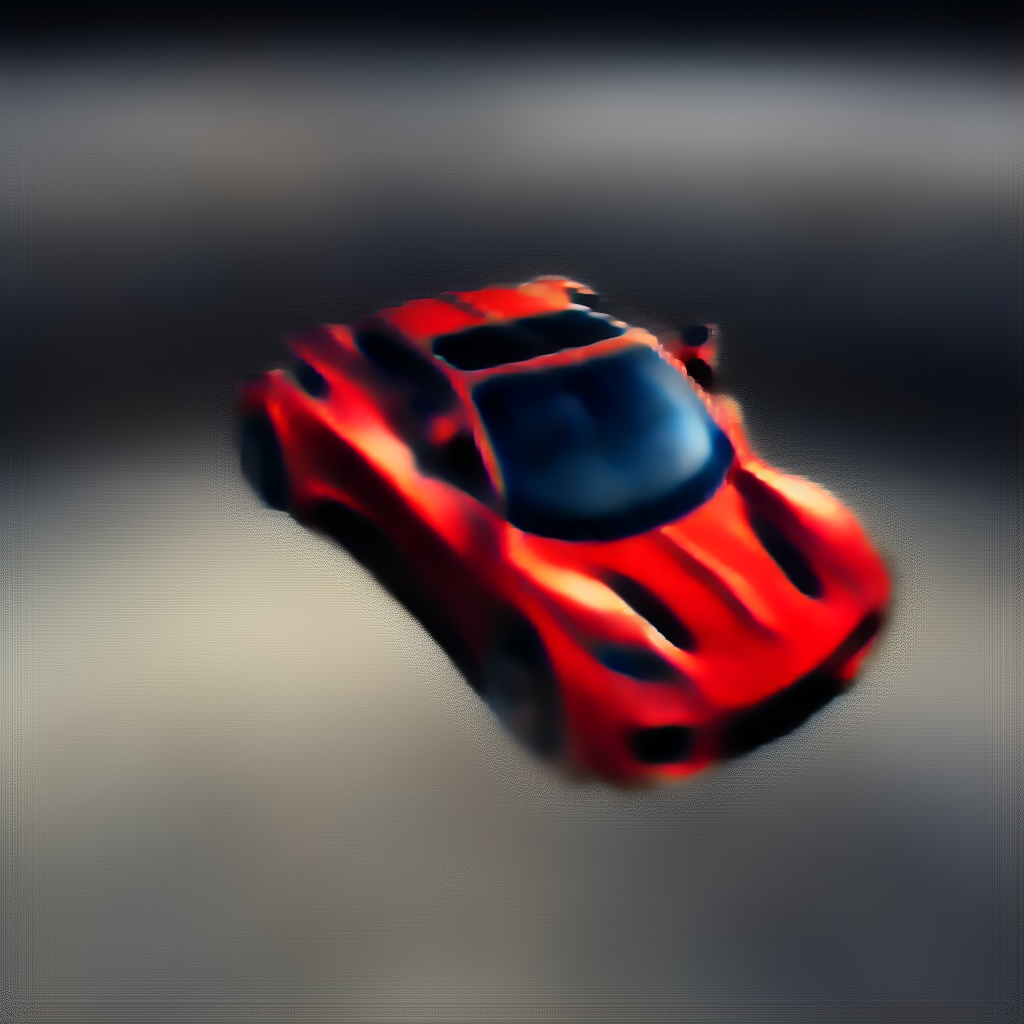} & \includegraphics[width=0.24\linewidth]{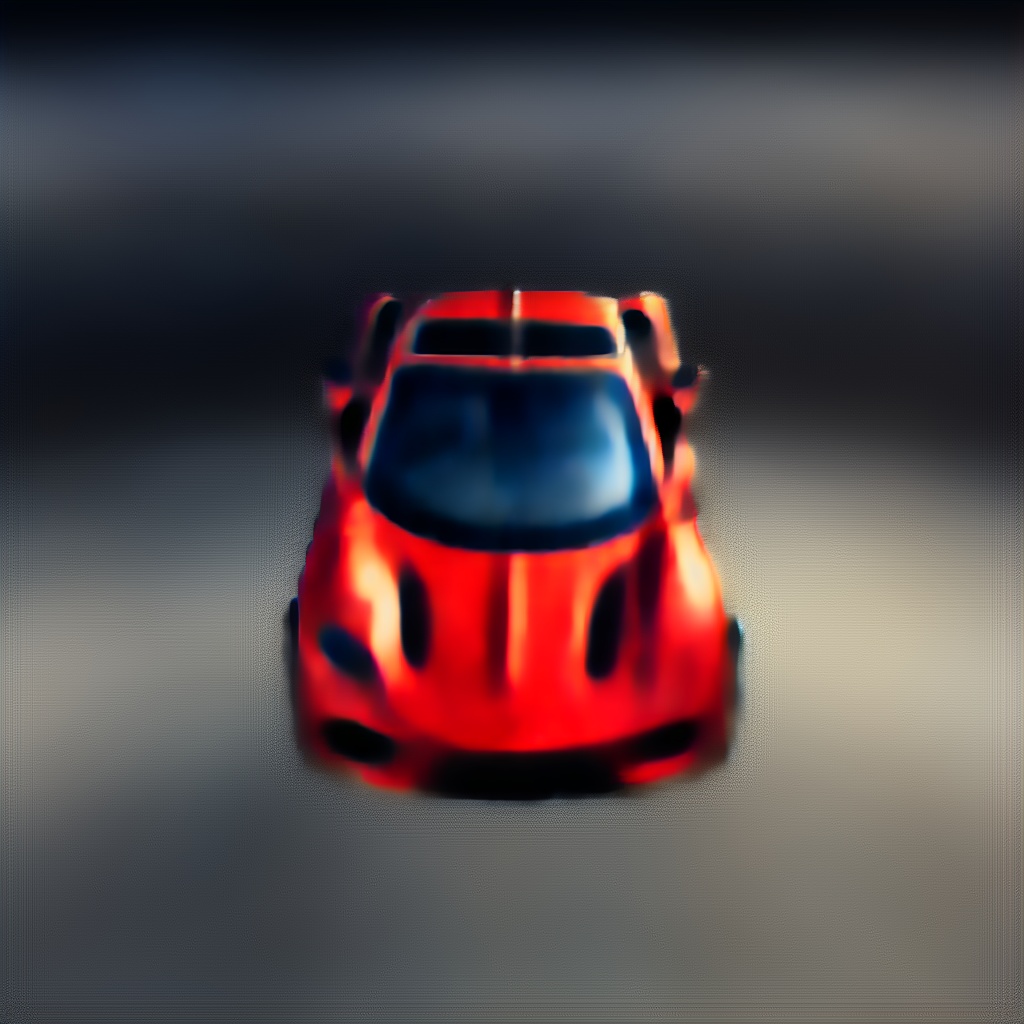} & 
    \includegraphics[width=0.24\linewidth]{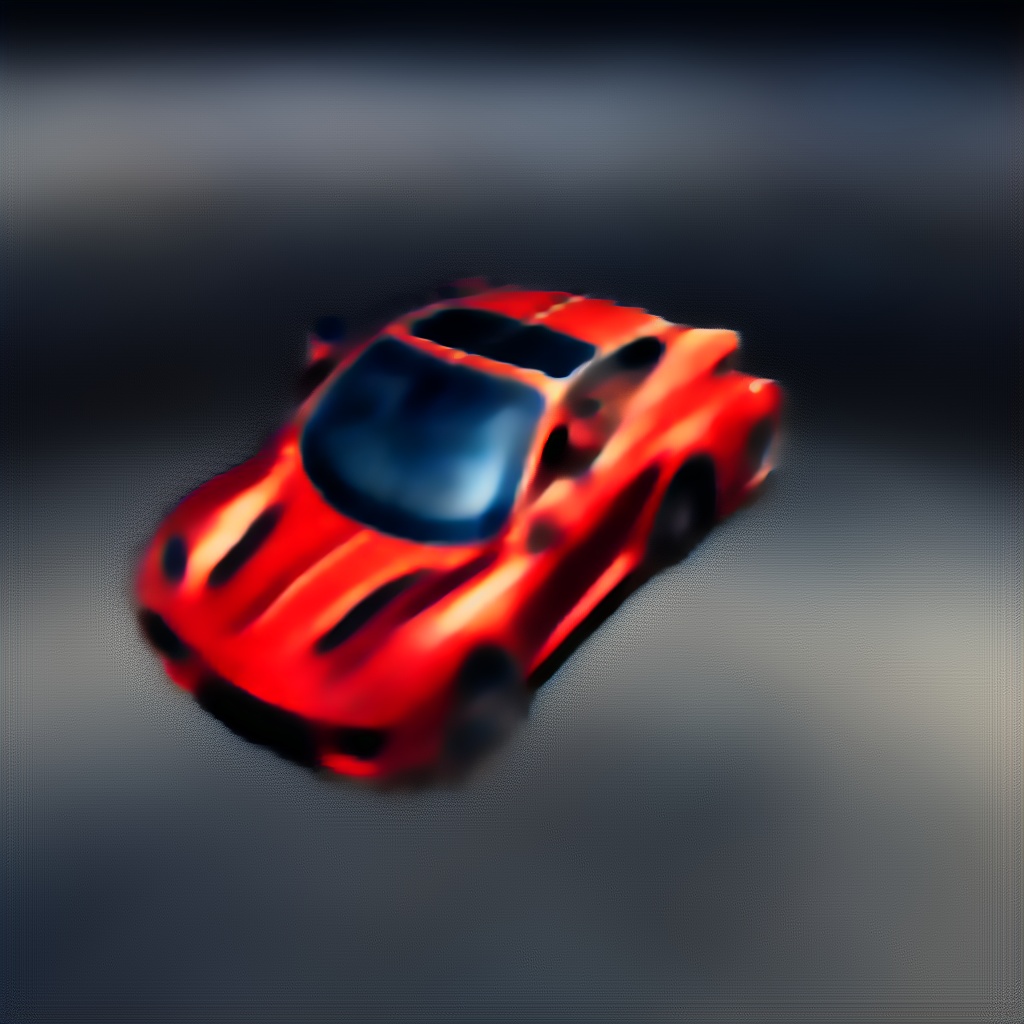}\tabularnewline
    &\multicolumn{3}{c}{`a sports car, highly detailed''} \tabularnewline
    \end{tabular}
    
    }
    \caption{Additional \meshsketch-guided results.}
    \label{fig:general_sketchshapes}
\end{figure} 

%% file: figures/4_exp/latent-mesh/misc/misc.tex
\begin{figure}[h!]
    \centering
    \scriptsize{Input}
    \vspace{-10pt}
    \setlength{\tabcolsep}{0.5pt}
    {\scriptsize
    \begin{tabular}{c c c c}
        \includegraphics[width=0.24\linewidth]{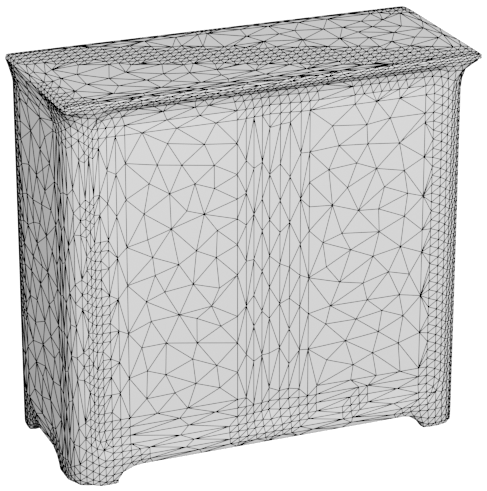} & 
        \includegraphics[width=0.35\linewidth]{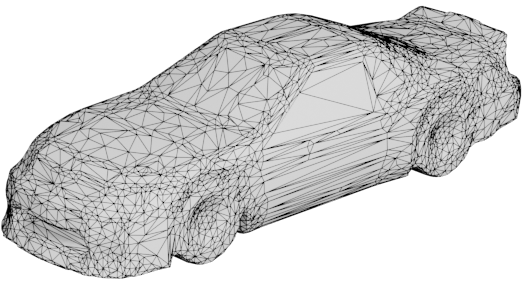} & 
        \includegraphics[width=0.24\linewidth]{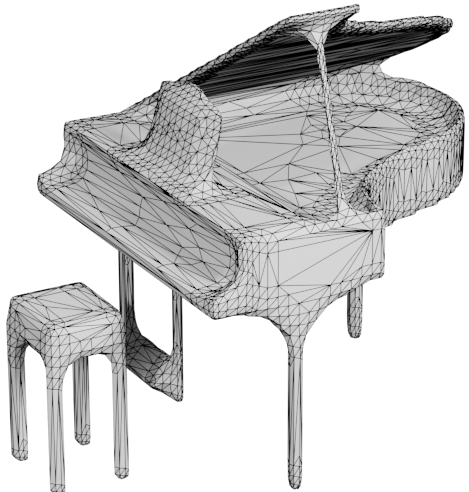} & 
        \includegraphics[width=0.15\linewidth]{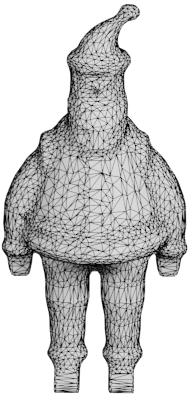} \\
        \includegraphics[width=0.24\linewidth]{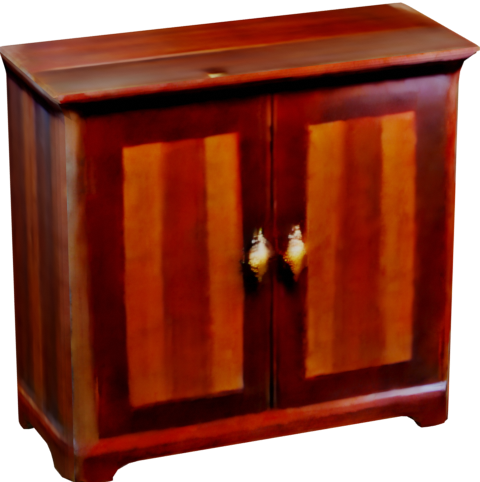} & 
        \includegraphics[width=0.35\linewidth]{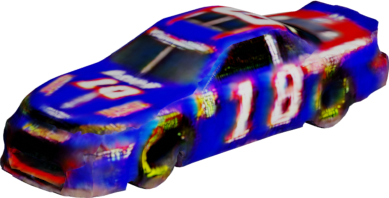} & 
        \includegraphics[width=0.24\linewidth]{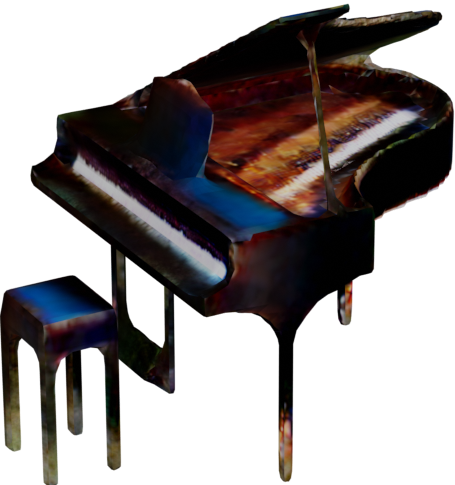} & 
        \includegraphics[width=0.15\linewidth]{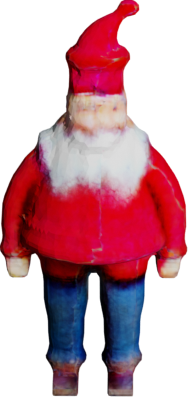} \\
        ``A wooden  & ``A next & ``A black & ``A garden gnome  \\
        brown cabinet'' & gen nascar" & grand piano'' & with a red hat''\\
        \includegraphics[width=0.24\linewidth]{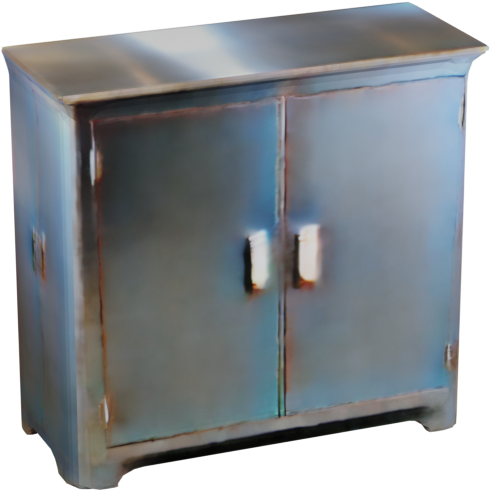} & 
        \includegraphics[width=0.35\linewidth]{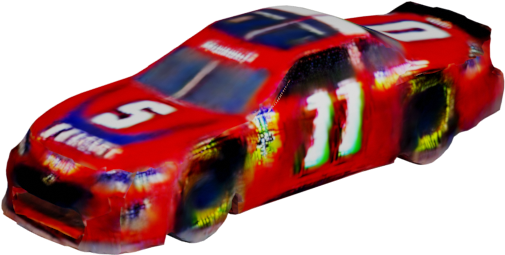} & 
        \includegraphics[width=0.24\linewidth]{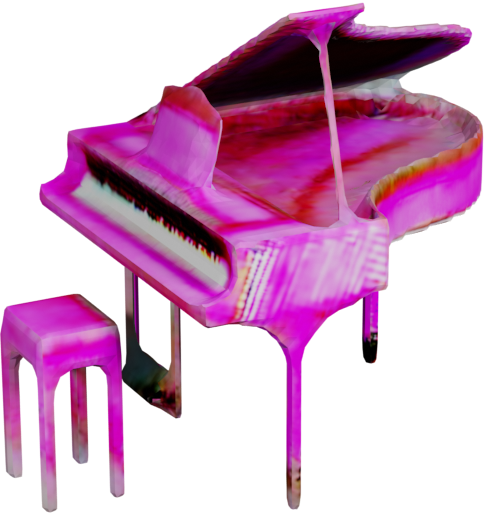} & 
        \includegraphics[width=0.15\linewidth]{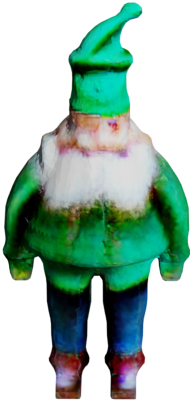} \\
        ``A steel  & ``A next & ``A pink & ``A garden gnome  \\
        cabinet'' & gen nascar" & grand piano'' & with a green hat''
        
    \end{tabular}}
    \vspace{0.1cm}
    \caption{\textit{\latentpaint} results on shapes from  ModelNet40~\cite{wu20153d}. UV parameterization was not given for any of the meshes in this figure.
    For the Nascar shapes, the same text prompt was used with a different initialization. }
    \label{fig:latent_mesh_misc}
\end{figure}

%% file: figures/4_exp/latent-mesh/shoes2.tex
\begin{figure}[h]
    \vspace{-6pt}

    \centering
    \setlength{\tabcolsep}{1pt}
    {\scriptsize
    \begin{tabular}{c c}
        \raisebox{0.050\textwidth}{\rotatebox[origin=t]{90}{Tango~\cite{chen2022tango}}} & 
        \includegraphics[width=0.95\linewidth]{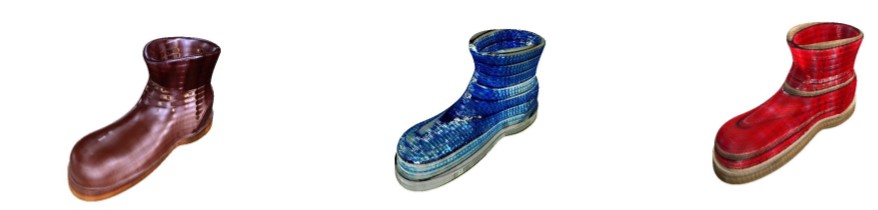} \\
        \raisebox{0.050\textwidth}{\rotatebox[origin=t]{90}{CLIPMesh~\cite{khalid2022clipmesh}}} & 
        \includegraphics[width=0.95\linewidth]{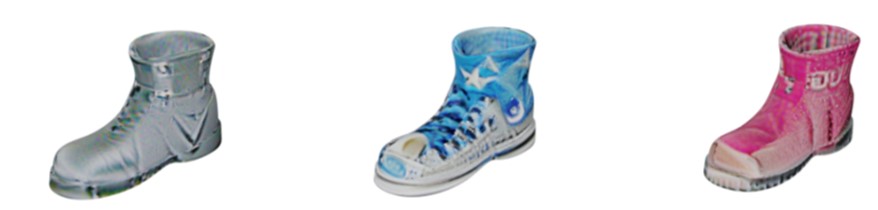} \\
        \raisebox{0.050\textwidth}{\rotatebox[origin=t]{90}{\latentpaint~(Ours)}} & 
        \includegraphics[width=0.95\linewidth]{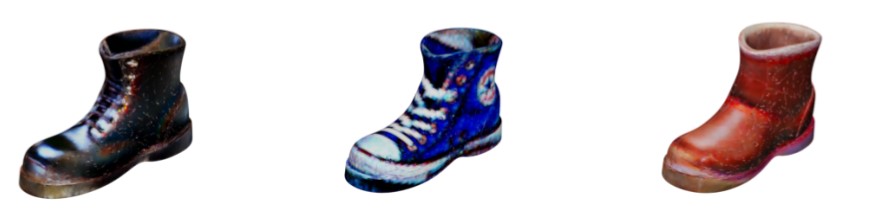} \\
        &
        \begin{tabular*}{0.98\linewidth}{P{0.28\linewidth}P{0.35\linewidth}P{0.36\linewidth}@{}}
        \centering
        ``A black boot'' & ``A blue converse allstar shoe'' & ``UGG boot''
        \end{tabular*}
    \end{tabular}
    }
    \vspace{-10pt}
    \caption{Generating different shoe textures over the same input mesh by only changing the conditioning text prompt. The surface parameterization was computed using XAtlas~\cite{xatlas}.}
    \vspace{-2pt}
    
    \label{fig:latent_mesh_shoe}    
\end{figure} 

%% file: figures/4_exp/sketch_mesh/ablation/fig.tex
\begin{figure}
    \centering
    \setlength{\tabcolsep}{0.5pt}
    {\scriptsize
    \begin{tabular}{c c c c c}
        Shape guidance & No shape guidance & Shape guidance & No shape guidance \\
        \includegraphics[width=0.24\linewidth]{figures/4_exp/sketch_mesh/general/images/hand_ep0050_0002_rgb.jpg} & 
        \includegraphics[width=0.24\linewidth]{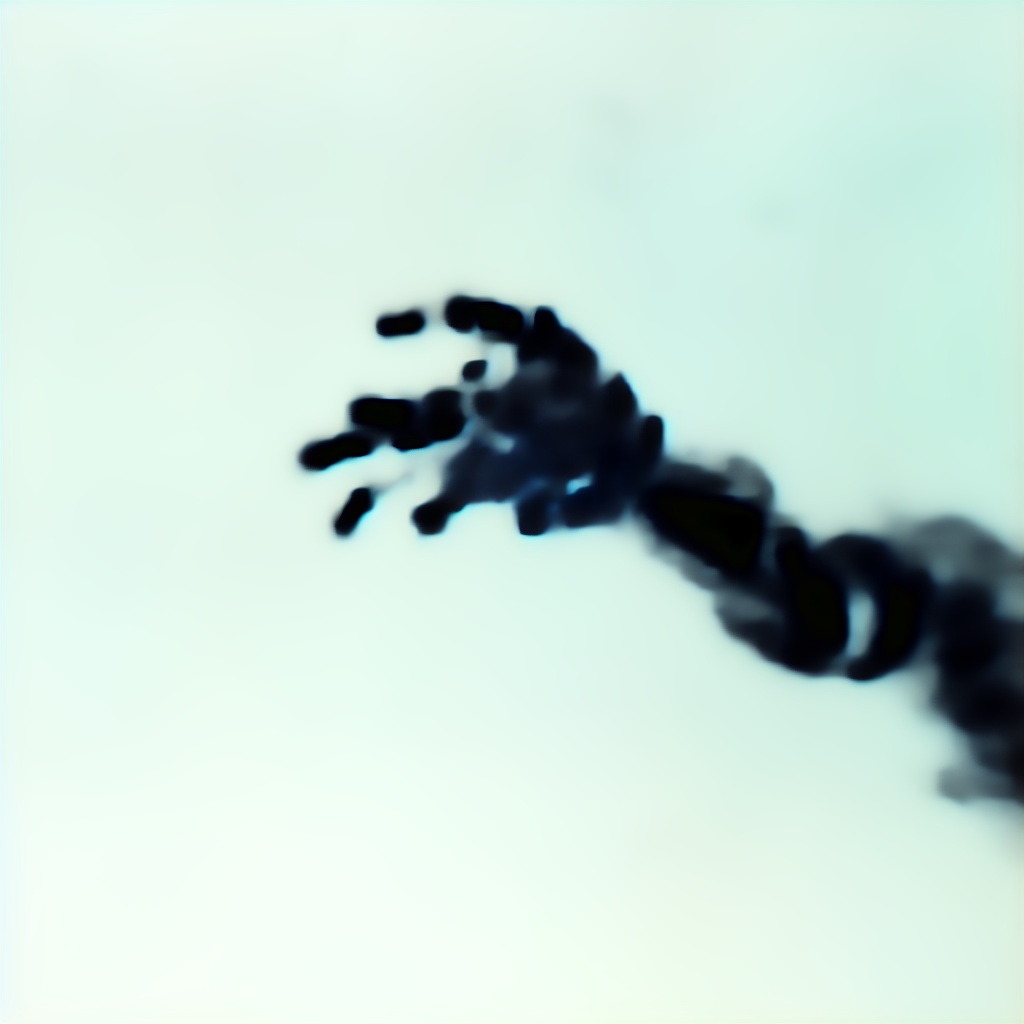} & 
        \includegraphics[width=0.24\linewidth]{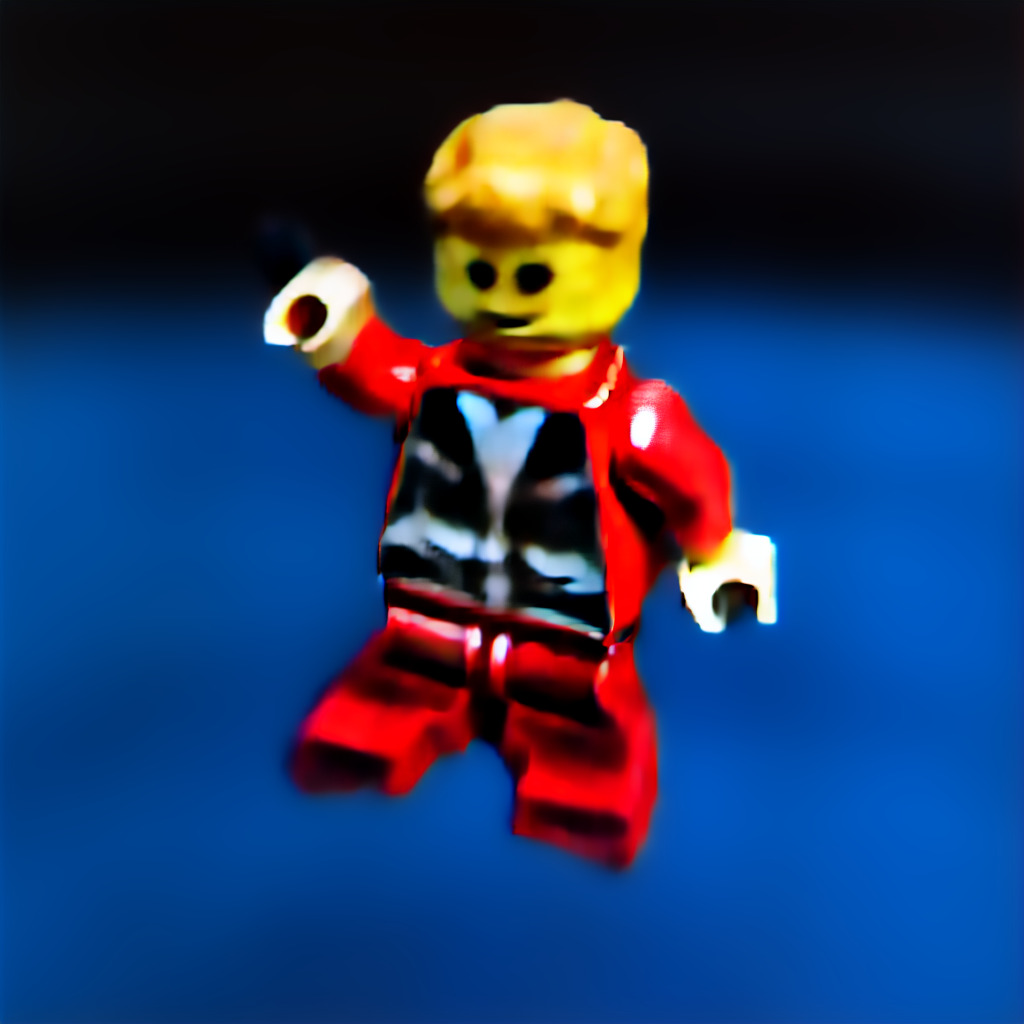} & 
        \includegraphics[width=0.24\linewidth]{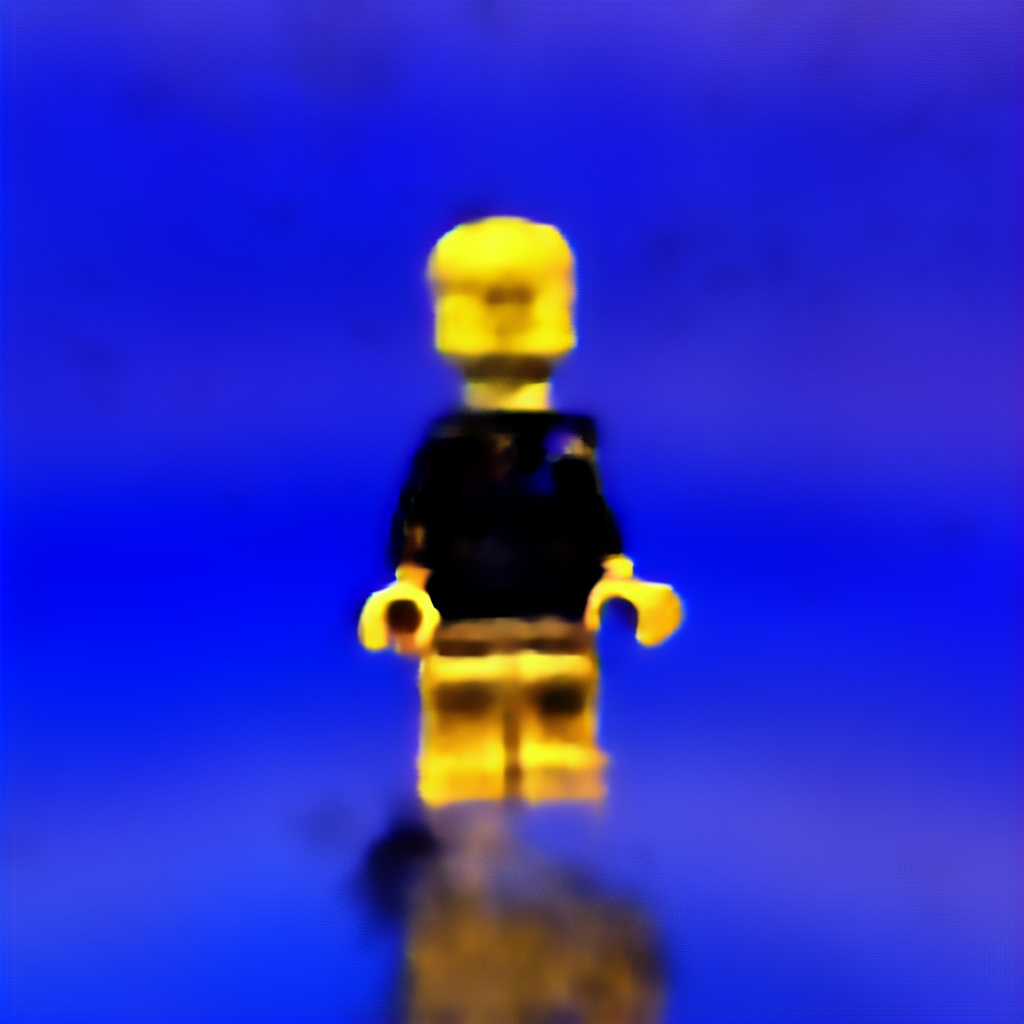} \\

        \multicolumn{2}{c}{``A robot hand, realistic''} & \multicolumn{2}{c}{``a lego man''} & 
        
    \end{tabular}}
    \vspace{-5pt}
    \caption{Results for the same prompts with and without shape-guidance. The guiding shapes are the same as in Figure~\ref{fig:general_sketchshapes}.}
    \label{fig:ablation_prompt}
\end{figure} 

%% file: figures/4_exp/latent-mesh/fish.tex
\begin{figure}[h!]
    \centering
    {\scriptsize
    \newcommand{\pl}{-8.0}
    \begin{overpic}[width=\columnwidth]{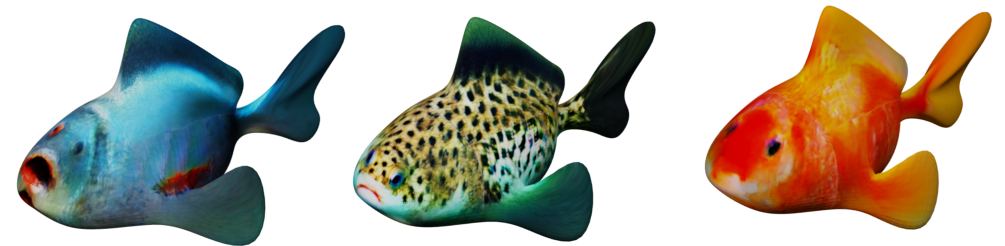}
    \end{overpic}
    \begin{tabular*}{\linewidth}{P{0.26\linewidth}P{0.26\linewidth}P{0.33\linewidth}@{}}
    \centering
    ``Piranha Fish'' & ``A fish with leopard spots'' & ``Goldfish''
    \end{tabular*}
    }
    \vspace{-8pt}
    \caption{\textit{\latentpaint} results for an input mesh containing precomputed UV parameterization. Model courtesy of "Keenan's 3D Model Repository"~\cite{keenan3D}.}
    \label{fig:latent_mesh_fish}
\end{figure}

%% file: figures/4_exp/limitations/fig.tex
\begin{figure}
    \centering
    \setlength{\tabcolsep}{0.5pt}
    {\scriptsize
    \begin{tabular}{c c c c c}
        \includegraphics[width=0.19\linewidth]{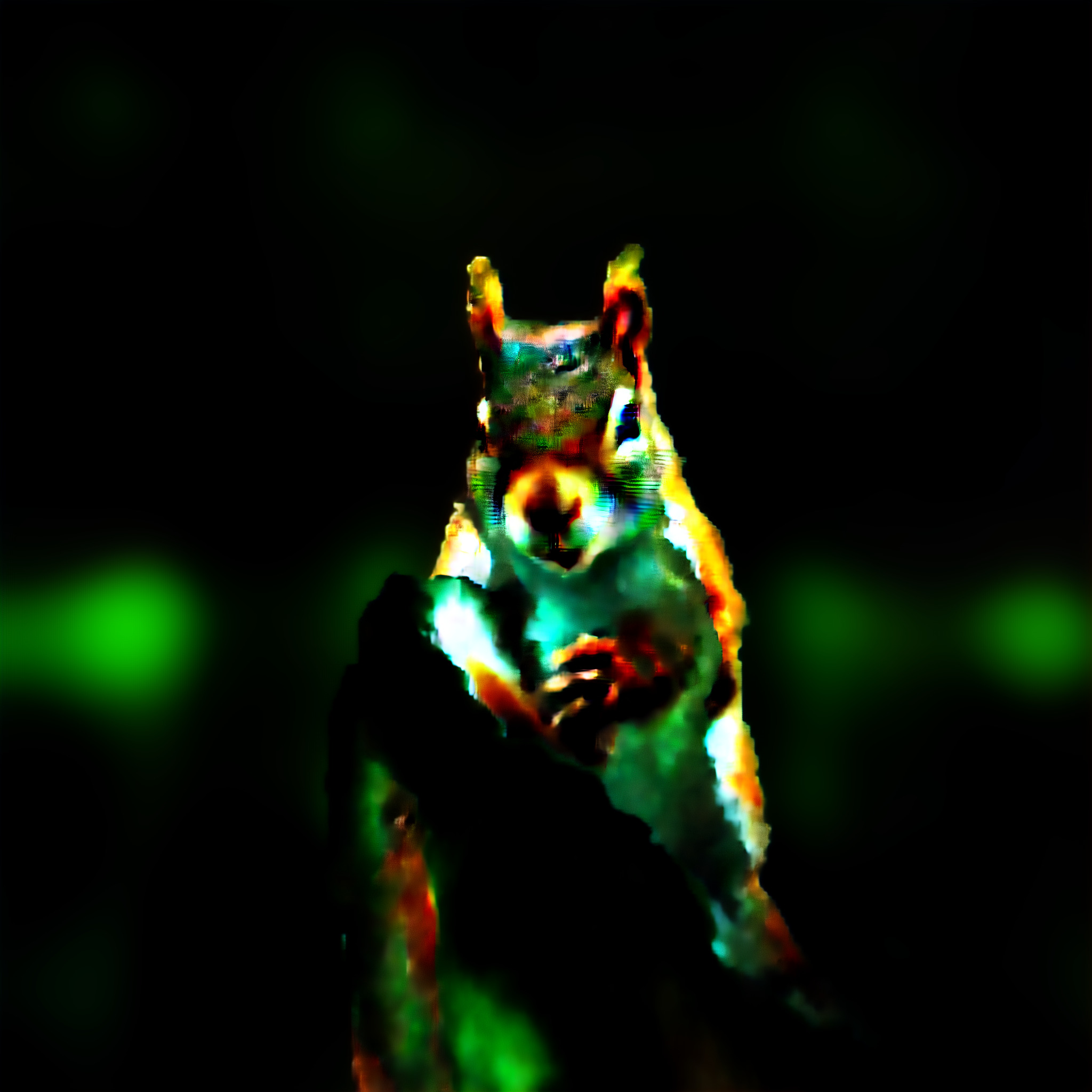} & 
        \includegraphics[width=0.19\linewidth]{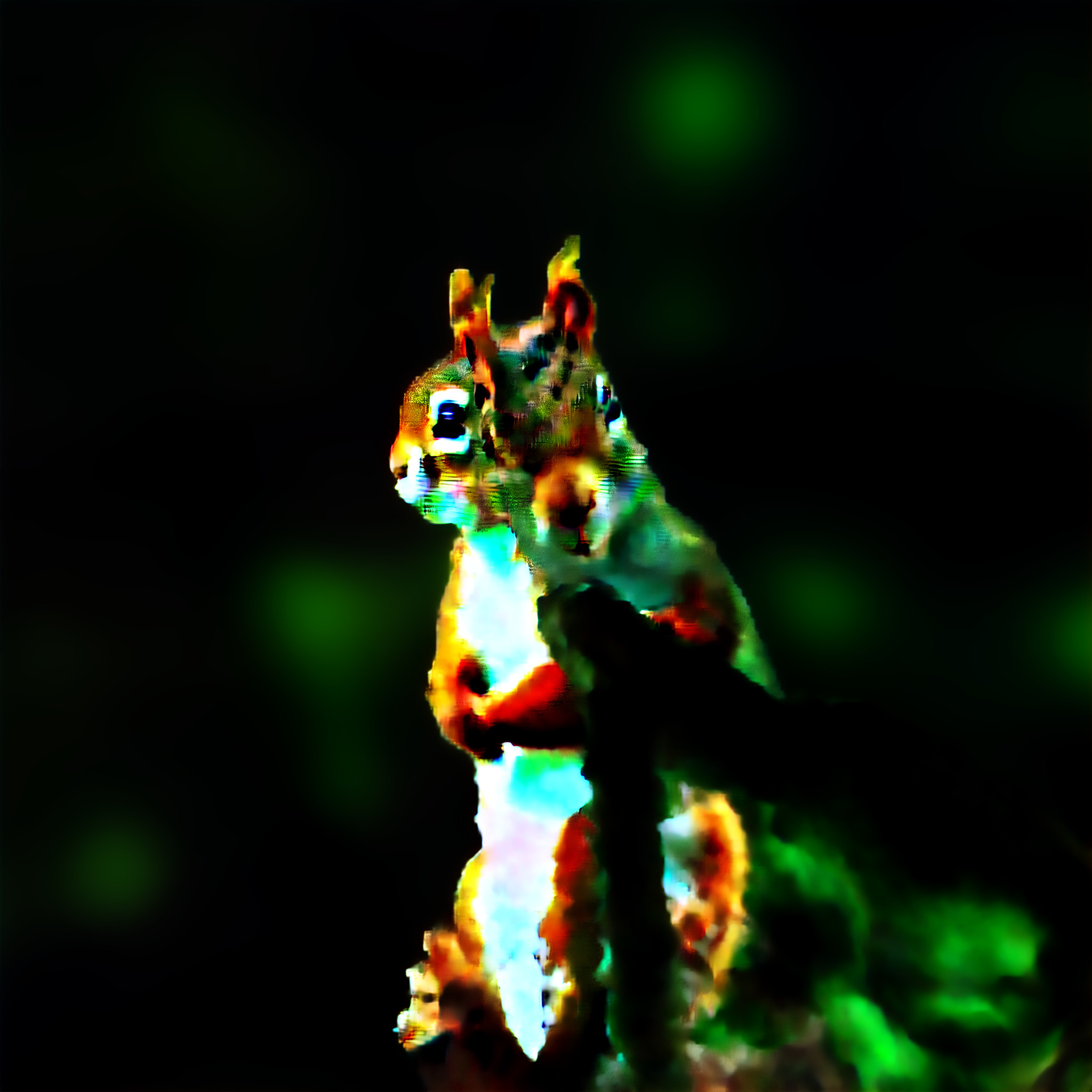} & 
        \includegraphics[width=0.19\linewidth]{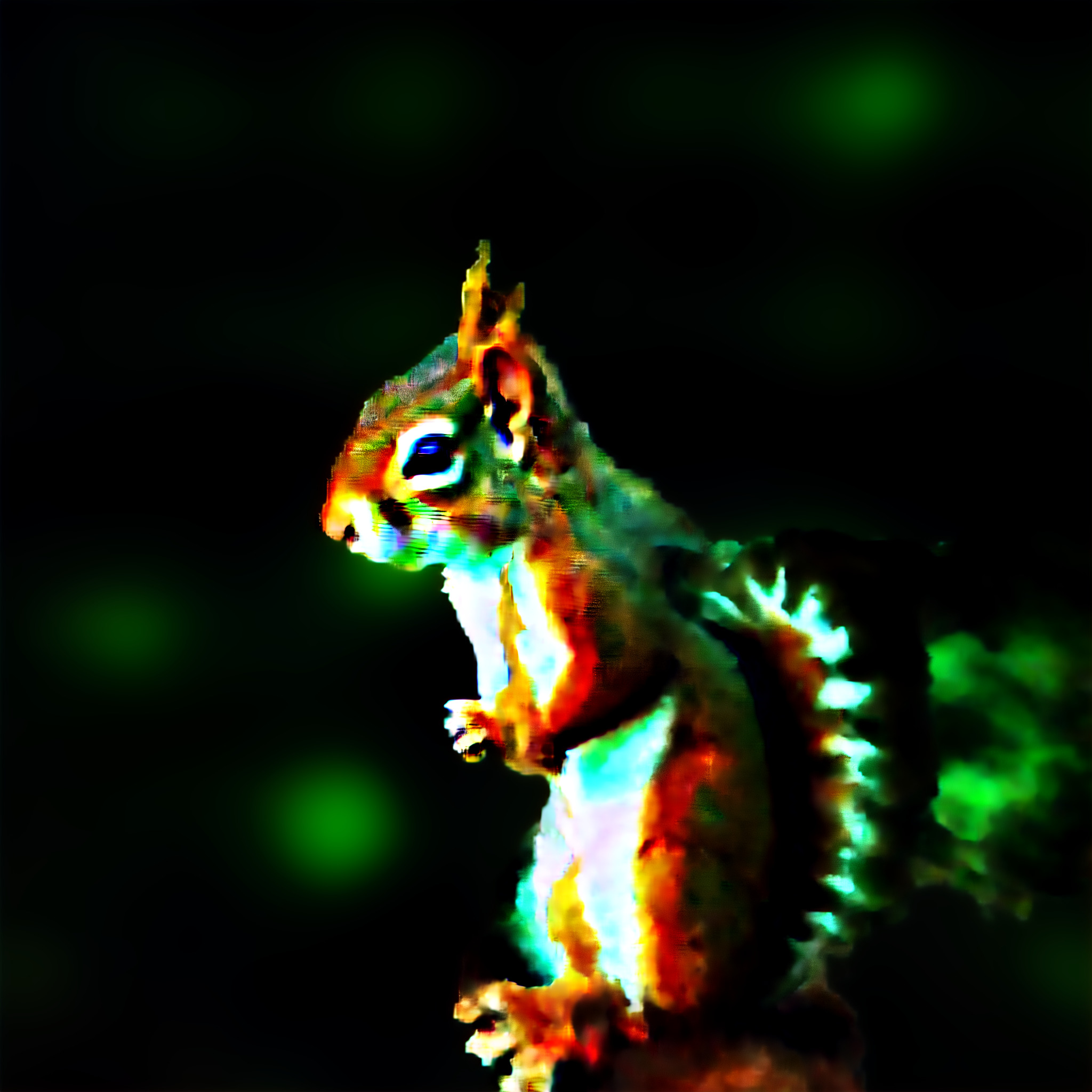} & 
        \includegraphics[width=0.19\linewidth]{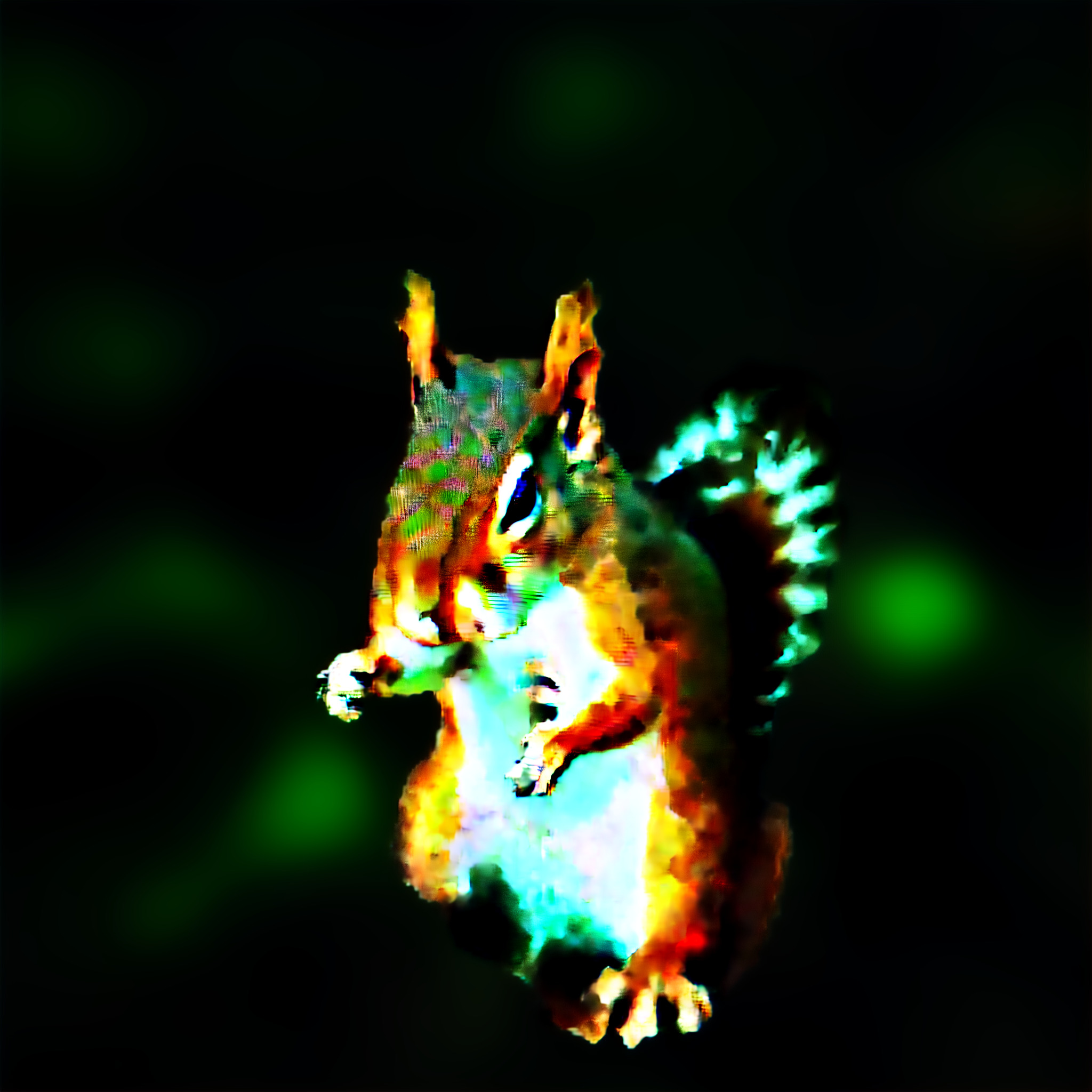} &
        \includegraphics[width=0.19\linewidth]{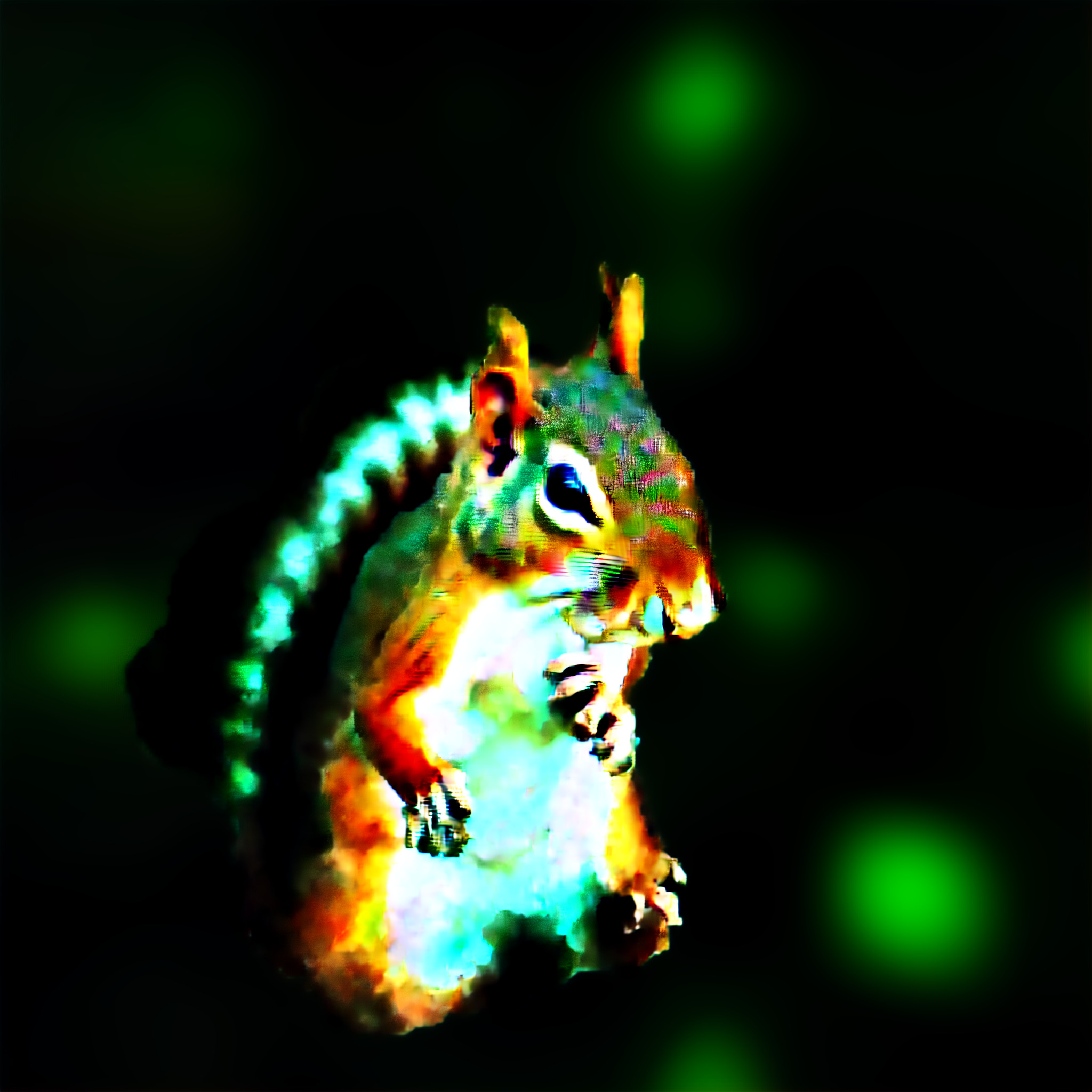} \\

        \multicolumn{5}{c}{Latent-NeRF with ``A photo of a squirrel''} \\
        \includegraphics[width=0.19\linewidth]{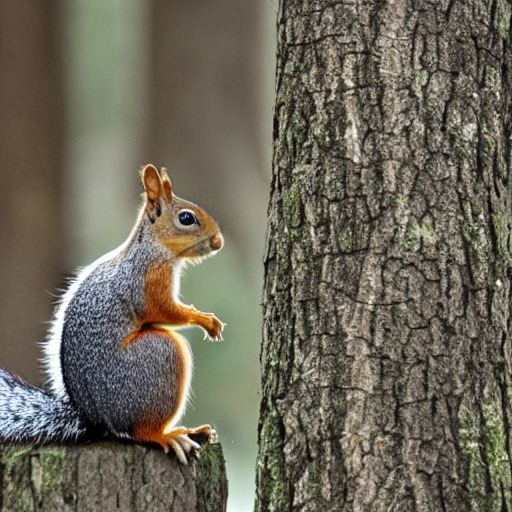} & 
        \includegraphics[width=0.19\linewidth]{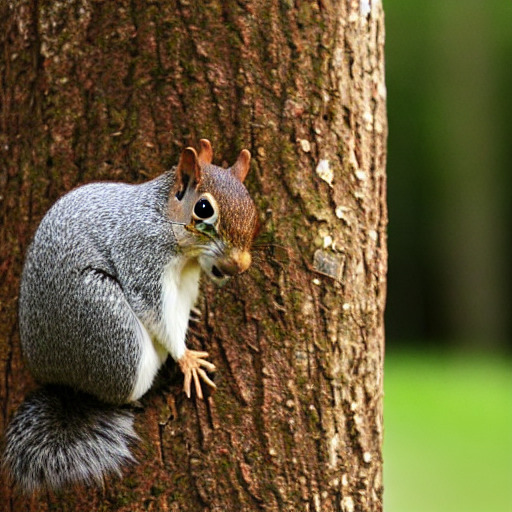} & 
        \includegraphics[width=0.19\linewidth]{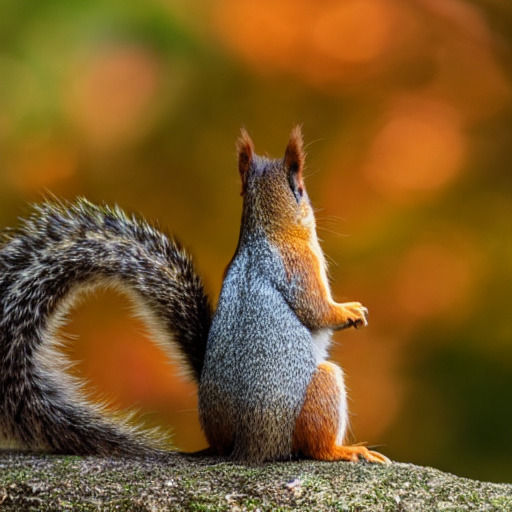} & 
        \includegraphics[width=0.19\linewidth]{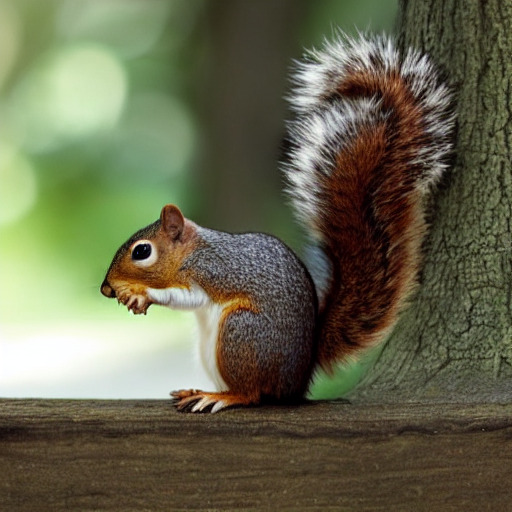} &
        \includegraphics[width=0.19\linewidth]{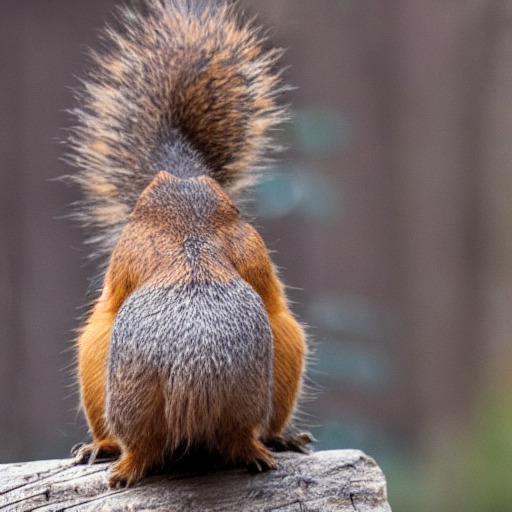} \\

        \multicolumn{5}{c}{Stable Diffusion with ``A photo of a squirrel, \textbf{backview}''}
        
    \end{tabular}}
    \vspace{-7pt}
    \caption{note that the generated squirrel in the first row contains two faces from some view directions; see the second to the left view. This problem may be attributed to the fact that Stable Diffusion fails to generate back views of a squirrel.}
    \label{fig:limitations}
\end{figure}

%% file: figures/4_exp/sketch_mesh/weight_ablation/fig.tex
\begin{figure}[h!]
    \centering
    \vspace{-7pt}
    \setlength{\tabcolsep}{0.5pt}
    {\scriptsize
    \begin{tabular}{c c c c c}
         \includegraphics[width=0.2\linewidth]{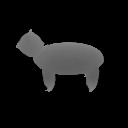} &
        \includegraphics[width=0.2\linewidth]{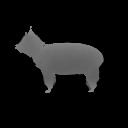} & 
        \includegraphics[width=0.2\linewidth]{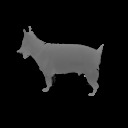} & 
        \includegraphics[width=0.2\linewidth]{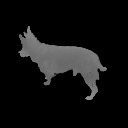} & 
        \includegraphics[width=0.2\linewidth]{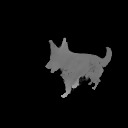} \\
        \includegraphics[width=0.2\linewidth]{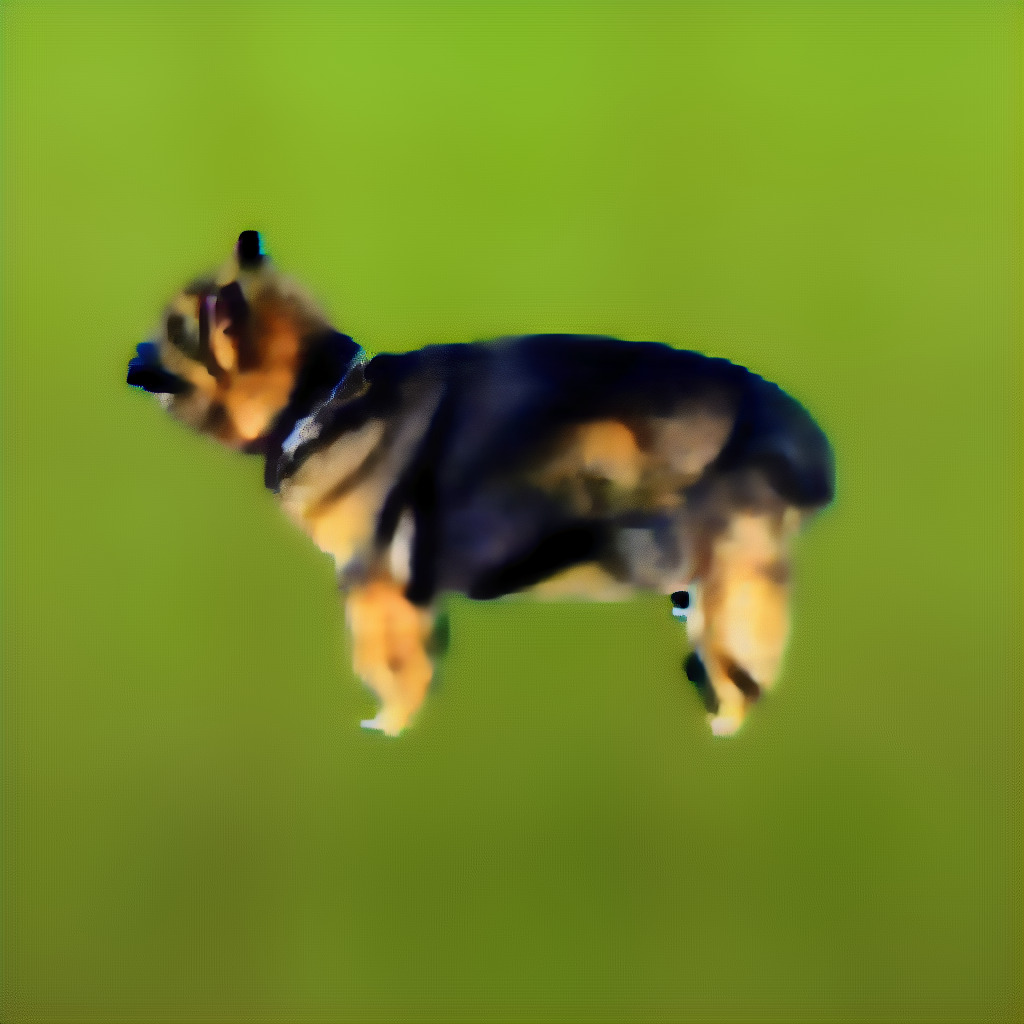} &
        \includegraphics[width=0.2\linewidth]{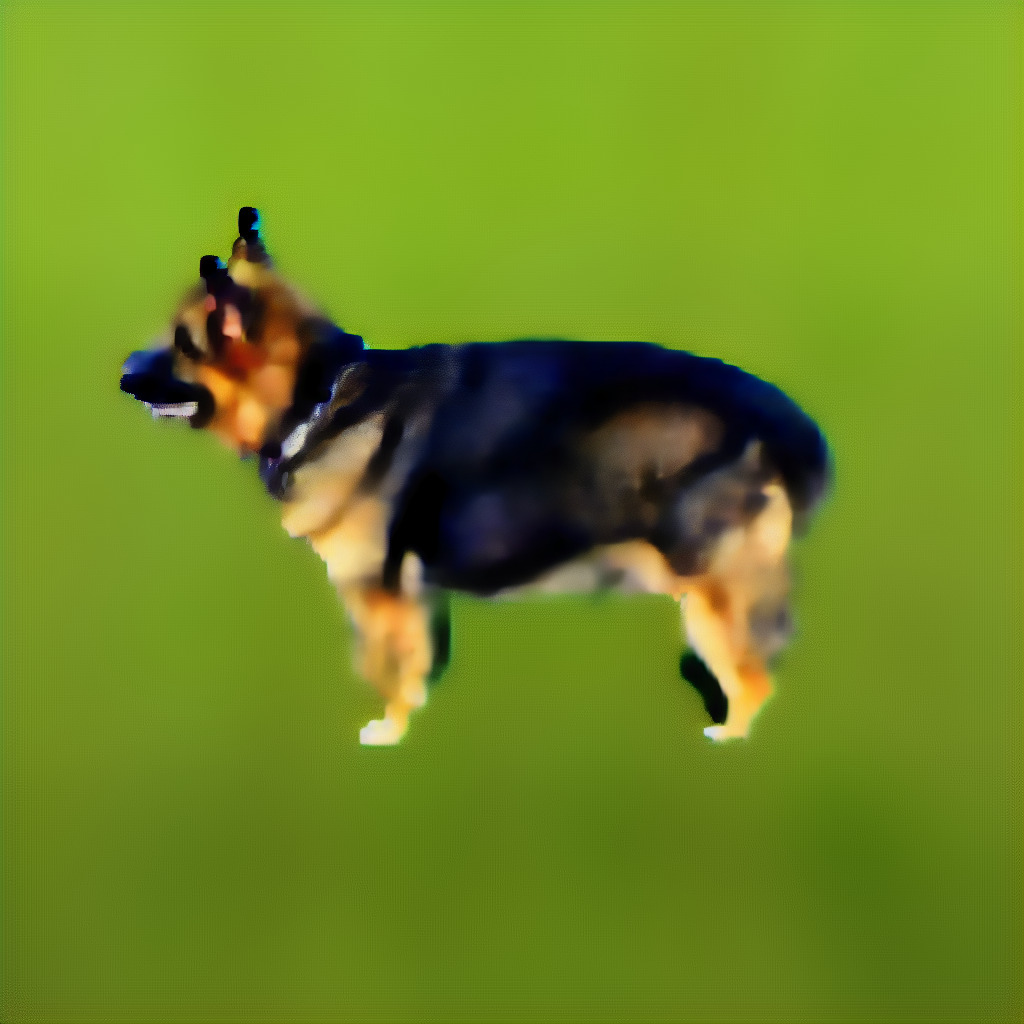} & 
        \includegraphics[width=0.2\linewidth]{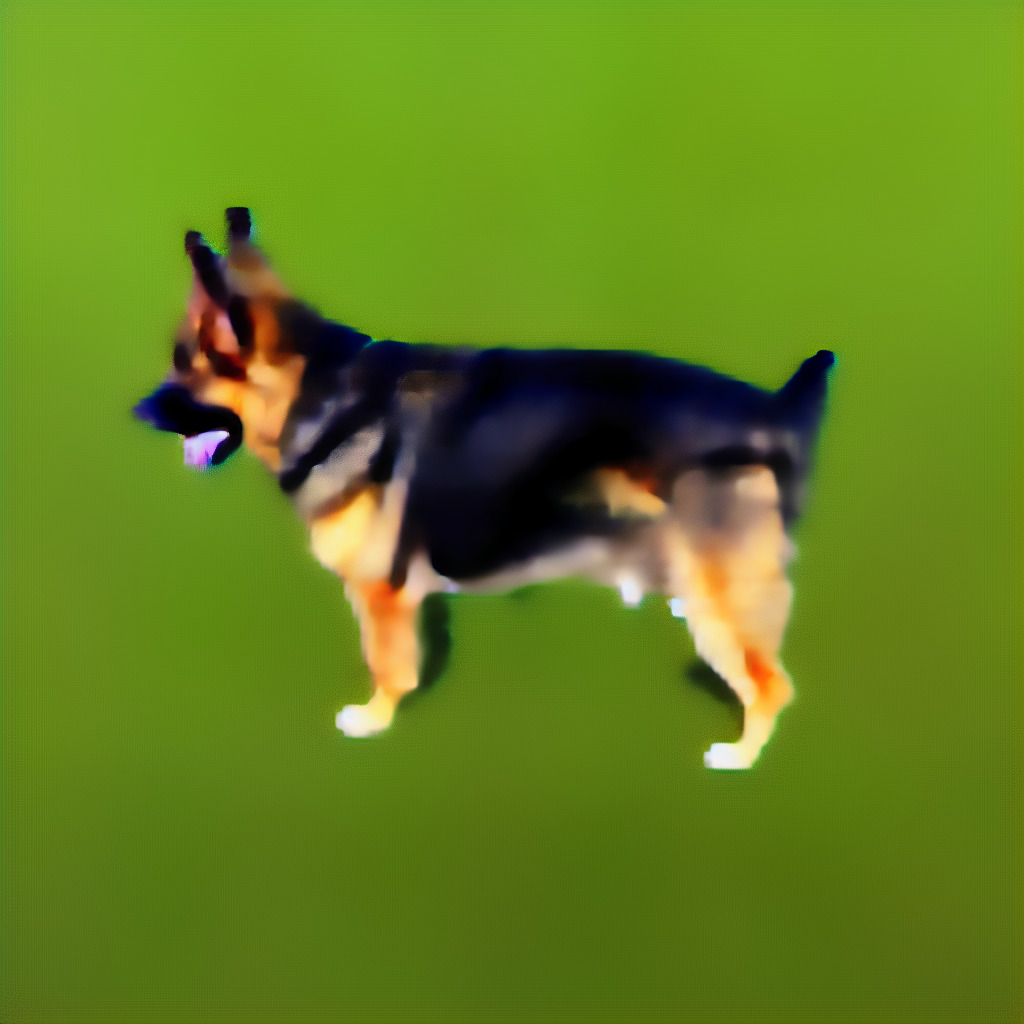} & 
        \includegraphics[width=0.2\linewidth]{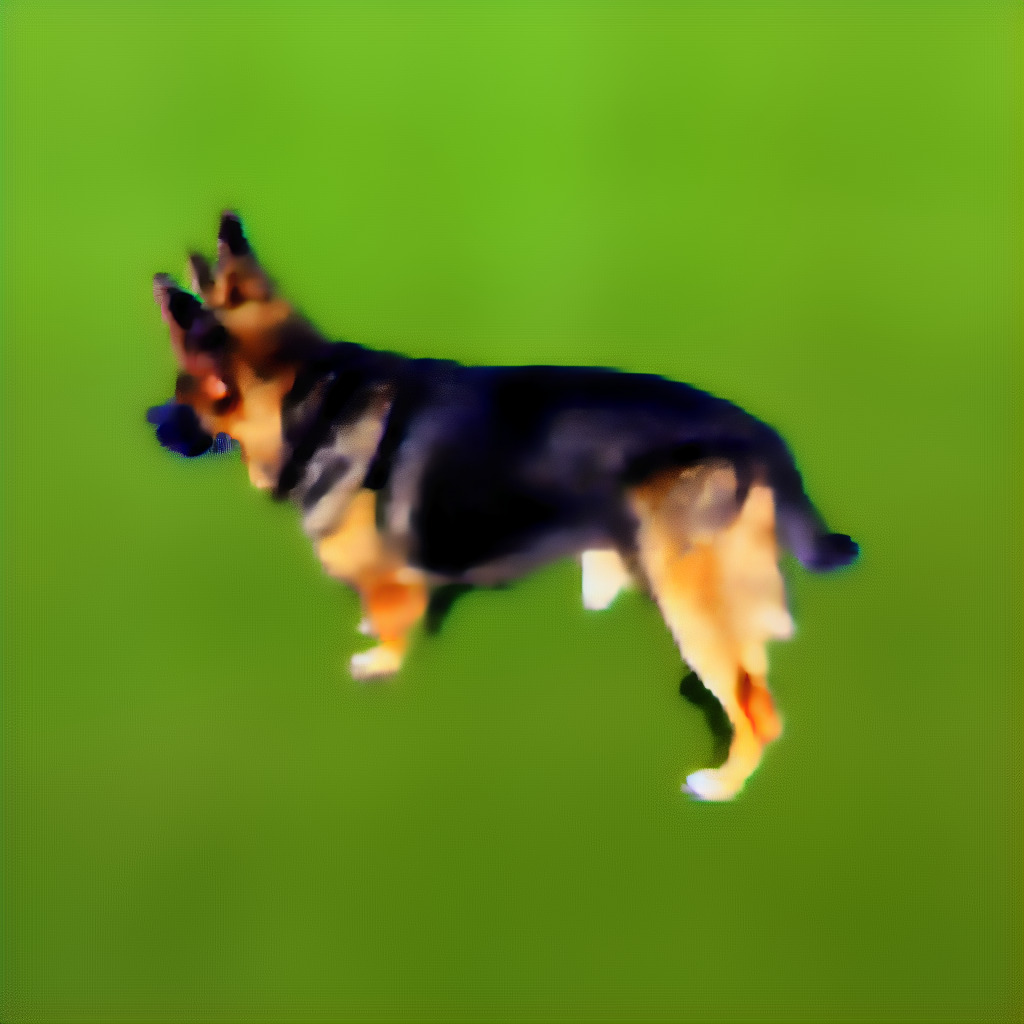} & 
        \includegraphics[width=0.2\linewidth]{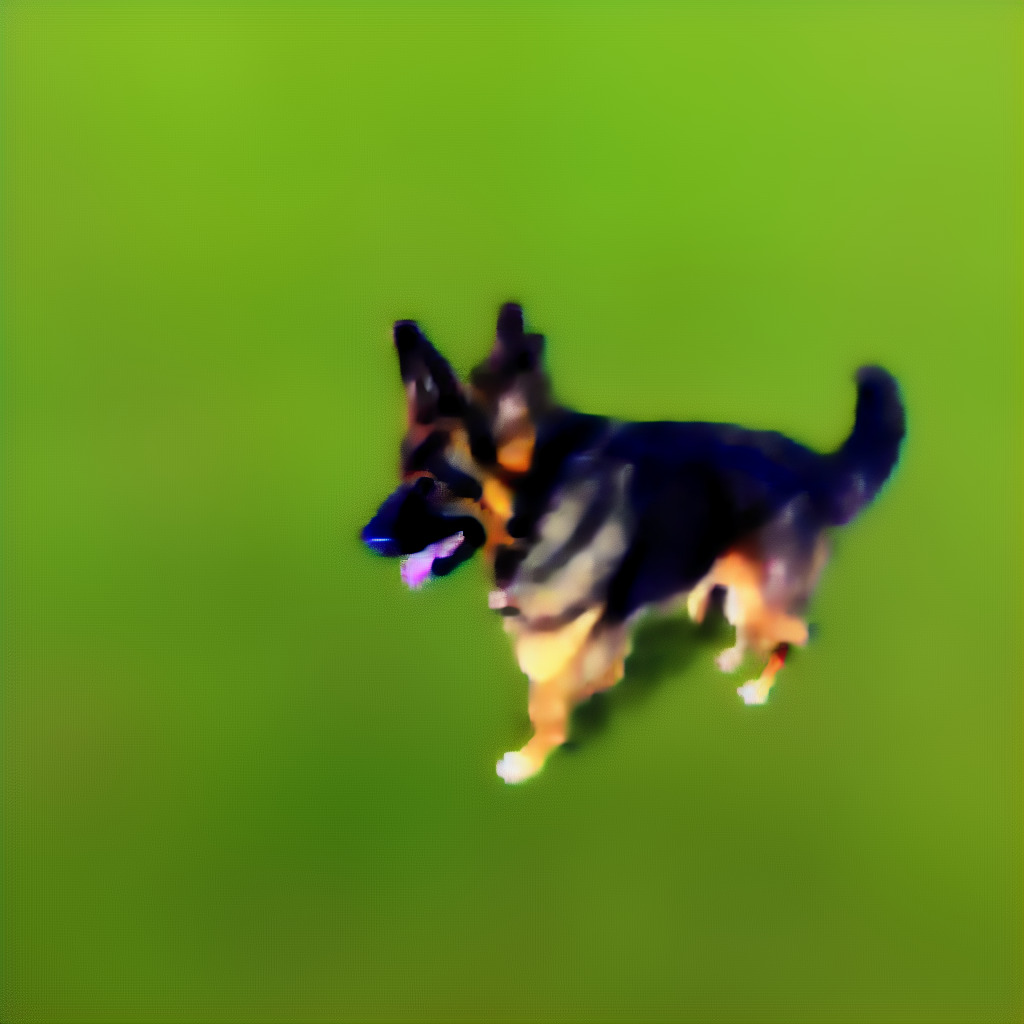} \\
        0.05 & 0.1 & 0.3 & 0.7 & 1.5
        
    \end{tabular}}
    \vspace{-7pt}
    \caption{\meshsketch{} ablation over $\sigma_S$ values (\cref{eq:lenient_constraint}) and input shape from Figure~\ref{fig:skecth_mesh_animals}. As $\sigma_S$ grows, the constraint becomes more lenient, and enables the shape to evolve into new geometries.}
    \vspace{-14pt}
    
    \label{fig:skecth_weight_ablation}
\end{figure}

%% file: 6_conclusion.tex
\section{Limitations}
Our presented technique is yet a preliminary step towards the challenging goal of a comprehensive text-to-shape model that uses no 3D supervision.
Still, the proposed latent framework has its limitations.
To attain a plausible 3D shape, we use the same ``prompt tweaking'' used by DreamFusion~\cite{poole2022dreamfusion}, \ie, adding a directional text-prompt (e.g., "front", "side" with respect to the camera) to the input text prompt. We find that this assistance tends to fail with our approach when applied to certain objects.
Moreover, we find that even Stable Diffusion tends to generate unsatisfactory images when specifying the desired direction as shown in Figure~\ref{fig:limitations}.
Additionally, similar to most works that employ diffusion models, there is a stochastic behavior to the results, such that the quality of the results may significantly vary between different seeds.

\section{Conclusions}
\label{sec:conclusion}
In this work, we introduced a latent framework for generating 3D shapes and textures using different forms of text and shape guidance. We first adapted the score distillation loss for LDMs, enabling the use of recent, powerful and publicly available text-to-image generation models for 3D shape generation.
Successfully applying score distillation on LDMs results in a fast and flexible object generation framework. We then introduced shape-guided control on the generated model. We showed two versions of shape-guided generation, \meshsketch{} and \latentpaint{}, and demonstrated their effectiveness for providing additional control over the generation process. 

Typically, the notion of rendering refers to generating an image in pixel space. Here, we have presented a method that renders directly into the latent space of a neural model. We believe that our Latent-NeRF approach opens the avenue for more latent space rendering methods, which can gain from a compact and effective latent representation, and advance the use of neural models that operate in latent space rather than pixel space.
Furthermore, our novel approach and its ease-of-use nature would encourage further research toward effective text-guided shape generation.

\subsection{Acknowledgement}
We thank Yuval Alaluf, Rinon Gal and Kfir Goldberg for their insightful comments. 
We would also like to thank Ido Richardson for his excellent mesh model designs used throughout the paper.